\RequirePackage[svgnames,table]{xcolor}

\documentclass[11pt,letterpaper]{mystyle}

\usepackage[all]{hypcap}
\usepackage[comma,numbers,compress]{natbib}

\tcbuselibrary{skins,breakable}

\usepackage{bxcoloremoji}

\usepackage{multirow}
\usepackage{adjustbox}
\usepackage{wrapfig}
\usepackage{placeins}
\usepackage{xspace}
\usepackage{array}
\usepackage{nicefrac}
\usepackage{xurl}
\usepackage{hyperref}
\usepackage{tcolorbox}

\newcommand{\sys}{MemEye\xspace}
\newcommand{\bestcell}[1]{\cellcolor{orange!18}{#1}}
\newcommand{\secondcell}[1]{\cellcolor{blue!18}{#1}}

\newcommand{\modelname}[1]{\nolinkurl{#1}}
\definecolor{darkblue}{rgb}{0.3, 0.5, 0.9}
\definecolor{selfblue}{RGB}{65,105,225}
\definecolor{Gray}{gray}{0.93}
\hypersetup{citecolor=darkblue, linkcolor=darkblue, urlcolor=darkblue}
\definecolor{axisHeaderGray}{RGB}{125,125,125}
\definecolor{axisBodyGray}{RGB}{248,248,248}
\definecolor{axisBorder}{RGB}{20,20,20}
\newtcolorbox{axisbox}[1][]{%
  enhanced,
  breakable,
  colback=axisBodyGray,
  colframe=axisBorder,
  boxrule=0.8pt,
  arc=1mm,
  outer arc=1mm,
  left=2.6mm,
  right=2.6mm,
  top=2.0mm,
  bottom=2.0mm,
  before skip=0.8em,
  after skip=0.8em,
  title={#1},
  fonttitle=\bfseries\color{white},
  coltitle=white,
  colbacktitle=axisHeaderGray,
  toptitle=0.7mm,
  bottomtitle=0.7mm,
  lefttitle=2.6mm,
  righttitle=2.6mm,
  boxed title style={
    colback=axisHeaderGray,
    colframe=axisBorder,
    boxrule=0.8pt,
    arc=1mm,
    outer arc=1mm
  }
}

\def\Snospace~{Section }

\newcommand{\iconsize}{1em}
\newcommand{\icorutgers}{\raisebox{-2pt}{\includegraphics[height=1.2\iconsize]{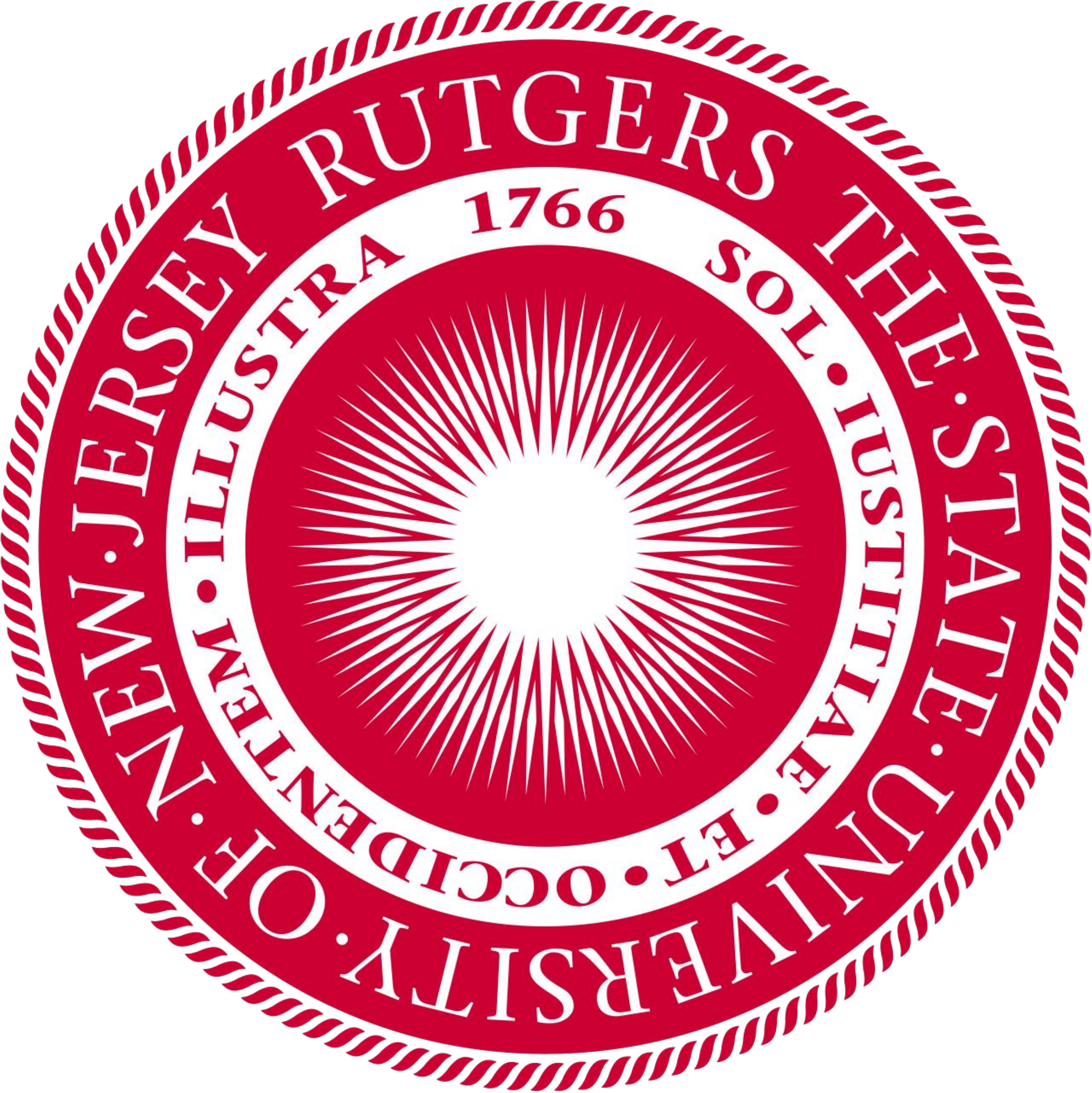}}}
\newcommand{\iconotredame}{\raisebox{-2pt}{\includegraphics[height=1.2\iconsize]{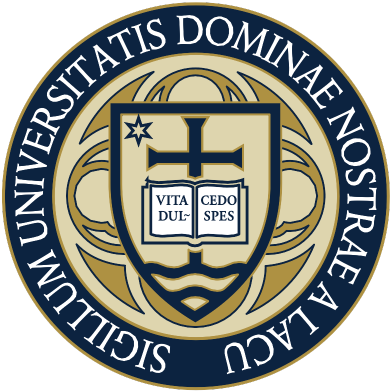}}}
\newcommand{\icoprinceton}{\raisebox{-2pt}{\includegraphics[height=1.25\iconsize]{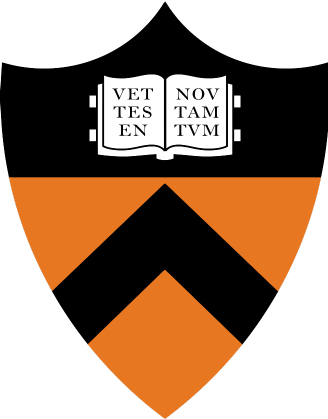}}}
\newcommand{\icoumn}{\raisebox{-2pt}{\includegraphics[height=\iconsize]{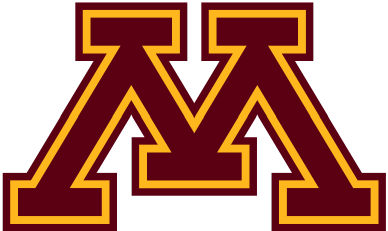}}}
\newcommand{\icoamd}{\raisebox{-0.75pt}{\includegraphics[height=0.82em]{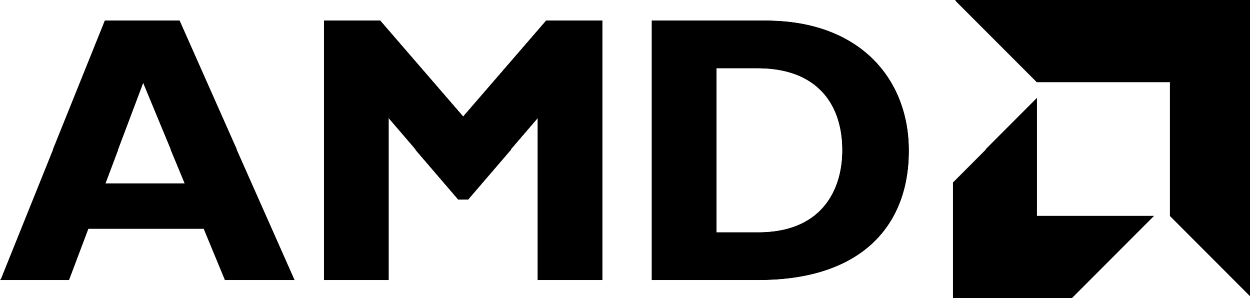}}}

\renewcommand{\titleboxlogo}{\includegraphics[height=2.5cm]{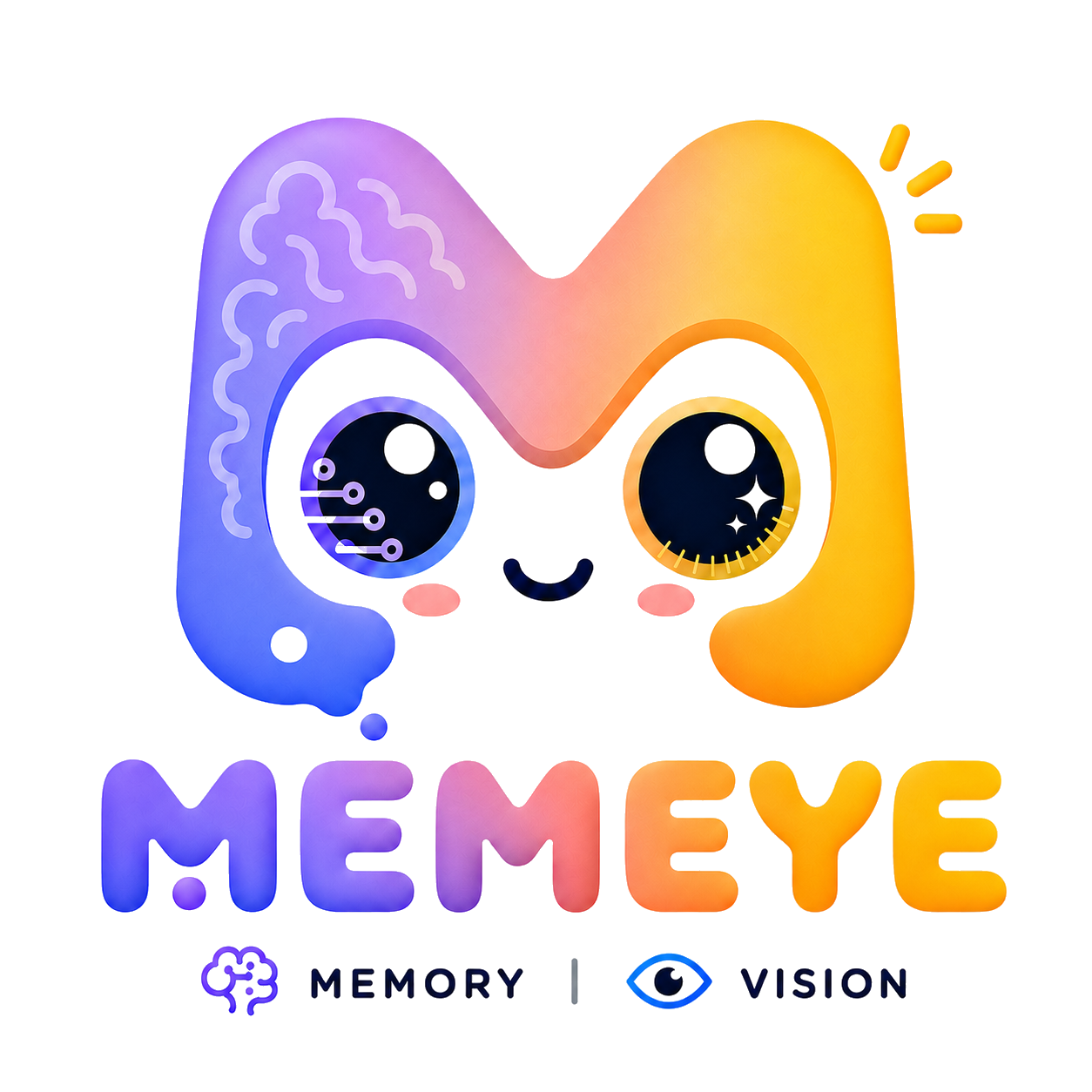}}
\newcommand{\memeyelogo}{\raisebox{-3pt}{\includegraphics[height=1.3em]{Fig/memeye-logo.png}}}

\title{MemEye: A Visual-Centric Evaluation Framework for Multimodal Agent Memory}

\runningtitle{\memeyelogo{} MemEye: A Visual-Centric Evaluation Framework for Multimodal Agent Memory}

\author{
  Minghao Guo$^{1,*}$,
  Qingyue Jiao$^{2,*}$,
  Zeru Shi$^{1,*}$,
  Yihao Quan$^{1}$,
  Boxuan Zhang$^{1}$,
  Danrui Li$^{1}$,
  Liwei Che$^{1}$,
  Wujiang Xu$^{1}$,
  Shilong Liu$^{3}$,
  Zirui Liu$^{4}$,
  Mubbasir Kapadia$^{1}$,
  Vladimir Pavlovic$^{1}$,
  Jiang Liu$^{5}$,\protect\\
  Mengdi Wang$^{3}$,
  Yiyu Shi$^{2,\dagger}$,
  Dimitris N. Metaxas$^{1,\dagger}$, and
  Ruixiang Tang$^{1,\dagger}$
  \protect\\[2mm]
  {\normalfont\small
  $^1$\icorutgers\,Rutgers \quad
  $^2$\iconotredame\,Notre Dame \quad
  $^3$\icoprinceton\,Princeton \quad
  $^4$\icoumn\,UMN \quad
  $^5$\icoamd}
}

\date{}

\newcommand{\githublogo}{\raisebox{-1.5pt}{\includegraphics[height=1.05em]{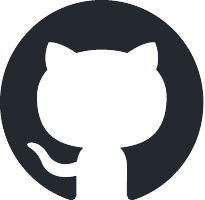}}}
\newcommand{\hflogo}{\raisebox{-1.5pt}{\includegraphics[height=1.05em]{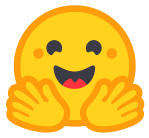}}}

\begin{document}

\begin{abstract}

Long-term agent memory is increasingly multimodal, yet existing evaluations rarely test whether agents preserve the visual evidence needed for later reasoning.
In prior work, many visually grounded questions can be answered using only captions or textual traces, allowing answers to be inferred without preserving the fine-grained visual evidence. Meanwhile, harder cases that require reasoning over changing visual states are largely absent.
Therefore, we introduce \textbf{\sys}, a framework that evaluates memory capabilities from two dimensions: one measures the granularity of decisive visual evidence (from scene-level to pixel-level evidence), and the other measures how retrieved evidence must be used (from single evidence to evolutionary synthesis). Under this framework, we construct a new benchmark across 8 life-scenario tasks, with ablation-driven validation gates for assessing answerability, shortcut resistance, visual necessity, and reasoning structure. By evaluating 13 memory methods across 4 VLM backbones, we show that current architectures still struggle to \textit{preserve fine-grained visual details} and \textit{reason about state changes over time}. Our findings show that long-term multimodal memory depends on evidence routing, temporal tracking, and detail extraction.


\vspace{2mm}
{\centering
\coloremojicode{1F310}\,\href{https://minghokwok.github.io/MemEye/}{\textbf{Project Page}} \quad
\githublogo\,\href{https://github.com/MinghoKwok/MemEye}{\textbf{Code}} \quad
\hflogo\,\href{https://huggingface.co/datasets/MemEyeBench/MemEye}{\textbf{Dataset}}\par}

\end{abstract}

\maketitle
\correspondingauthor{$^{*}$\textit{Equal contribution} \quad $^{\dagger}$\textit{Corresponding authors}}

\section{Introduction}

Long-term memory has recently become a major focus in building mordern intelligent agents~\cite{zhang2025survey, chhikara2025mem0}. However, the rapid development of Vision-Language Models (VLMs)~\cite{bai2025qwen3, xie2024large} has changed the way agents interact, allowing them to process both textual and visual inputs. In multimodal conversations, an agent must remember not only the dialogue history but also the visual information shown across different sessions. Human visual memory can retain rich object details and track changes in scenes over time~\cite{brady2008visual}, yet existing evaluations provide limited evidence about whether VLM-based agents can preserve and reason over visual information in long-term interactions~\cite{pang2026steering, 10.1145/3746027.3755646}. 

\begin{wrapfigure}{r}{0.7\textwidth}
  \centering
  \vspace{-12pt}
  \includegraphics[width=\linewidth]{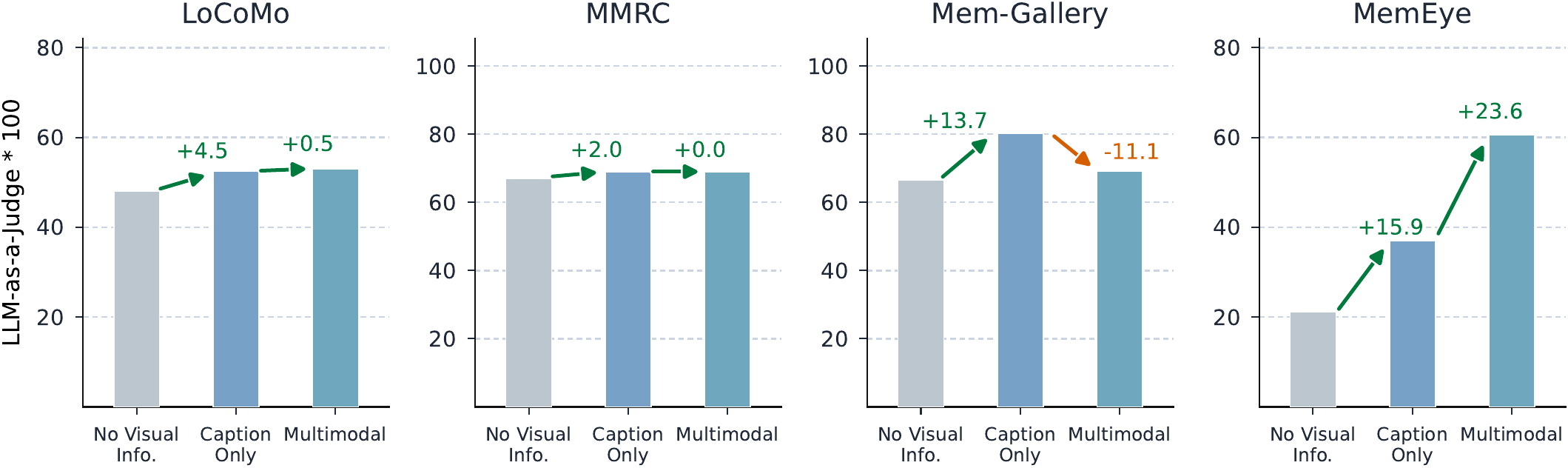}
  \vspace{-8pt}
  \caption{
Measuring visual information necessity across long-term memory benchmarks. We compare three settings with the same textual question: \textit{No Visual Info.} removes images, \textit{Caption Only} replaces images with captions, and \textit{Multimodal} provides the original images. MemEye exhibits stronger visual irreplaceability than LoCoMo~\citep{maharana2024evaluating}, MMRC~\citep{xue2025mmrc}, and Mem-Gallery~\citep{bei2026mem}. Details are in Appendix~\ref{app:effective-visual-information}.}
  \label{fig:caption-irreplaceability-bar}
  \vspace{-10pt}
\end{wrapfigure}

Most existing benchmarks~\citep{maharana2024evaluating,wu2024longmemeval,hu2025evaluating,guo2026individualturingtestcase,das2017visual,kottur2019clevr,liu2024mmdu,xue2025mmrc,xu2025crab} either focus on short-context image understanding or evaluate long-term memory settings where the primary information is textual.
While recent efforts~\cite{bei2026mem} distribute images across multiple sessions, many visually grounded questions remain answerable from captions, surrounding dialogue, or answer options rather than retained visual evidence.
Figure~\ref{fig:caption-irreplaceability-bar} shows that prior long-term memory benchmarks such as LoCoMo~\citep{maharana2024evaluating}, MMRC~\citep{xue2025mmrc}, and Mem-Gallery~\citep{bei2026mem} have smaller caption-to-multimodal gains, suggesting weaker dependence on original images.
Moreover, state changes are often described in text rather than through evolving visual evidence, making it difficult to test whether agents can track visual updates over time.

\begin{figure}[t]
    \centering
    \vspace{-8pt}
    \includegraphics[width=\linewidth]{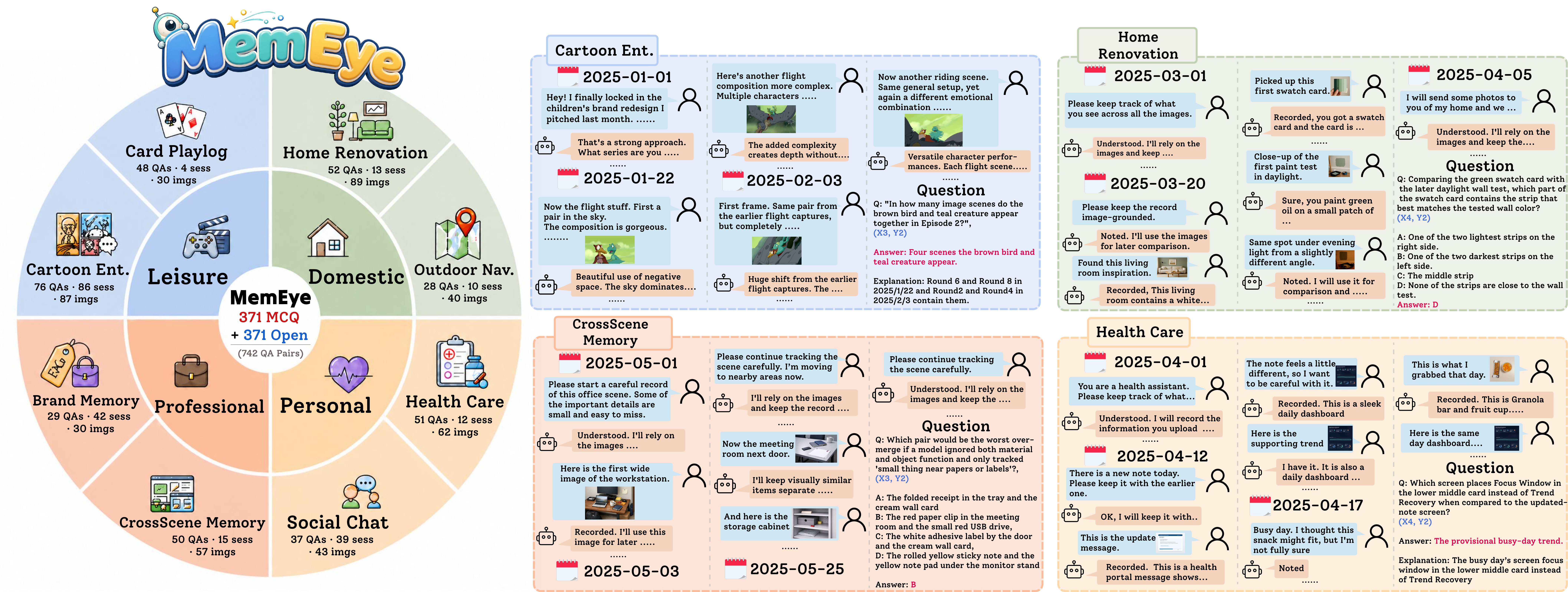}
    \caption{The left sub-figure shows the \sys dataset overview, with inner rings grouping tasks and outer rings showing statistics. The right sub-figure presents example cases.}
    \label{fig:teaser}
    \vspace{-6pt}
  \end{figure}

These limitations leave two core challenges unaddressed. First, agents often fail to preserve visual evidence at the necessary level of detail. While a text caption can capture the general scene information, it frequently loses specific attributes, such as region layouts, object identities, and fine-grained textures. These details can be easily lost when images are compressed into text. Second, agents struggle to reason over their history beyond simple retrieval. This involves linking evidence across sessions and synthesizing the current state as new observations override previous ones. Without separating these two factors, current evaluations make it difficult to track the system failure causes.


To bridge this gap, we propose \sys, a framework that evaluates multimodal agent memory along two orthogonal dimensions. 
The first dimension, \textbf{\textit{visual evidence granularity}}, ranges from scene-level evidence to pixel-level evidence and measures whether memory systems can preserve the visual details needed to answer a question.
The second dimension, \textbf{\textit{memory reasoning depth}}, ranges from atomic retrieval to evolutionary synthesis and measures whether memory systems can reason over retrieved evidence to answer a question.
Based on this framework, we construct a benchmark with 371 mirrored multiple-choice and open-ended questions across eight life-scenario tasks, as shown in Figure~\ref{fig:teaser}. 
Each question is annotated with its visual-evidence granularity and memory-reasoning depth, and filtered to reduce textual shortcut cases. As shown in Figure~\ref{fig:caption-irreplaceability-bar}, MemEye shows a larger caption-to-multimodal gain than prior long-term memory benchmarks, indicating stronger dependence on original visual evidence.

Using this benchmark, we evaluate \(13\) memory methods across four vision-language model backbones.
Current systems remain far from reliable long-term visual memory.
We identify a trade-off: text-based memory can help organize state transitions and updates, but often loses fine-grained visual details during abstraction.
In contrast, native image memory preserves visual evidence more directly, but struggles to identify which visual state remains valid over time.
Furthermore, cross-topic scaling shows that specialized memory mechanisms become more important as history length and thematic diversity grow.
Together, these findings suggest that effective multimodal memory must preserve visual details, track temporal validity, and select the right evidence over long histories.

\begin{table}[t]
\centering
\fontsize{6.1pt}{7pt}\selectfont
\newcommand{\cmark}{\textcolor{green!50!black}{\ding{51}}}
\newcommand{\pmark}{\textcolor{orange!85!black}{\ding{108}}}
\newcommand{\xmark}{\textcolor{red!70!black}{\ding{55}}}
\caption{Comparison with multimodal conversational benchmarks.}
\label{tab:image-conv-benchmark-comparison}
\vspace{-3pt}
\setlength{\tabcolsep}{2pt}
\renewcommand{\arraystretch}{0.92}
\begin{tabular*}{\columnwidth}{@{\extracolsep{\fill}}l*{8}{c}c@{}}
\toprule
& VisDial~\citep{das2017visual}
& CLEVR-dialog~\citep{kottur2019clevr}
& LoCoMo~\cite{maharana2024evaluating}
& MMDU~\cite{liu2024mmdu}
& ConvBench~\citep{liu2024convbench}
& MMRC~\cite{xue2025mmrc}
& MultiVerse~\citep{lee2025multiverse}
& Mem-Gallery~\citep{bei2026mem}
& \cellcolor{orange!10}\textbf{MemEye} \\
\midrule
Caption-Proof  & \xmark & \xmark & \xmark & \xmark & \xmark & \xmark & \xmark & \xmark & \cellcolor{orange!10}\textbf{\cmark} \\
Long Memory    & \xmark & \xmark & \cmark & \xmark & \xmark & \pmark & \xmark & \cmark & \cellcolor{orange!10}\textbf{\cmark} \\
Fine Visual    & \cmark & \cmark & \xmark & \cmark & \cmark & \xmark & \cmark & \pmark & \cellcolor{orange!10}\textbf{\cmark} \\
Cross-Context  & \cmark & \cmark & \cmark & \pmark & \cmark & \cmark & \cmark & \cmark & \cellcolor{orange!10}\textbf{\cmark} \\
State Revision & \xmark & \xmark & \pmark & \xmark & \xmark & \cmark & \xmark & \pmark & \cellcolor{orange!10}\textbf{\cmark} \\
Visual-State   & \xmark & \xmark & \xmark & \xmark & \xmark & \xmark & \xmark & \xmark & \cellcolor{orange!10}\textbf{\cmark} \\
\bottomrule
\end{tabular*}
\vspace{1pt}

\begin{minipage}{\columnwidth}
\fontsize{6pt}{7pt}\selectfont
\raggedright
* \cmark\,= substantial ($\geq$20\%), \pmark\,= limited (5--20\%), \xmark\,= negligible ($<$5\%) coverage under MemEye labeling.
\textbf{Caption-Proof}: verifies native images cannot be replaced by captions.
\textbf{Long Memory}: requires extended conversational memory.
\textbf{Fine Visual}: $X_3$--$X_4$ visual bottlenecks (instance binding, pixel attributes, OCR).
\textbf{Cross-Context}: $Y_2$ associations across turns, sessions, or modalities.
\textbf{State Revision}: $Y_3$ reasoning under updates, conflicts, or overrides.
\textbf{Visual-State}: joint high-visual, evolving-memory region.
\end{minipage}
\vspace{-10pt}
\end{table}

In summary, our main contributions are:

\begin{itemize}
    \item \textbf{A Multi-dimensional Evaluation Framework.} We propose \sys, a novel framework that categorizes multimodal memory challenges along two orthogonal axes: visual evidence granularity (ranging from scene-level to pixel-level) and memory reasoning depth (ranging from atomic retrieval to evolutionary synthesis).
    \item \textbf{A Vision-Centric Long-Term Memory Benchmark.} We introduce a rigorous benchmark with $371$ mirrored questions across real-world scenarios to test whether multimodal agents can preserve and reason over irreplaceable visual evidence.
    \item \textbf{Comprehensive Evaluation and Empirical Insights.} We comprehensively evaluate existing methods and reveal a key trade-off: text-based memory helps manage state changes but loses fine-grained visual details, while image-based memory preserves visual evidence but struggles with temporal validity.
\end{itemize}






\section{Related Work}

\subsection{Agent Memory Systems}

Long-term memory management is a central design problem for deployed agents~\cite{zhang2025survey, hu2025}. Prior work has explored memory mechanisms for computer-use and interactive agents~\cite{shi2024commands, mei2025r, guo-etal-2026-deepsieve,chen2025harmonyguardsafetyutilityweb,chen2026safepredpredictiveguardrailcomputerusing,li2026avenir}, including textual memory systems with explicit memory writing, updating, and maintenance procedures~\cite{xu2025amem, kang2025memoryos, chhikara2025mem0,xu2026ael}. These methods improve an agent's ability to store and reuse past information, but they primarily operate over textual memories or text abstractions of prior experience.

Recent multimodal memory methods extend this line of work by retaining or retrieving visual experience, including MIRIX~\cite{wang2025mirix}, MMA~\cite{lu2026mma}, M2A~\cite{feng2026m2a}, and FluxMem~\cite{xie2026fluxmem}. These systems make different architectural trade-offs among coverage, retrieval selectivity, abstraction, and revision. However, existing evaluations often report end-task performance without isolating which memory operation fails. A memory system may discard perceptual details during captioning or memory writing, retrieve a semantically relevant but temporally invalid clue, or fail to synthesize the valid state even when relevant evidence is available. MemEye is designed to evaluate these mechanisms rather than only compare aggregate accuracy, exposing when a memory architecture loses visual evidence, selects stale evidence, or fails to recover the valid memory state.

\subsection{Memory Benchmarks for Long-Horizon Multimodal Agents}

Long-horizon agent benchmarks increasingly evaluate whether systems can retain information across extended interactions. Text-centric benchmarks such as LoCoMo~\cite{maharana2024evaluating}, LongMemEval~\cite{wu2024longmemeval}, TwinVoice~\cite{du2025twinvoicemultidimensionalbenchmarkdigital}, and MemoryAgentBench~\cite{hu2025evaluating} primarily measure whether linguistic facts can be recovered, summarized, or used after many turns. Multimodal benchmarks such as MMDU~\cite{liu2024mmdu}, ATM-Bench~\cite{mei2026atm}, and MMRC~\cite{xue2025mmrc} introduce image information within dialogue, while Mem-Gallery~\cite{bei2026mem} further extends this direction to a multi-session multimodal memory setting where images appear throughout the conversation.

The missing object of study is not only another task domain, but the coupled failure mode between visual evidence compression and state-evolving memory use. As shown in~\autoref{tab:image-conv-benchmark-comparison}, prior benchmarks rarely ask whether the decisive image content can be bypassed by captions, whether fine-grained visual evidence must be preserved at instance or pixel granularity, or whether visual evidence changes over time. For example, although Mem-Gallery introduces knowledge conflicts, these conflicts are primarily textual rather than visual-state updates. MemEye therefore treats visual evidence as the central memory bottleneck: each item specifies the visual granularity that must be retained and how it must be used over time.




\section{MemEye Framework and Benchmark}
\label{sec:benchmark}

\subsection{The Two-Dimensional Evaluation Framework}

\begin{figure}[t]
  \centering
  \includegraphics[width=\linewidth]{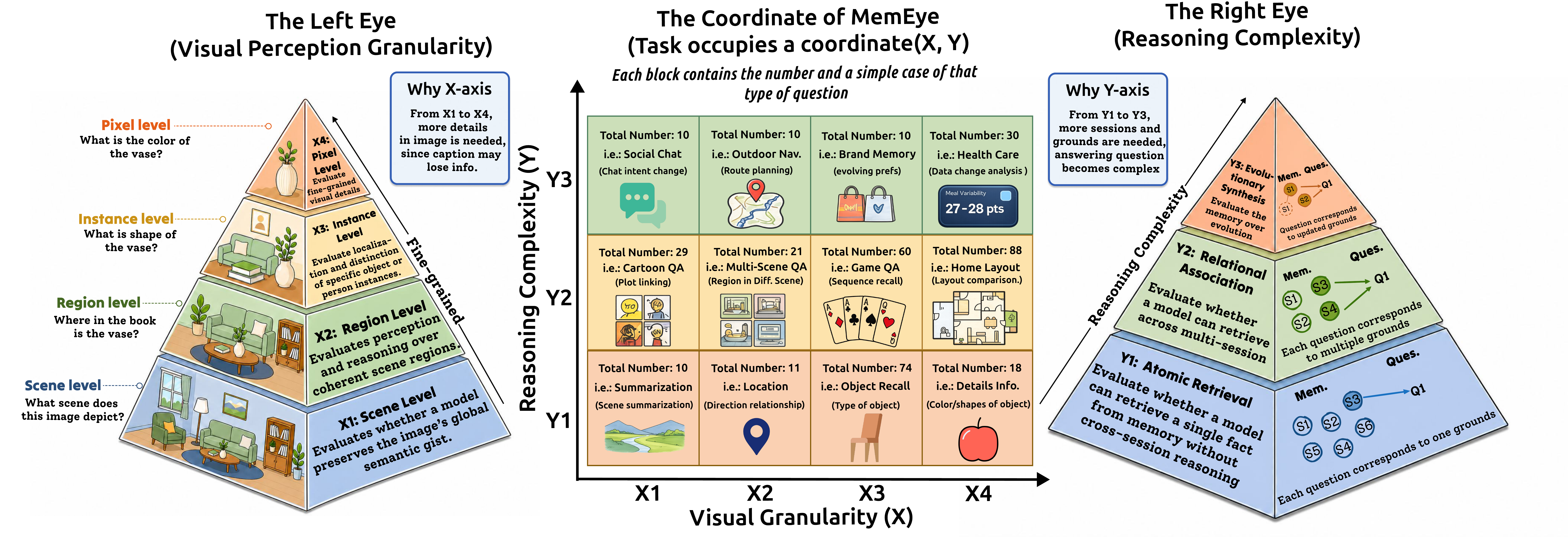}
  \caption{The MemEye two-axis taxonomy. The X-axis captures the granularity of decisive visual evidence, while the Y-axis captures the required reasoning operation over memory.}
  \label{fig:xyaxis}
  \vspace{-10pt}
\end{figure}

\sys's evaluation framework is organized as follows. As shown in Figure~\ref{fig:xyaxis}, \sys contains two dimensions that form a coordinate system. The X-axis represents the first dimension, defined by the granularity of visual perception. From $X_1$ to $X_4$, the granularity of the required visual evidence becomes increasingly fine-grained. The definitions of $X_1$ to $X_4$ are as follows:

\begin{axisbox}[Visual Evidence Granularity]
\begin{itemize}[label={}, leftmargin=0pt, itemsep=0pt, parsep=0pt, topsep=0pt]
\item \textbf{Scene-level.}
$X_1$ measures whether the model preserves scene-level evidence, including scene type, activity, and global semantic gist. This level captures the coarsest form of visual evidence and serves as the baseline for visual granularity.

\item \textbf{Region-level.}
$X_2$ evaluates the ability to perceive and reason over semantically coherent regions. The model must identify meaningful subregions and capture their local context and interactions. Unlike scene-level understanding, this level focuses on localized semantics, where information is organized within regions rather than the entire scene. 

\item \textbf{Instance-level.}
$X_3$ evaluates whether the model can localize and distinguish specific object or person instances within and across images. The key challenge is preserving entity identity when multiple similar candidates exist. Caption-based representations often flatten such distinctions, marking the transition from region-level understanding to instance-level visual memory.

\item \textbf{Pixel-level.}
$X_4$ requires reasoning over pixel-level evidence, including fine-grained details such as color, texture, or small text. These cues are often absent from text, reflecting pixel-level necessity where critical evidence exists only in the visual signal.
\end{itemize}
\end{axisbox}

The Y-axis corresponds to the second dimension, capturing the reasoning depth required for memory retrieval during question answering. It emphasizes not only whether sufficient evidence can be located, but also whether that evidence can be associated, revised, and synthesized into the valid answer. This dimension is organized into three levels, from $Y_1$ to $Y_3$, as detailed below:

\begin{axisbox}[Memory-Reasoning Depth]
\begin{itemize}[label={}, leftmargin=0pt, itemsep=0pt, parsep=0pt, topsep=0pt]
\item \textbf{Atomic Retrieval.}
$Y_1$ measures whether the model can retrieve a single fact from memory without cross-session reasoning. It primarily tests basic memory access rather than composition, serving as the lowest reasoning baseline.

\item \textbf{Relational Association.}
$Y_2$ evaluates the ability to associate distributed evidence across sessions and modalities. The reasoning remains monotonic: information accumulates without contradiction. This level captures referential resolution and implicit memory traversal beyond isolated retrieval.

\item \textbf{Evolutionary Synthesis.}
$Y_3$ tests non-monotonic synthesis over evolving memory. The model must handle updates, conflicts, and overrides, maintaining a coherent world state under revision. Answers are not explicitly stated but must be inferred from changing evidence.
\end{itemize}
\end{axisbox}

The two dimensions form \sys's framework. Each question is assigned an $(X,Y)$ coordinate indicating its level of visual evidence and depth of reasoning over memory. The middle sub-figure of~\autoref{fig:xyaxis} illustrates this assignment.


The benchmark contains 371 questions across 221 sessions, 848 dialogue rounds, and 438 images. Each question has two mirrored forms (a multiple-choice version and an open-ended version). For MCQ questions, to mitigate VLM bias, we create four rotated variants with the correct answer cycling through A–D. As shown in Figure~\ref{fig:teaser}, the benchmark spans eight tasks grouped into four life-scenario domains: \emph{Leisure} (Card Playlog and Cartoon Entertainment), \emph{Domestic} (Home Renovation and Outdoor Navigation), \emph{Professional} (Brand Memory and CrossScene Memory), and \emph{Personal} (Health Care and Social Chat). The images come from both public and archival media, as well as generated content, covering a wide range of image types, including photographs, screenshots, comic panels, and user interface renderings. Each question receives the most demanding $X$ and $Y$ labels needed to answer it; the full $(X,Y)$ cell distribution is provided in Appendix~\ref{app:cell-distribution}. After generating the candidate visual-memory questions and assigning the label to each question, we use three mechanisms to verify that our benchmark is visual-centric and that these questions arise from limitations in the agent’s memory capabilities, rather than from the underlying foundation model.

\begin{wrapfigure}[9]{r}{0.56\textwidth}
    \vspace{-10pt}
    \centering
    \includegraphics[height=0.22\textheight,width=0.54\textwidth,keepaspectratio]{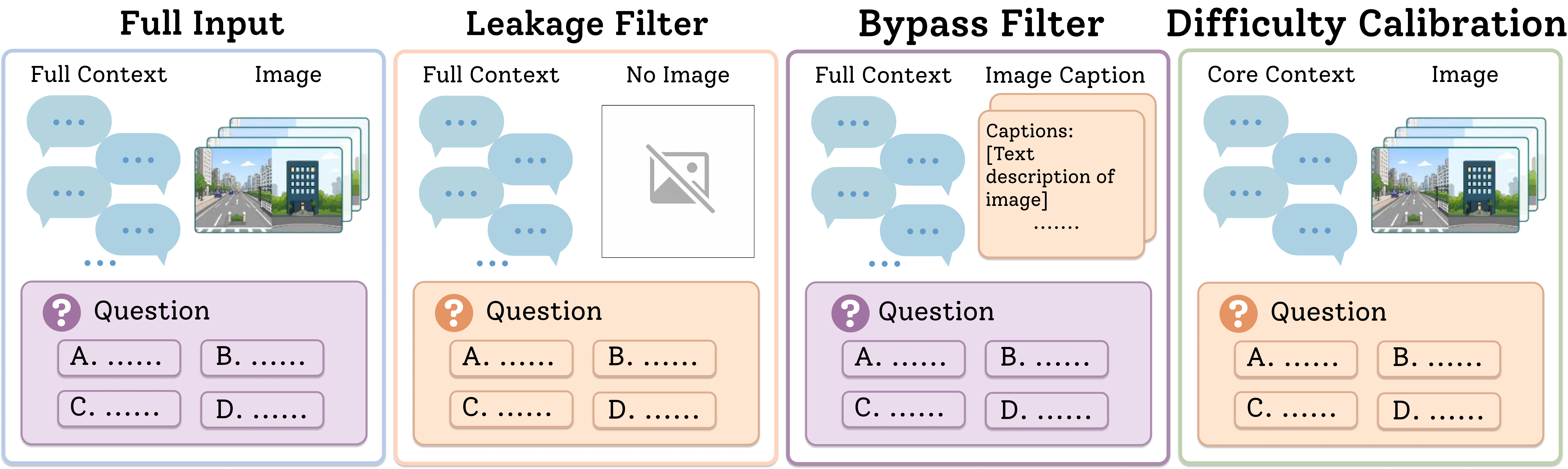}
    \caption{The filtering process used to build the benchmark.}
    \label{fig:datafilter}
    \vspace{-14pt}
\end{wrapfigure}

To be more specific, we perform three filtering mechanisms on each question. To mitigate VLM bias, we use four answer rotations during these checks. \textbf{(1) Eliminating answer leakage in dialogue.} For each question, we provide only the question, answer choices, and gold clue-round text, with no images or captions, and test whether the agent can answer correctly across answer rotations. If so, the item is considered solvable without visual evidence and is removed. \textbf{(2) Eliminating visual bypassability via minimal captions.} We replace each image with a very short caption and test whether the question can still be answered. The caption only keeps the rough image type, such as a room photo, a game board, or a phone screenshot. If a candidate can still be answered from these captions, we revise or remove it because its visual evidence is too easily replaced by text and therefore does not satisfy the visual-centric requirement. \textbf{(3) Controlling for problem difficulty.} We provide the image along with the answer-relevant context to assess whether the question is inherently solvable. This setting does not evaluate memory; rather, it isolates answerability. If the model fails, the difficulty is attributed to limitations of the underlying foundation model rather than its memory. Through these mechanisms, we retain questions that require visual information and are suitable for evaluating memory capabilities rather than only foundation-model recognition ability. More details are provided in Appendix~\ref{app:item-filtering-details}.

\section{Experiments and Analysis}

In this section, we use our benchmark to analyze current multi-modal agent memory systems. Our analysis moves from locating failures to explaining their causes. We first validate the rationality of our framework's configuration using our benchmark. Then we ask three questions:
\textbf{RQ1}: Where do current memory systems fail in the MemEye matrix?
\textbf{RQ2}: Why do memory systems lose visual information?
\textbf{RQ3}: Why do memory systems lose evolving visual states?
Together, these questions first map the failure landscape, then isolate the visual-evidence bottleneck in high-$X$ questions, and finally diagnose why retrieval remains insufficient when memory evidence evolves over time.

\begin{figure*}[htbp]
  \centering
  \includegraphics[width=\textwidth]{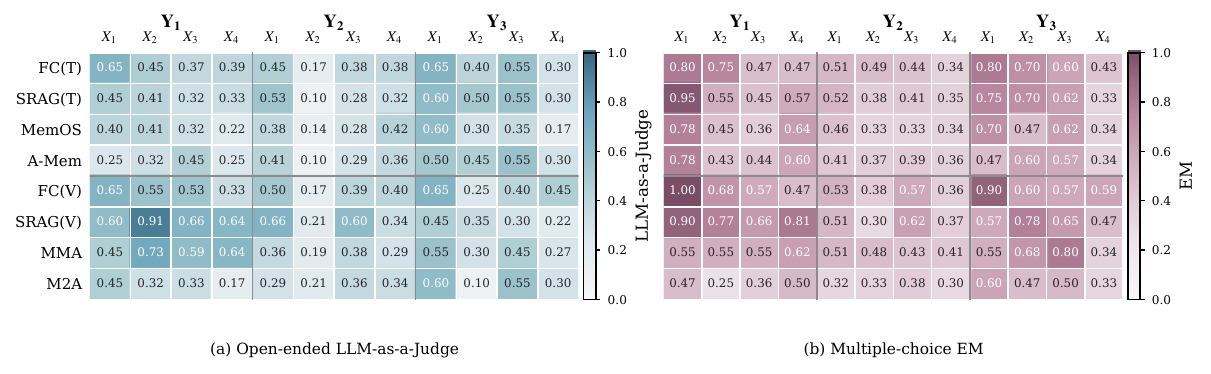}
  \caption{Representative method performance heatmap using \texttt{gpt-5.4-mini}. Left: LLM-as-a-Judge; Right: MCQ EM. }
  \label{fig:cell_level_representative_heatmap}
\end{figure*}

\subsection{Evaluation Setup}

\paragraph{Models and memory methods.}
We evaluate 13 methods across 4 model backbones including Qwen3-VL-8B-Instruct~\citep{bai2025qwen3}, GPT-4.1-nano, GPT-5.4-mini~\citep{openai_models_2026}, and Gemini-2.5-flash-lite~\citep{google_gemini_models_2026}. 

The evaluated methods include seven text-based memory approaches and six multimodal memory approaches. The text-based methods are Full Context (FC(T)), Semantic RAG (SRAG(T)), Reflexion (Refl.)~\citep{shinn2023reflexion}, Generative Agents (Gen.Ag.)~\citep{park2023generativ}, MemoryOS (MemOS)~\citep{kang2025memoryos}, A-Mem~\citep{xu2025amem}, and SimpleMem (SM(T))~\citep{simplemem2025}. These methods replace each image with a dense GPT-5.2 caption. The multimodal methods are Full Context (FC(V)), Semantic RAG (SRAG(V)), MIRIX~\citep{wang2025mirix}, MMA~\citep{lu2026mma}, M2A~\citep{feng2026m2a}, and SimpleMem (SM(V))~\citep{omnisimplemem2026}. These methods operate on the original visual inputs. For retrieval-based methods, we use top-$K{=}10$ and standardize text and image embedding backbones where possible. The full model identifiers, embedding settings, context budgets, and implementation details are provided in Appendix~\ref{app:method-implementation-details} and~\ref{app:evaluation-implementation-details}. We follow each method's official or recommended retrieval stack when available, so method comparisons should be interpreted as system-level comparisons rather than encoder-controlled ablations.

\paragraph{Metrics and diagnostics.}
For multiple-choice evaluation, we report exact-match accuracy (EM) averaged over the four answer rotations. For open-ended evaluation, we use LLM-as-a-Judge as the primary metric. We report BLEU-1 as an auxiliary lexical metric in Appendix~\ref{app:full-main-matrix}. To validate the judge, we conduct a human-judge agreement study on a stratified sample of 72 predictions. The automated accept/reject judgments show strong agreement with human labels, with Cohen's $\kappa = 0.94$. Details are provided in Appendix~\ref{app:judge-agreement}.


\begin{table*}[!t]
  \centering
  \caption{
  Main results on the MemEye evaluation matrix using \texttt{gpt-5.4-mini}. Columns correspond to memory methods, grouped into text-only and multimodal families. Within each coordinate, EM is reported for multiple-choice questions, while LLM-as-a-Judge (LLM-Judge) is reported for free-response questions. The first- and second-performing memory model(s) are highlighted with orange and blue backgrounds, respectively. Results on other backbones are shown in Appendix~\ref{app:additional-results}.}
  \label{tab:memeye_main_matrix_gpt_5_4_mini_full}

  \vspace{-1mm}
  \scriptsize
  \setlength{\tabcolsep}{2.2pt}
  \renewcommand{\arraystretch}{0.88}

  \resizebox{\textwidth}{!}{%
   \begin{tabular}{lllcccccccccccccc}
  \toprule
  \multirow{2}{*}{Y} & \multirow{2}{*}{X} & \multirow{2}{*}{Metric}
  & \multicolumn{7}{c}{Textual memory}
  & \multicolumn{6}{c}{Multimodal memory} \\
  \cmidrule(lr){4-10}\cmidrule(lr){11-16}
  & &
  & FC(T) & SRAG(T) & Refl. & Gen.Ag. & MemOS & A-Mem & SM(T)
  & FC(V) & SRAG(V) & MIRIX & MMA & M2A & SM(V) \\
  \midrule

  \multirow{8}{*}{Y1}
  & \multirow{2}{*}{X1} & EM   & 0.8000 & \secondcell{0.9500} & 0.6750 & 0.2500 & 0.7750 & 0.7750 & 0.8000 & \bestcell{1.0000} & 0.9000 & 0.6750 & 0.5500 & 0.4750 & 0.8500 \\
  & & LLM-Judge   & \bestcell{0.6500} & 0.4500 & \secondcell{0.6000} & 0.3000 & 0.4000 & 0.2500 & 0.5500 & \bestcell{0.6500} & \secondcell{0.6000} & 0.4000 & 0.4500 & 0.4500 & 0.5000 \\
  \cmidrule(lr){2-16}
  & \multirow{2}{*}{X2} & EM   & \secondcell{0.7500} & 0.5455 & 0.2045 & 0.2273 & 0.4545 & 0.4318 & 0.5000 & 0.6818 & \bestcell{0.7727} & 0.5227 & 0.5455 & 0.2500 & 0.4773 \\
  & & LLM-Judge   & 0.4545 & 0.4091 & 0.1364 & 0.1818 & 0.4091 & 0.3182 & 0.2727 & 0.5455 & \bestcell{0.9091} & 0.5000 & \secondcell{0.7273} & 0.3182 & 0.1818 \\
  \cmidrule(lr){2-16}
  & \multirow{2}{*}{X3} & EM   & 0.4662 & 0.4527 & 0.2534 & 0.2500 & 0.3649 & 0.4392 & 0.3209 & \secondcell{0.5709} & \bestcell{0.6554} & 0.5473 & 0.5507 & 0.3615 & 0.3784 \\
  & & LLM-Judge   & 0.3716 & 0.3176 & 0.3108 & 0.2230 & 0.3176 & 0.4459 & 0.2230 & 0.5338 & \bestcell{0.6554} & 0.2568 & \secondcell{0.5946} & 0.3311 & 0.2838 \\
  \cmidrule(lr){2-16}
  & \multirow{2}{*}{X4} & EM   & 0.4722 & 0.5694 & 0.4444 & 0.2361 & \secondcell{0.6389} & 0.5972 & 0.4583 & 0.4722 & \bestcell{0.8056} & 0.3750 & 0.6250 & 0.5000 & 0.5000 \\
  & & LLM-Judge   & \secondcell{0.3889} & 0.3333 & 0.1111 & 0.0556 & 0.2222 & 0.2500 & 0.2222 & 0.3333 & \bestcell{0.6389} & 0.3056 & \bestcell{0.6389} & 0.1667 & 0.2222 \\
  \midrule

  \multirow{8}{*}{Y2}
  & \multirow{2}{*}{X1} & EM   & 0.5086 & 0.5172 & 0.4483 & 0.2500 & 0.4569 & 0.4052 & 0.5000 & \secondcell{0.5345} & 0.5086 & 0.2845 & 0.5086 & 0.3190 & \bestcell{0.5690} \\
  & & LLM-Judge   & 0.4483 & 0.5345 & \secondcell{0.5517} & 0.2759 & 0.3793 & 0.4138 & 0.4828 & 0.5000 & \bestcell{0.6552} & 0.4138 & 0.3621 & 0.2931 & 0.4138 \\
  \cmidrule(lr){2-16}
  & \multirow{2}{*}{X2} & EM   & \bestcell{0.4881} & 0.3810 & 0.1905 & 0.2619 & 0.3333 & 0.3690 & 0.3095 & 0.3810 & 0.2976 & 0.3214 & \secondcell{0.4762} & 0.3333 & 0.3452 \\
  & & LLM-Judge   & 0.1667 & 0.0952 & 0.1667 & 0.0238 & 0.1429 & 0.0952 & 0.1429 & 0.1667 & \bestcell{0.2143} & 0.1429 & \secondcell{0.1905} & \bestcell{0.2143} & 0.0952 \\
  \cmidrule(lr){2-16}
  & \multirow{2}{*}{X3} & EM   & 0.4417 & 0.4125 & 0.3833 & 0.2292 & 0.3333 & 0.3917 & 0.3917 & \secondcell{0.5750} & \bestcell{0.6250} & 0.4583 & 0.4292 & 0.3792 & 0.3708 \\
  & & LLM-Judge   & 0.3750 & 0.2833 & 0.2000 & 0.1167 & 0.2833 & 0.2917 & 0.2333 & \secondcell{0.3917} & \bestcell{0.6000} & 0.2667 & 0.3750 & 0.3583 & 0.2667 \\
  \cmidrule(lr){2-16}
  & \multirow{2}{*}{X4} & EM   & 0.3438 & 0.3523 & 0.2841 & 0.2472 & 0.3409 & 0.3551 & 0.3097 & 0.3636 & \secondcell{0.3722} & 0.3665 & \bestcell{0.4119} & 0.2955 & 0.2869 \\
  & & LLM-Judge   & 0.3807 & 0.3182 & 0.3466 & 0.2614 & \bestcell{0.4205} & 0.3636 & 0.2500 & \secondcell{0.3977} & 0.3352 & 0.3352 & 0.2898 & 0.3352 & 0.2159 \\
  \midrule

  \multirow{8}{*}{Y3}
  & \multirow{2}{*}{X1} & EM   & \secondcell{0.8000} & 0.7500 & 0.7000 & 0.2500 & 0.7000 & 0.4750 & 0.5000 & \bestcell{0.9000} & 0.5750 & 0.6750 & 0.5500 & 0.6000 & 0.5500 \\
  & & LLM-Judge   & \secondcell{0.6500} & 0.6000 & \bestcell{0.7000} & 0.4500 & 0.6000 & 0.5000 & 0.6000 & \secondcell{0.6500} & 0.4500 & 0.6000 & 0.5500 & 0.6000 & 0.6000 \\
  \cmidrule(lr){2-16}
  & \multirow{2}{*}{X2} & EM   & \secondcell{0.7000} & \secondcell{0.7000} & 0.5500 & 0.2500 & 0.4750 & 0.6000 & 0.6500 & 0.6000 & \bestcell{0.7750} & 0.4750 & 0.6750 & 0.4750 & \secondcell{0.7000} \\
  & & LLM-Judge   & 0.4000 & \bestcell{0.5000} & \secondcell{0.4500} & \secondcell{0.4500} & 0.3000 & \secondcell{0.4500} & 0.3000 & 0.2500 & 0.3500 & 0.2000 & 0.3000 & 0.1000 & 0.3000 \\
  \cmidrule(lr){2-16}
  & \multirow{2}{*}{X3} & EM   & 0.6000 & 0.6250 & 0.5750 & 0.2750 & 0.6250 & 0.5750 & 0.5250 & 0.5750 & \secondcell{0.6500} & \secondcell{0.6500} & \bestcell{0.8000} & 0.5000 & 0.6000 \\
  & & LLM-Judge   & \bestcell{0.5500} & \bestcell{0.5500} & \bestcell{0.5500} & 0.4000 & 0.3500 & \bestcell{0.5500} & 0.3500 & 0.4000 & 0.3000 & 0.4000 & \secondcell{0.4500} & \bestcell{0.5500} & 0.3500 \\
  \cmidrule(lr){2-16}
  & \multirow{2}{*}{X4} & EM   & 0.4333 & 0.3250 & 0.2833 & 0.2333 & 0.3417 & 0.3417 & 0.3167 & \bestcell{0.5917} & \secondcell{0.4750} & 0.2583 & 0.3417 & 0.3250 & 0.3083 \\
  & & LLM-Judge   & 0.3000 & 0.3000 & 0.2833 & \secondcell{0.3167} & 0.1667 & 0.3000 & 0.2000 & \bestcell{0.4500} & 0.2167 & 0.1667 & 0.2667 & 0.3000 & 0.2000 \\
  \midrule
  \multirow{2}{*}{Avg.} & \multirow{2}{*}{--} & EM   & 0.5670 & 0.5484 & 0.4160 & 0.2467 & 0.4866 & 0.4797 & 0.4651 & \secondcell{0.6038} & \bestcell{0.6177} & 0.4674 & 0.5386 & 0.4011 & 0.4947 \\
  & & LLM-Judge   & 0.4280 & 0.3909 & 0.3672 & 0.2546 & 0.3326 & 0.3524 & 0.3189 & \secondcell{0.4391} & \bestcell{0.4937} & 0.3323 & 0.4329 & 0.3347 & 0.3025 \\
  \bottomrule
  \end{tabular}%
  }
  \vspace{-2pt}
\end{table*}

\subsection{Validation of \sys}

Before reporting diagnostic findings, we verify that the MemEye axes discriminate as intended: X captures visual evidence granularity, and Y captures reasoning depth over memory.


\paragraph{Caption-Proof Diagnostic.}
To validate the \(X\)-axis, we compare native-image memory with dense-caption memory and measure \(\Delta = \mathrm{Acc}_{\mathrm{image}} - \mathrm{Acc}_{\mathrm{caption}}\).
During benchmark construction, all \(X_1\)--\(X_4\) items already pass a minimal-caption bypass filter, removing questions answerable from very short captions that only preserve coarse image type.
Here, GPT-5.2 dense captions serve as a stronger textual substitute for testing how much visual evidence is lost when images are stored as text.
If the \(X\)-axis captures visual granularity, the image-caption gap should be smaller for scene- and region-level evidence and larger for instance- and pixel-level evidence.
Detailed results are reported in Appendix~\ref{app:caption-proof-validation-details} and analyzed in \S\ref{sec:caption-proof-validation}.

\paragraph{Oracle-Evidence Diagnostic.}
To validate the $Y$-axis, we evaluate an oracle-evidence setting where each question is answered using its ground-truth rounds and original images, removing retrieval as the main bottleneck. Here, ``oracle'' means that the annotated gold clue rounds are provided directly, rather than retrieved by the memory system. The results are shown in Appendix Table~\ref{tab:oracle-diagnostics}. In this setting, GPT-5.4-mini shows a steady drop in LLM-as-a-Judge performance from $Y_1$ to $Y_3$ ($0.673 \rightarrow 0.601 \rightarrow 0.558$), indicating that the $Y$-axis captures reasoning depth beyond retrieval. System-level results are consistent: retrieval-based methods perform well in $Y_1$, while full-context or state-aware methods become more competitive in $Y_3$. Thus, the $Y$-axis reflects differences in memory usage, not just task difficulty.

\begin{figure*}[t]
  \centering
  \includegraphics[width=\textwidth]{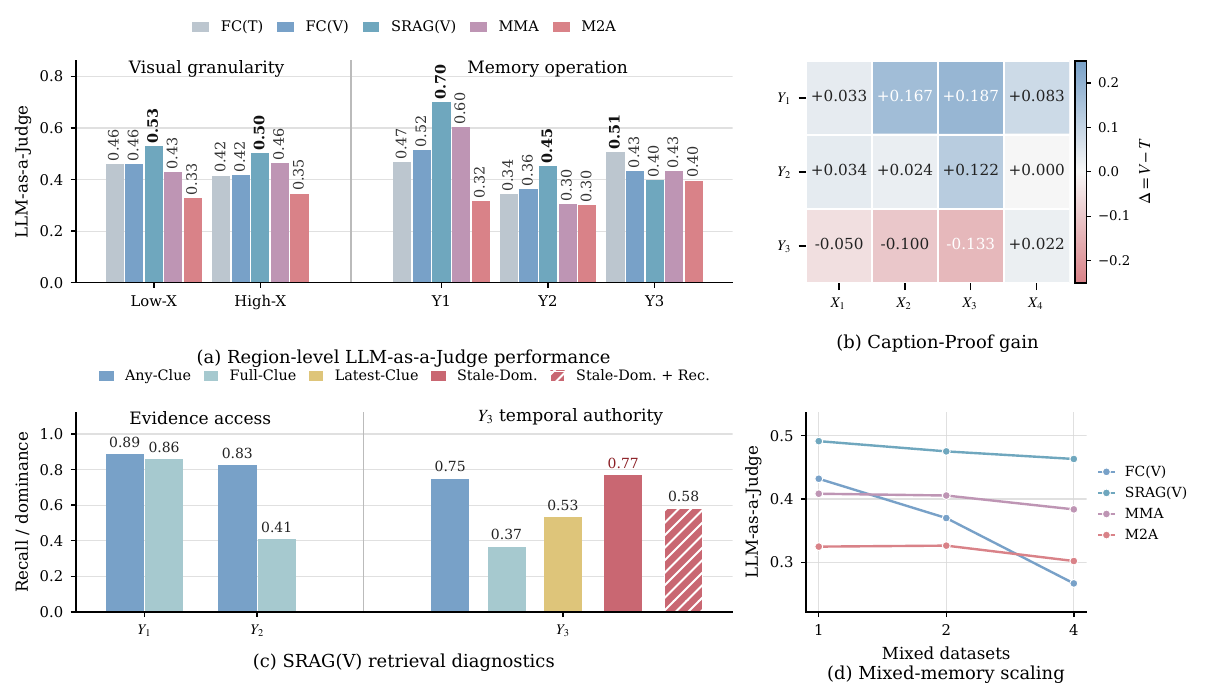}
  \caption{
  Experimental diagnostics on MemEye under \texttt{gpt-5.4-mini}. 
  (a) Region-level LLM-as-a-Judge performance. Bars report coordinate-balanced macro-averages: performance is first averaged within each relevant $(X,Y)$ coordinate, and the resulting coordinate scores are then averaged equally.
  (b) Average Caption-Proof gain, $\Delta=\mathrm{Score}_{V}-\mathrm{Score}_{T}$, shows where native visual evidence improves over dense captions.
  (c) SRAG(V) retrieval diagnostics separate evidence access from temporal-authority failures; the hatched bar shows a retrieval-only recency counterfactual.
  (d) Cross-topic dialogue scaling evaluates robustness as unrelated histories from other tasks are added. Results are averaged across two controlled four-task combinations. Full method-specific Caption-Proof heatmaps are reported in Appendix~\ref{app:caption-proof-validation-details}. Together, these panels show current systems trade off visual evidence preservation, evidence selection, and robustness, rather than solving all MemEye regions uniformly.
  }
  \label{fig:experimental-diagnostics}
  \vspace{-10pt}
\end{figure*}

\subsection{RQ1: Where Do Current Memory Systems Fail in the MemEye Matrix?}
\label{sec:main-results}

Table~\ref{tab:memeye_main_matrix_gpt_5_4_mini_full} reports cell-level performance using EM for multiple-choice questions and LLM-as-a-Judge for open-ended questions, while Figure~\ref{fig:cell_level_representative_heatmap} visualizes representative method performance as heatmaps. Current systems are far from saturating MemEye. At the aggregate level, SRAG(V) achieves the best open-ended performance with LLM-Judge $=0.4937$ and the best multiple-choice performance with EM $=0.6177$. The gap between EM and LLM-as-a-Judge is informative: multiple-choice accuracy can benefit from answer options and broad context coverage, whereas open-ended evaluation reveals whether the system can articulate the relevant memory state.

The results reveal two interacting stressors rather than a single memory challenge. First, fine-grained visual evidence exposes failures that are not visible at the scene level. At low $X$, caption-based memory remains competitive; at high $X$, native visual memory becomes more important. For example, at $(X_3,Y_1)$, SRAG(V) reaches an LLM-as-a-Judge score of $0.6554$, outperforming the best text-based method, A-Mem, which reaches $0.4459$. At $(X_4,Y_1)$, MMA and SRAG(V) both reach an LLM-as-a-Judge score of $0.6389$, while the best text-based method reaches $0.3889$ (Appendix~\ref{app:full-main-matrix}).

Second, evolving-state reasoning changes the bottleneck after evidence is retrieved. Retrieval works well when the relevant evidence can be selected directly: SRAG(V) remains competitive at $(X_2,Y_2)$ and in high-$X$ relational cells. In $Y_3$ cells, however, the system must decide which evidence remains valid after updates or conflicts. This shifts the bottleneck from evidence access to state selection. Therefore, retrieval-oriented methods lose some of their advantage, and methods with abstraction or revision mechanisms, such as M2A, Reflexion, and MemOS, perform better in lower-$Y$ cells. Still, no method solves both axes at once: textual or agentic memory can help organize evolving states but may lose fine visual details, whereas image-based memory preserves more visual evidence but struggles to select the updated visual state. This motivates RQ2 and RQ3, which separately analyze visual-evidence loss and state-selection failure.

\subsection{RQ2:  Why Do Memory Systems Lose Visual Information?}
\label{sec:caption-proof-validation}

We next analyze why fine-grained visual evidence is often lost. Current multimodal agent systems adopt two main strategies for storing images. Methods such as MIRIX and SimpleMem convert images into text abstractions to store and index such evidence with text embedding. In contrast, methods like MMA and M2A retain access to native image evidence and index them by image embeddings. To enable text-based memory systems to receive the image input, we replace each image with a dense caption. To compare these storage schemes, we focus on $Y_1$, where each question corresponds to a single evidence source and does not require multi-hop reasoning. This setting isolates the agent’s ability to understand and preserve visual information. 
Text-based storage methods perform as well as image-based methods on coarse-grained questions (e.g., $X_1$ and $X_2$), while image-based methods excel on fine-grained questions (e.g., $X_3$ and $X_4$). We attribute this difference to the nature of the two representations: text can capture high-level, generalized descriptions, whereas native images can better preserve fine-grained visual details.

To quantify this effect, we compare each text-based method with its visual counterpart and compute the Caption-Proof gain,
$\Delta=\mathrm{Score}_{V}-\mathrm{Score}_{T}$.
Figure~\ref{fig:experimental-diagnostics}(b) reports the average LLM-as-a-Judge gain across the MemEye matrix, with method-specific heatmaps provided in Appendix~\ref{app:caption-proof-validation-details}. Image-based memory helps most when the decisive evidence is fine-grained. In the average heatmap, gains are small in scene-level regions and become positive in fine-grained cells. Bootstrap confidence intervals, reported in Table~\ref{tab:bootstrap-ci-key-comparisons}, are consistent with this diagnostic pattern. Overall, these results suggest that caption-based storage is more likely to lose decisive instance- and pixel-level evidence. Moreover, Appendix Table~\ref{tab:oracle-diagnostics} shows that, when the correct clue rounds are provided, the gap between text-based and multimodal methods widens as the required visual evidence becomes more fine-grained. More results and analysis are provided in Appendix~\ref{app:caption-robustness}.


\subsection{RQ3: Why Do Memory Systems Lose Evolving Visual States?}
\label{sec:retrieval-diagnostics}

RQ2 shows that native image evidence improves fine-grained visual preservation, 
especially in high-$X$ regions. However, this benefit weakens in $Y_3$ 
(Figure~\ref{fig:experimental-diagnostics}(b)), where the answer depends on which 
visual state remains valid after later updates. These cases are not static visual 
recall: the image-grounded evidence itself changes across sessions. This exposes 
a second bottleneck beyond visual preservation, which we call 
\textbf{evolving visual state tracking}.


To isolate evolving-state tracking, we compare representative memory systems with oracle evidence controls on the evolving visual-state subset. Table~\ref{tab:evolving-visual-state-probe} shows that directly providing only the latest valid visual state yields performance close to the full oracle evidence chain. This near match suggests that, for many \(Y_3\) questions, answering correctly depends primarily on identifying the currently valid visual state. In contrast, memory systems remain far below these oracle settings, indicating that they fail not because the final state is visually unreadable, but because they do not recover and prioritize it from an evolving memory history.


Retrieval diagnostics show that related evidence is not enough.
As shown in Figure~\ref{fig:experimental-diagnostics}(c), Appendix~\ref{app:retrieval-diagnostics}, and the case studies in Appendix Figure~\ref{fig:case-studies-3}, SRAG(V) often retrieves evidence about the right topic in \(Y_3\) questions, but can miss the decisive 
latest clue or the complete update chain. In other words, semantic similarity does not guarantee temporal validity. 
Figure~\ref{fig:y3-textual-memory-case-study} further illustrates this failure mode: text-based memory methods can preserve compact evolving-state evidence chains, while multimodal memory methods are distracted by visually similar, stale, or conflicting retrieved images.
In other words, finding related evidence is not the same as selecting the valid state.
If retrieval fails to select the updated evidence, one possible alternative is full-context conditioning: provide the entire history and let the model resolve the valid state. We further conduct a memory scaling analysis by concatenating conversations from different tasks into a larger, more diverse memory history. The results in Figure~\ref{fig:experimental-diagnostics}(d) show that memory mechanisms become increasingly important as history length and topic diversity grow, because they help filter unrelated evidence. 
These findings suggest that multimodal memory should combine image evidence, text, or structured state records, and mechanisms for selecting valid evidence over long histories.
Together, these results show that Y3 failures arise from incomplete evolving-state tracking: systems must not only preserve visual details, but also recover the update chain and prioritize the currently valid visual state over stale evidence.

\section{Implications for Memory Architecture Design}
The results above suggest that current multimodal memory should not be designed as a single retrieval module. MemEye's two axes point to complementary design requirements: the $X$-axis calls for preserving decisive visual evidence at the right granularity, while the $Y$-axis calls for selecting the temporally valid evidence once memory evolves.

\paragraph{How to store visual information?}

RQ2 shows that agent memory systems lose visual information when images are converted into text.
Text-based memory can capture coarse visual information, but it often misses fine-grained visual information.
Image-based memory is therefore needed for high-\(X\) questions, where the answer depends on visual information that captions may omit.
Thus, future memory systems should preserve image evidence rather than relying only on text summaries.

\paragraph{How to select valid visual memory states?}

RQ3 shows that preserving image evidence is still not enough.
When state information changes across sessions, the system must decide which visual state is currently valid.
Text-based or structured memory can do better for high-\(Y\) questions in recording updates, conflicts, and overrides, while image-based memory preserves the visual information needed to check those states. The cross-topic dialog scaling ablation further shows why memory mechanisms are useful as memory grows: full-context methods become more sensitive to unrelated histories, while retrieval-based or structured memory methods remain more stable. This suggests that memory systems need mechanisms that filter, compress, or reweight evidence before answering.
Meanwhile, current mechanisms are not yet state-aware enough: retrieval diagnostics show that semantically relevant evidence can still be stale.
The recency re-ranking probe in Appendix~\ref{app:retrieval-diagnostics} shows that temporal signals can reduce stale-over-latest errors, but they do not fully resolve errors that arise during answer generation.

\paragraph{Summary.}

Together, these findings suggest that multimodal memory should keep both image-based and text-based memory.
Image-based memory preserves fine-grained visual information, while text-based or structured memory helps update state information across sessions.
On top of both, the memory system needs mechanisms that select valid evidence: it should filter unrelated history, use temporal signals, and avoid selecting stale evidence.
A useful future direction is to combine image evidence, structured state records, and recency-aware selection in one memory system.

\section{Conclusion}
Long-term multimodal agents require memory systems that preserve visual evidence and reason over it across time. We introduced MemEye, a visual-centric evaluation framework for multimodal agent memory, organized by a two-axis taxonomy that separates visual evidence granularity from memory reasoning depth. Under this framework, we construct a benchmark with 371 questions across eight life-scenario tasks, using clue-centered construction and validation gates to ensure answerability, shortcut resistance, visual grounding, and reasoning-structure alignment.

Our evaluation of 13 memory architectures across 4 VLM backbones shows that current systems remain far from saturation. The MemEye matrix reveals three recurring failure modes: systems may lose fine-grained visual evidence, retrieve stale evidence, or fail to synthesize the currently valid state. These findings suggest that future multimodal memory systems should combine image evidence, text or structured state records, and mechanisms for selecting temporally valid evidence over long histories. MemEye is diagnostic rather than exhaustive: it focuses on curated life-scenario memory tasks, representative memory architectures, and system-level comparisons, with broader human baselines and deployment-scale studies to future work discussed in Appendix~\ref{app:limitations}.

\section*{Acknowledgments}
The authors would like to thank The Brand Memory Company (TBMC) for providing API funding that supported the large-scale experiments in this work.

\bibliographystyle{plainnat}
\bibliography{reference}

\appendix

\section{Benchmark Construction and Dataset Details}
\label{app:dataset}

\subsection{Task Statistics}
\label{app:task-stats}

Table~\ref{tab:app-task-stats} provides per-task statistics for the MemEye benchmark.

\begin{table}[h]
\centering
\small
\caption{Per-task statistics of the MemEye benchmark.}
\label{tab:app-task-stats}
\begin{tabular}{lrrrr}
\toprule
Task & Sessions & Rounds & Images & Questions \\
\midrule
Home Renovation   & 13 & 120 & 89 & 52 \\
Brand Memory      & 42 & 72 & 30 & 29 \\
Card Playlog      & 4 & 30 & 30 & 48 \\
Cartoon Ent.      & 86 & 299 & 87 & 76 \\
CrossScene Memory & 15 & 117 & 57 & 50 \\
Outdoor Nav.      & 10 & 60 & 40 & 28 \\
Health Care       & 12 & 97 & 62 & 51 \\
Social Chat       & 39 & 53 & 43 & 37 \\
\midrule
Total             & 221 & 848 & 438 & 371 \\
\bottomrule
\end{tabular}
\end{table}

\paragraph{Image provenance.}
Table~\ref{tab:task-overview} marks each task as archival/public (\textbf{A}) or generated/rendered (\textbf{G}).
Brand Memory uses advertisement images from the Pitt Image Ads dataset~\citep{hussain2017automatic}.
Cartoon Ent.\ uses scanned pages from the public-domain comic strip \emph{Alley Oop} and story illustrations from Seed-Story~\citep{yang2025seed}.
Home Renovation uses stock interior-design photographs.
Card Playlog uses HTML-rendered game-state screenshots based on Cardiverse~\citep{li2025cardiverse}.
Social Chat uses synthetic face images from StyleGAN~\citep{karras2019style} (no real people) combined with PIL-rendered chat-UI screenshots.
Outdoor Nav.\ uses dashcam frames from the Japan Open Driving Dataset~\citep{jodd2024}.
CrossScene Memory and Health Care use AI-generated images produced with \texttt{gpt-5.2} (DALL-E) under controlled scene blueprints.

\begin{table}[t]
\centering
\small
\caption{
MemEye dataset composition across eight life-scenario tasks.
Each task contains $n$ multiple-choice questions and an equal number of mirrored open-ended questions.
Source denotes whether visual evidence is archival/public (\textbf{A}) or generated/rendered (\textbf{G}).
}
\label{tab:task-overview}

\setlength{\tabcolsep}{4pt}
\renewcommand{\arraystretch}{1.08}

\definecolor{leisurebg}{RGB}{226,238,255}
\definecolor{domesticbg}{RGB}{232,242,223}
\definecolor{professionalbg}{RGB}{255,229,213}
\definecolor{personalbg}{RGB}{255,239,209}

\begin{tabular}{lllccr}
\toprule
\textbf{Domain} & \textbf{Task} & \textbf{Scenario type} & \textbf{Src.} & \textbf{Primary $(X,Y)$} & \textbf{$n$} \\
\midrule

\rowcolor{leisurebg}
Leisure
& Card Playlog
& game-state tracking
& G & $X_4$, $Y_2$--$Y_3$ & 48 \\

\rowcolor{leisurebg}
Leisure
& Cartoon Ent.
& character and narrative memory
& A & $X_1$--$X_2$, $Y_1$--$Y_3$ & 76 \\

\addlinespace[2pt]

\rowcolor{domesticbg}
Domestic
& Home Renovation
& room-state and design updates
& A & $X_3$--$X_4$, $Y_2$--$Y_3$ & 52 \\

\rowcolor{domesticbg}
Domestic
& Outdoor Nav.
& route and landmark memory
& G & $X_3$, $Y_1$--$Y_2$ & 28 \\

\addlinespace[2pt]

\rowcolor{professionalbg}
Professional
& Brand Memory
& logo and visual identity recall
& A & $X_2$--$X_4$, $Y_1$--$Y_2$ & 29 \\

\rowcolor{professionalbg}
Professional
& CrossScene Memory
& object-state updates across scenes
& G & $X_2$--$X_4$, $Y_2$--$Y_3$ & 50 \\

\addlinespace[2pt]

\rowcolor{personalbg}
Personal
& Health Care
& dashboard and portal updates
& G & $X_1$--$X_4$, $Y_1$--$Y_3$ & 51 \\

\rowcolor{personalbg}
Personal
& Social Chat
& personal visual-detail memory
& G & $X_2$--$X_3$, $Y_1$--$Y_2$ & 37 \\

\midrule
\multicolumn{5}{@{}l}{\textit{Total: 8 tasks across 4 life-scenario domains}} & \textbf{371} \\
\bottomrule
\end{tabular}
\end{table}

\subsection{Taxonomy Cell Distribution}
\label{app:cell-distribution}

Table~\ref{tab:app-xy-distribution} shows the number of questions in each $(X_i, Y_j)$ cell of the two-axis taxonomy.

\begin{table}[h]
\centering
\small
\caption{Question distribution across the $(X, Y)$ taxonomy matrix.}
\label{tab:app-xy-distribution}
\begin{tabular}{lcccc|c}
\toprule
 & $X_1$ & $X_2$ & $X_3$ & $X_4$ & Total \\
\midrule
$Y_1$ & 10 & 11 & 74 & 18 & 113 \\
$Y_2$ & 29 & 21 & 60 & 88 & 198 \\
$Y_3$ & 10 & 10 & 10 & 30 & 60 \\
\midrule
Total & 49 & 42 & 144 & 136 & 371 \\
\bottomrule
\end{tabular}
\end{table}

\subsection{Detailed Taxonomy Definitions}
\label{app:detailed-taxonomy-definitions}

This section provides the detailed annotation definitions behind the compact taxonomy description in \S\ref{sec:benchmark}. MemEye assigns each question an $(X,Y)$ coordinate using a highest-bottleneck rule: the $X$ label is determined by the finest level of decisive visual evidence required for the answer, and the $Y$ label is determined by the deepest memory operation required after the relevant evidence is available.

Table~\ref{tab:taxonomy-definitions} summarizes the definitions used to assign axis labels and presents representative MemEye examples for each axis level. The examples are intentionally drawn from different task domains to show that the taxonomy is attached to evidence requirements and memory operations rather than to task names.

\begin{table*}[t]
\centering
\small
\caption{Detailed MemEye taxonomy definitions with representative dataset examples.}
\label{tab:taxonomy-definitions}

\setlength{\tabcolsep}{5pt}
\renewcommand{\arraystretch}{1.16}

\begin{tabular}{p{0.16\textwidth}p{0.31\textwidth}p{0.45\textwidth}}
\toprule
\textbf{Level} & \textbf{Definition} & \textbf{Representative MemEye example} \\
\midrule

\rowcolor{gray!10}
\multicolumn{3}{@{}l}{\textbf{Visual evidence granularity} $(X)$} \\
\cmidrule(lr){1-3}

$X_1$ Scene-level
& Global scene type, activity, or semantic gist; often recoverable from a good caption.
& In Cartoon Ent., identifying the open blue-green sky and clouds behind a baby dinosaur climbing a tree. \\[4pt]

$X_2$ Region-level
& Semantically coherent subregions, grouped entities, or local spatial structure within a scene.
& In Home Renovation, deciding which cabinet sample is closest to a brass hardware piece lying on the floor. \\[4pt]

$X_3$ Instance-level
& Specific object or person identity among visually or semantically similar candidates.
& In Home Renovation, matching the labeled cabinet sample from a three-sample comparison to the same sample later shown beside a tape measure and pencil. \\[4pt]

$X_4$ Pixel-level
& Fine visual details such as small text, exact color, texture, count, or OCR-like evidence.
& In CrossScene Memory, reading the current identification tag number displayed inside the fossil-room case. \\

\midrule

\rowcolor{gray!10}
\multicolumn{3}{@{}l}{\textbf{Memory-reasoning depth} $(Y)$} \\
\cmidrule(lr){1-3}

$Y_1$ Atomic retrieval
& One sufficient evidence unit answers the question; no cross-session composition is needed.
& A single clue image in Cartoon Ent. is enough to answer what the background looks like in the tree-climbing scene. \\[4pt]

$Y_2$ Relational association
& Multiple non-conflicting clues must be linked across sessions, modalities, or references.
& In Cartoon Ent., a palace-arc question requires comparing two remembered events to decide whether the crowd pelting or dinner fight occurred first. \\[4pt]

$Y_3$ Evolutionary synthesis
& Temporally ordered clues include updates, conflicts, or overrides; the answer depends on the valid current state.
& In the fossil-room case, an earlier identification tag is later replaced, so the answer must use the updated tag rather than the stale one. \\

\bottomrule
\end{tabular}
\end{table*}

\begin{figure*}[t]
  \centering
  \includegraphics[width=\textwidth]{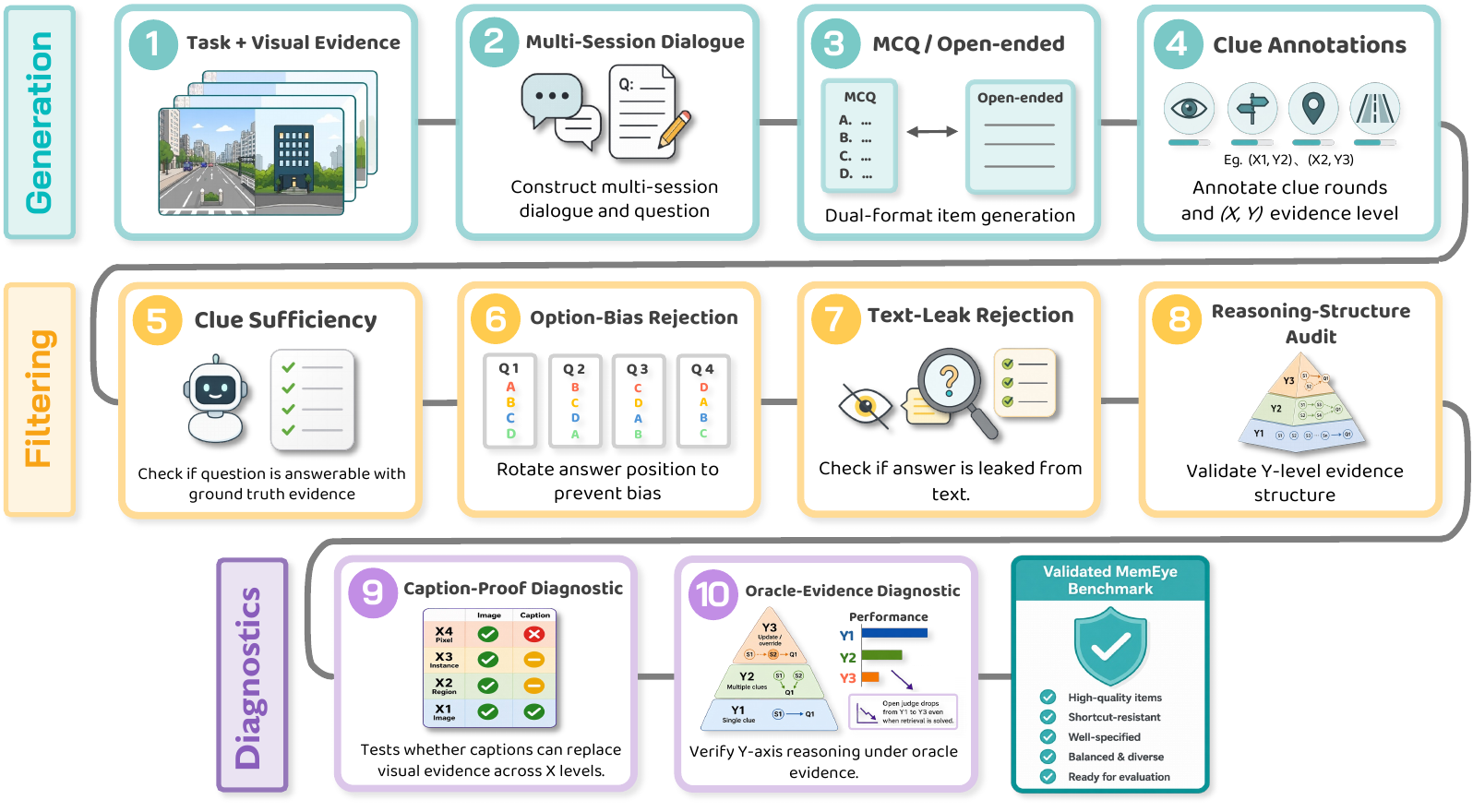}
  \caption{Complete MemEye construction and validation pipeline. Candidate questions are generated from task scenarios, visual evidence, and target $(X,Y)$ regions, then authored as mirrored MCQ/open-ended items with clue-round annotations and four-way MCQ rotations. Item-level checks establish multimodal validity, memory answerability, and taxonomy alignment by removing shortcut-answerable or under-specified candidates and verifying that the clue structure matches the annotated $Y$ level. Finally, aggregate diagnostics validate the two benchmark axes: caption substitution probes $X$-axis visual irreplaceability, and oracle evidence probes $Y$-axis reasoning over retrieved memory.}
  \label{fig:benchmark-validation-pipeline}
\end{figure*}

\subsection{Item-Level Filtering and Taxonomy Audit Details}
\label{app:item-filtering-details}

Stage 2 filters and audits candidate questions before they enter the locked benchmark. All MCQ-based gates use four-way answer rotations and a two-level model panel consisting of \texttt{gpt-5.4-mini} and \texttt{gpt-5.2}. For shortcut-resistance gates, robust success means that both panel models answer all four rotations correctly (rotation-averaged EM \(=1.0\)). For the oracle answerability gate, a candidate is treated as unsolved if both panel models remain at or below chance level under oracle visual evidence, i.e., rotation-averaged EM \(\le 0.25\). A candidate is flagged when shortcut success is robust across rotations, when it remains unsolved under oracle visual evidence, or when its clue structure does not match the intended $Y$ label. Flagged candidates are revised, removed, or relabeled.

\paragraph{Shortcut-Resistance Gates.}
We first test whether a candidate can be answered without the intended multimodal memory signal. In the option-only gate, the model receives only the question and rotated answer options, with no context, image, or caption. Robust success indicates that the option set, world knowledge, or stylistic artifacts leak the answer, so the item is flagged. In the text-only clue gate, the model receives the question, options, and clue-round text, but no image or caption. Robust success indicates textual leakage from the dialogue, so the item is flagged. In the minimal-caption gate, original images are replaced by very short captions that only keep the rough image type, such as a room photo, a game board, a cartoon panel, or a phone screenshot. This gate is applied to all \(X_1\)--\(X_4\) candidates. Robust success indicates that the question is too easily answered without the original image, so the item is flagged for revision or removal. This gate does not define the \(X\) label; the \(X\) label is assigned by the finest visual evidence needed to answer the question.

\paragraph{Oracle Answerability Gate.}
We then provide the gold-clue rounds and the original images. This oracle-evidence setting removes memory search as the bottleneck and checks whether the candidate is well-defined, visually answerable, and within the intended difficulty range. Candidates that remain unsolved under oracle visual evidence are treated as visually ambiguous, underspecified, or above the intended difficulty level, and are flagged. This gate ensures that MemEye evaluates memory access and evidence use rather than impossible visual recognition.

\paragraph{Taxonomy-Structure Audit.}
Finally, we audit whether the clue structure matches the annotated $Y$ level. A $Y_1$ item must be answerable from one evidence unit. A $Y_2$ item must require associating multiple non-redundant evidence units under monotonic accumulation; no single clue should collapse it into atomic retrieval. A $Y_3$ item must require temporally ordered evidence in which a later clue updates, overrides, or conflicts with an earlier state; the answer must depend on resolving the valid current state rather than retrieving either clue in isolation. This audit ensures that the $Y$-axis reflects evidence use rather than task naming.

\subsection{Annotation Agreement}
\label{app:annotation-agreement}

MemEye questions are initially generated by \texttt{gpt-5.2} conditioned on target $(X,Y)$ coordinates, then reviewed and adjudicated by human annotators who inspect the visual evidence, clue rounds, and question--answer pairs. The agreement study below is a reproducibility check on the taxonomy definitions, not a substitute for human adjudication. To quantify label reproducibility, we measure inter-annotator agreement on a stratified subset of 100 questions using two strong LLMs as independent annotators: GPT-5.4 and Gemini-2.5-Pro.

\paragraph{Protocol.}
We sample 100 questions stratified across all $(X,Y)$ cells (8--10 per cell). Each model receives the question, ground-truth answer, annotator explanation, and the full taxonomy definitions, including boundary examples and decision rules (e.g., ``X2 is about WHERE things are in a local region; X3 is about WHICH specific instance is being referred to among similar candidates''). The models independently assign $X$ and $Y$ labels. Neither model has access to the original images; labels are assigned from the textual description of visual evidence in the question and explanation. Therefore, this agreement study measures the reproducibility of the written taxonomy definitions and annotation rationales, rather than serving as primary evidence that visual bottlenecks are objectively labeled from images. Final label adjudication is performed by human annotators who inspect the original images, dialogue context, clue rounds, and question-answer pairs. Visual bottleneck validity is evaluated separately through the Caption-Proof diagnostic and caption-robustness ablation.

\paragraph{Results.}
Inter-annotator agreement is substantial on both axes: $\kappa_X = 0.66$ and $\kappa_Y = 0.63$ (Cohen's $\kappa$). The two models agree on 76 of 100 $X$ labels and 77 of 100 $Y$ labels. Remaining $X$-axis disagreement concentrates on the $X_2$/$X_3$ boundary (region-level vs.\ instance-level), which often requires inspecting the actual image to determine whether the bottleneck is spatial layout or identity binding. $Y$-axis disagreement is distributed more evenly, with occasional confusion between $Y_1$ (atomic retrieval) and $Y_2$ (relational association) when the number of required evidence units is ambiguous from the textual description alone.

\paragraph{Human adjudication.}
All 371 questions in the final benchmark have been reviewed by a human annotator who inspects the images, dialogue context, and clue structure. The LLM-based agreement measurement reported here serves as a reproducibility check on the taxonomy definitions rather than as a replacement for human review. Cases where the two LLM annotators disagreed were cross-checked against the human-adjudicated gold labels; the gold labels were revised only when the disagreement revealed a genuine annotation error rather than a boundary judgment call.

\section{Benchmark Validation}
\label{app:benchmark-validation}

\subsection{Caption-Proof Validation Details}
\label{app:caption-proof-validation-details}

\begin{figure*}[t]
  \centering
  \includegraphics[width=\textwidth]{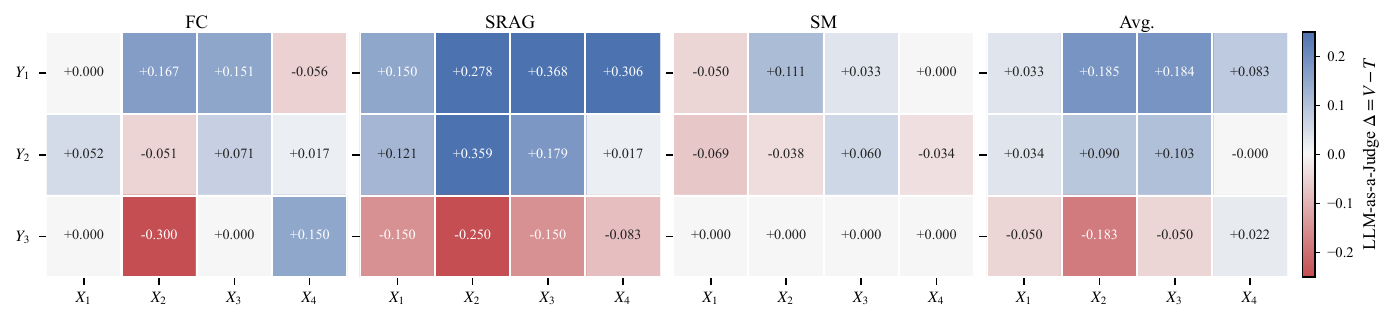}
  \caption{
  Full Caption-Proof heatmaps on MemEye under \texttt{gpt-5.4-mini}. Each cell reports
  $\Delta=\mathrm{Score}_{V}-\mathrm{Score}_{T}$ for matched textual and visual
  streams, where textual streams use dense captions and visual streams use native
  images. The main text reports the average heatmap in Figure~\ref{fig:experimental-diagnostics}(b).
  }
  \label{fig:caption-proof-delta-full}
\end{figure*}

\begin{table}[t]
  \centering
  \small
  \caption{Caption-proof validation on \texttt{gpt-5.4-mini}. Rows aggregate over $Y_1$-$Y_3$ for each visual granularity level. Judge(T) and Judge(V) denote LLM-Judge scores for matched caption-based and native-visual streams; $\Delta$ is the visual-stream gain.}
  \label{tab:caption_proof_validation}
  \setlength{\tabcolsep}{2.6pt}
  \renewcommand{\arraystretch}{1.02}
  \begin{tabular}{llrrrrr}
  \toprule
  Method & $X$ & $n$ & Judge(T) & Judge(V) & $\Delta_{\mathrm{J}}$ & $\Delta_{\mathrm{EM}}$ \\
  \midrule
  FC   & X1 & 49  & 0.5306 & 0.5612 & +0.0306 & +0.0765 \\
  FC   & X2 & 42  & 0.2976 & 0.2857 & -0.0119 & -0.0952 \\
  FC   & X3 & 144 & 0.3854 & 0.4653 & +0.0799 & +0.1076 \\
  FC   & X4 & 136 & 0.3640 & 0.4007 & +0.0368 & +0.0478 \\
  \midrule
  SRAG & X1 & 49  & 0.5306 & 0.6020 & +0.0714 & -0.0510 \\
  SRAG & X2 & 42  & 0.2738 & 0.4286 & +0.1548 & +0.0357 \\
  SRAG & X3 & 144 & 0.3194 & 0.6076 & +0.2882 & +0.1944 \\
  SRAG & X4 & 136 & 0.3162 & 0.3493 & +0.0331 & +0.0772 \\
  \midrule
  SM   & X1 & 49  & 0.5204 & 0.4694 & -0.0510 & +0.0612 \\
  SM   & X2 & 42  & 0.2143 & 0.1667 & -0.0476 & +0.0238 \\
  SM   & X3 & 144 & 0.2361 & 0.2812 & +0.0451 & +0.0260 \\
  SM   & X4 & 136 & 0.2353 & 0.2132 & -0.0221 & -0.0110 \\
  \bottomrule
  \end{tabular}
\end{table}

Caption-Proof validation assesses whether a benchmark question requires native visual evidence, rather than being solvable from captions alone. For each question, we evaluate matched textual and multimodal memory systems under the same memory architecture. The textual memory replaces every image with a dense caption. The multimodal memory keeps the original images. We then report the visual gain
$\Delta=\mathrm{score}_{V}-\mathrm{score}_{T}$, where positive values indicate that native images preserve critical evidence that captions do not.

\begin{figure*}[htbp!]
\centering
\footnotesize

\newcommand{\panelrule}{\vspace{4pt}\hrule\vspace{5pt}}
\newcommand{\caseinnerwidth}{0.94\linewidth}

\newcommand{\caseheader}[4]{%
\parbox[t][0.78in][t]{\caseinnerwidth}{\centering
\textbf{#1}\\[2pt]
\textit{Cell:} #2\\[1pt]
FC-V\,=\,#3\\[1pt]
FC-T\,=\,#4
}%
}

\newcommand{\caseqa}[1]{%
\parbox[t][1.08in][t]{\caseinnerwidth}{\scriptsize #1}%
}

\newcommand{\casecaptionbox}[1]{%
\colorbox{gray!10}{%
\parbox[t][1.10in][t]{0.91\linewidth}{\scriptsize #1}%
}%
}

\newcommand{\casewhy}[1]{%
\parbox[t][0.66in][t]{\caseinnerwidth}{\scriptsize #1}%
}

\fbox{%
\begin{minipage}[t]{0.305\textwidth}
\centering
\vspace{4pt}

\caseheader
{Case 1: Cross-Session Identity Tracking}
{$(X_3,Y_2)$}
{1.00}
{0.00}

\panelrule

\begin{minipage}[t][0.75in][t]{\caseinnerwidth}
\centering
\includegraphics[width=0.32\linewidth,height=0.43in,keepaspectratio]{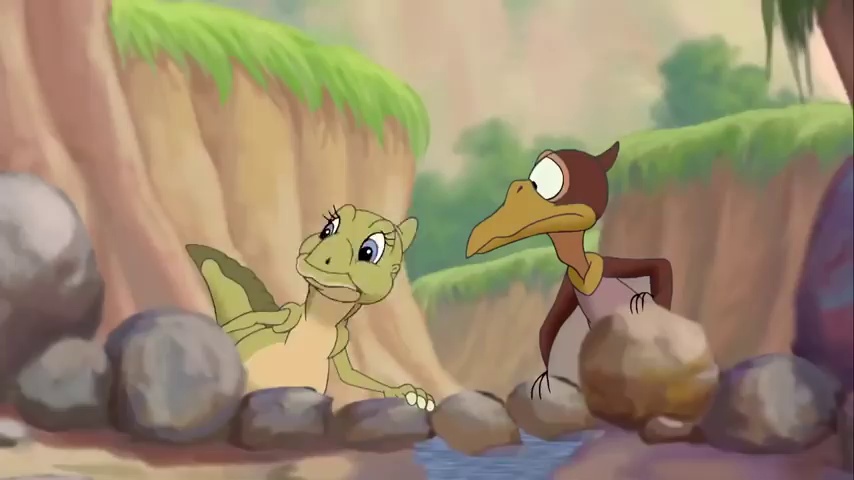}\hfill
\includegraphics[width=0.32\linewidth,height=0.43in,keepaspectratio]{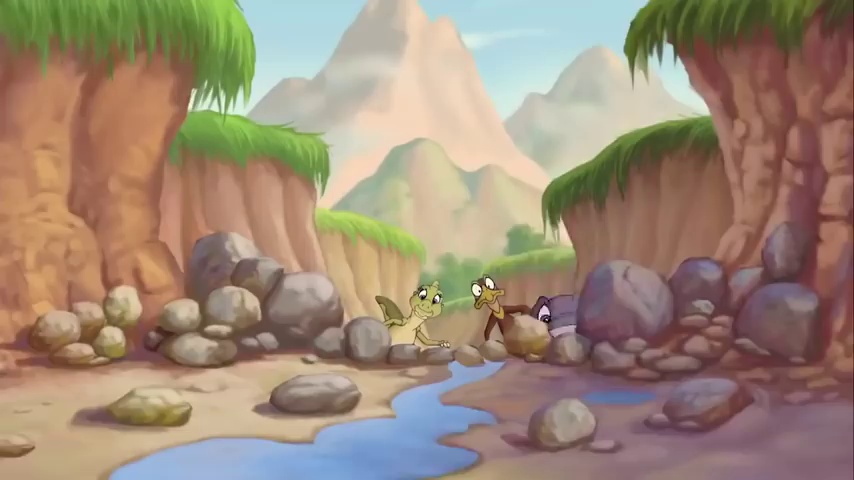}\hfill
\includegraphics[width=0.32\linewidth,height=0.43in,keepaspectratio]{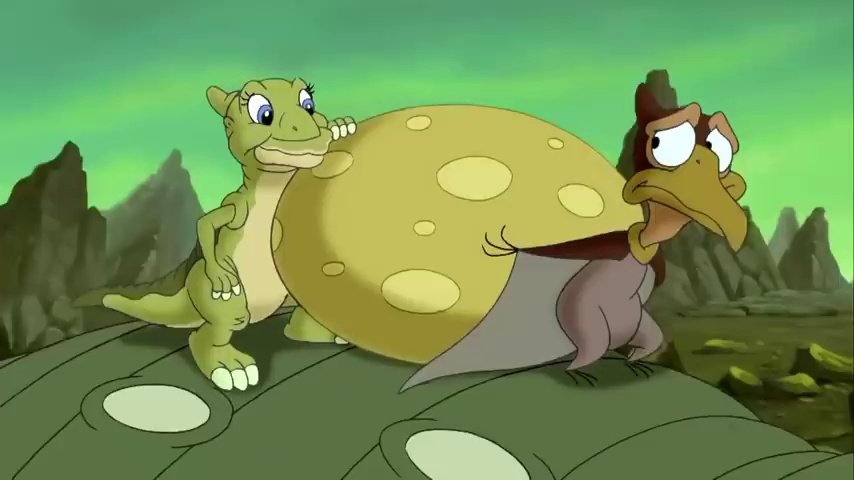}

\vspace{2pt}
\parbox[t]{0.32\linewidth}{\centering\scriptsize S5: Pair by rocks}\hfill
\parbox[t]{0.32\linewidth}{\centering\scriptsize S5: Pair in valley}\hfill
\parbox[t]{0.32\linewidth}{\centering\scriptsize S9: Pair with egg}
\end{minipage}

\panelrule

\caseqa{
\textbf{Q:} A small green dinosaur and a brown bird appeared as a pair in two Episode~1 scenes. Were these same two characters also seen together holding a large egg in a later session?

\vspace{3pt}
\textbf{A:} Yes---the same green dinosaur and brown bird appear consistently across all three scenes.
}

\vspace{1pt}

\casecaptionbox{
\textbf{Captions seen by text-only model:}\\
S5-R6: \textit{``Two animated dinosaurs peek over rocks beside a stream in a canyon.''}\\
S5-R7: \textit{``Two cartoon dinosaurs stand in a rocky canyon beside a small stream.''}\\
S9-R3: \textit{``A green dinosaur stands on the back of a large yellow-spotted dinosaur in a rocky landscape.''}
}

\vspace{4pt}

\casewhy{
\textit{Why caption fails:} The captions lose species identity, body shape, and skin texture---the visual cues needed to confirm the same pair across sessions.
}

\vspace{4pt}
\end{minipage}
}
\hfill
\fbox{%
\begin{minipage}[t]{0.305\textwidth}
\centering
\vspace{4pt}

\caseheader
{Case 2: Micro-Attribute Comparison}
{$(X_4,Y_2)$}
{1.00}
{0.00}

\panelrule

\begin{minipage}[htbp][0.75in][t]{\caseinnerwidth}
\centering
\includegraphics[width=0.48\linewidth,height=0.43in,keepaspectratio]{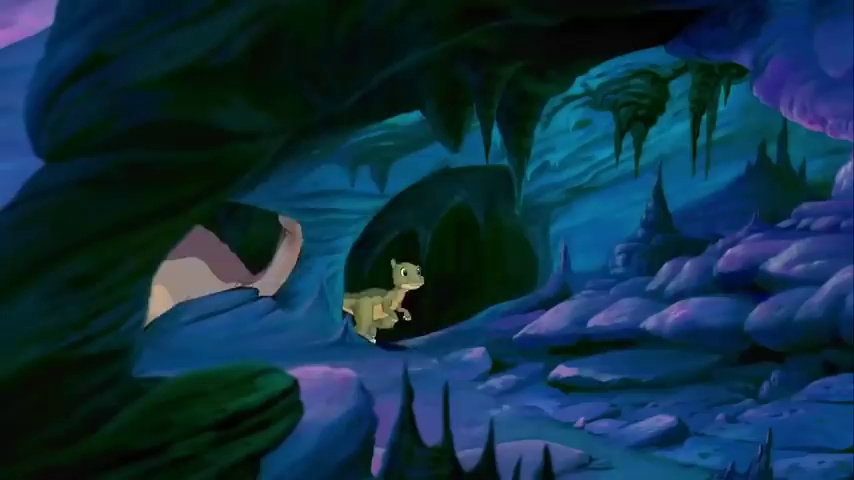}\hfill
\includegraphics[width=0.48\linewidth,height=0.43in,keepaspectratio]{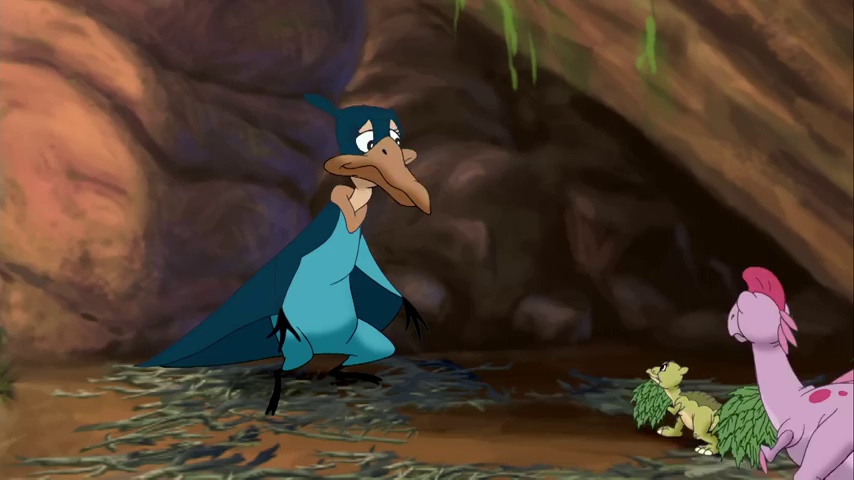}

\vspace{2pt}
\parbox[t]{0.48\linewidth}{\centering\scriptsize Ep.~1 cave (S1): stalactites}\hfill
\parbox[t]{0.48\linewidth}{\centering\scriptsize Ep.~2 cave (S7): vegetation}
\end{minipage}

\panelrule

\caseqa{
\textbf{Q:} Both episodes open with a cave scene. In Episode~1's dark cave, icicle-shaped rock formations hang from the ceiling. In Episode~2's cave, what hangs from the ceiling instead?

\vspace{3pt}
\textbf{A:} Green hanging vegetation and vines.
}

\vspace{1pt}

\casecaptionbox{
\textbf{Captions seen by text-only model:}\\
S1-R4: \textit{``A small dinosaur stands inside a blue cave with stalactites and rocky walls.''}\\
S7-R3: \textit{``A blue bird stands in a cave facing two small dinosaurs.''}
}

\vspace{4pt}

\casewhy{
\textit{Why caption fails:} S1's caption mentions ``stalactites,'' but S7's caption omits the ceiling entirely, making the cross-episode comparison impossible.
}

\vspace{4pt}
\end{minipage}
}
\hfill
\fbox{%
\begin{minipage}[t]{0.305\textwidth}
\centering
\vspace{4pt}

\caseheader
{Case 3: Fine-Grained Color Memory}
{$(X_4,Y_2)$}
{1.00}
{0.25}

\panelrule

\begin{minipage}[t][0.75in][t]{\caseinnerwidth}
\centering
\includegraphics[width=0.48\linewidth,height=0.43in,keepaspectratio]{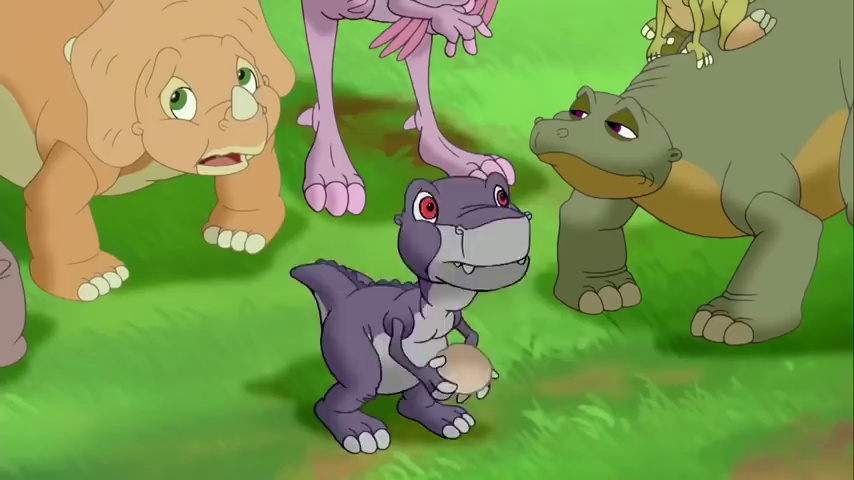}\hfill
\includegraphics[width=0.48\linewidth,height=0.43in,keepaspectratio]{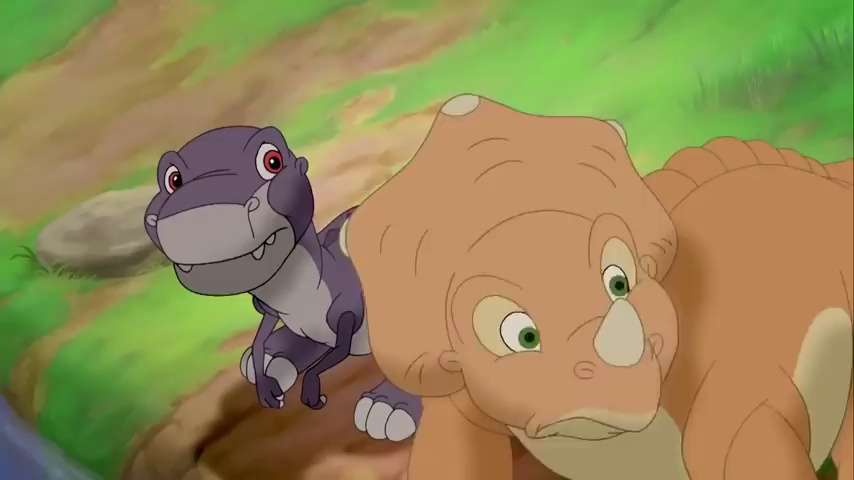}

\vspace{2pt}
\parbox[t]{0.48\linewidth}{\centering\scriptsize S5: First close-up}\hfill
\parbox[t]{0.48\linewidth}{\centering\scriptsize S10: Later close-up}
\end{minipage}

\panelrule

\caseqa{
\textbf{Q:} In the close-up where two characters sit on green grass---one purple, one orange---what are their exact eye colors?

\vspace{3pt}
\textbf{A:} Purple character: red eyes; orange character: green eyes.
}

\vspace{1pt}

\casecaptionbox{
\textbf{Captions seen by text-only model:}\\
S5-R3: \textit{``A group of cartoon dinosaurs stands on green grass, with a small purple dinosaur holding an egg in the center.''}\\
S10-R4: \textit{``Two cartoon dinosaurs stand together outdoors.''}
}

\vspace{4pt}

\casewhy{
\textit{Why caption fails:} Neither caption mentions eye color. The text-only model must guess, while the visual model reads the exact hue from the pixel data.
}

\vspace{4pt}
\end{minipage}
}

\vspace{4pt}

\caption{
Three Caption-Proof examples from MemEye's Cartoon Ent. task.
Each question requires native visual input to answer; replacing images with dense captions causes accuracy to collapse.
\textbf{Case~1} tests cross-session character identity tracking $(X_3,Y_2)$.
\textbf{Case~2} tests micro-attribute comparison $(X_4,Y_2)$.
\textbf{Case~3} tests fine-grained color recall $(X_4,Y_2)$.
Together, these cases illustrate why MemEye's Caption-Proof protocol is essential: without it, benchmark scores may reflect text-based retrieval rather than genuine visual memory.
}
\label{fig:case-studies}
\end{figure*}

\begin{figure*}[htbp]
\centering
\footnotesize

\setlength{\fboxsep}{3pt}
\setlength{\fboxrule}{0.4pt}

\newcommand{\panelrule}{\vspace{4pt}\hrule\vspace{5pt}}
\newcommand{\caseinnerwidth}{0.94\linewidth}

\newcommand{\caseheader}[5]{%
\parbox[t][0.58in][t]{\caseinnerwidth}{\centering
\textbf{#1}\\[1pt]
\textit{Task:} #2 \quad \textit{Cell:} #3\\[1pt]
FC-V\,=\,#4 \quad FC-T\,=\,#5
}%
}

\newcommand{\caseqa}[2]{%
\parbox[t][#1][t]{\caseinnerwidth}{\scriptsize #2}%
}

\newcommand{\casecaptionbox}[2]{%
\colorbox{gray!10}{%
\parbox[t][#1][t]{0.91\linewidth}{\scriptsize #2}%
}%
}

\newcommand{\casewhy}[2]{%
\parbox[t][#1][t]{\caseinnerwidth}{\scriptsize #2}%
}

\fbox{%
\begin{minipage}[t]{0.535\textwidth}
\centering
\vspace{4pt}

\caseheader
{Case 4: State-Evolving Belief Revision ($A{\to}B{\to}A$)}
{Health Care}
{$(X_4,Y_3)$}
{1.00}
{0.00}

\panelrule

\begin{minipage}[t][0.92in][t]{\caseinnerwidth}
\centering
\includegraphics[width=0.30\linewidth,height=0.55in,keepaspectratio]{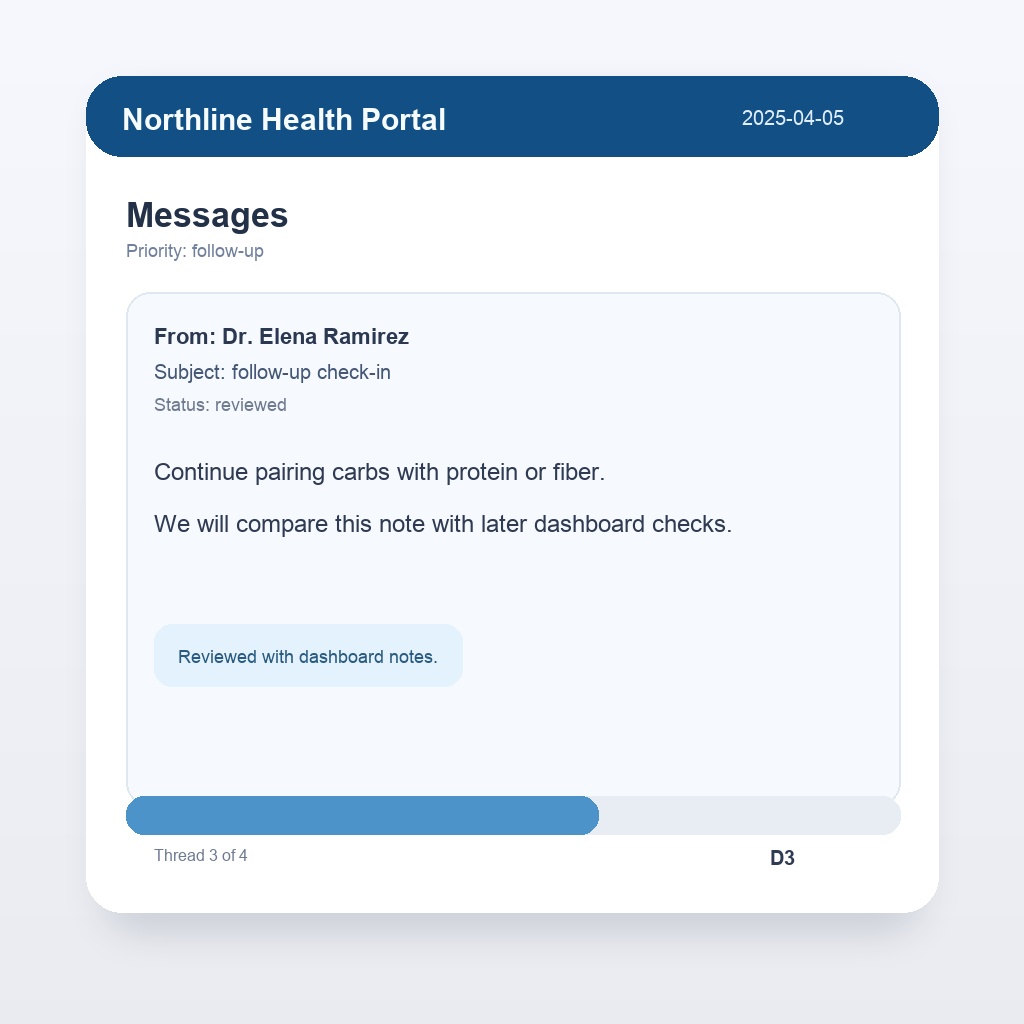}\hfill
\includegraphics[width=0.30\linewidth,height=0.55in,keepaspectratio]{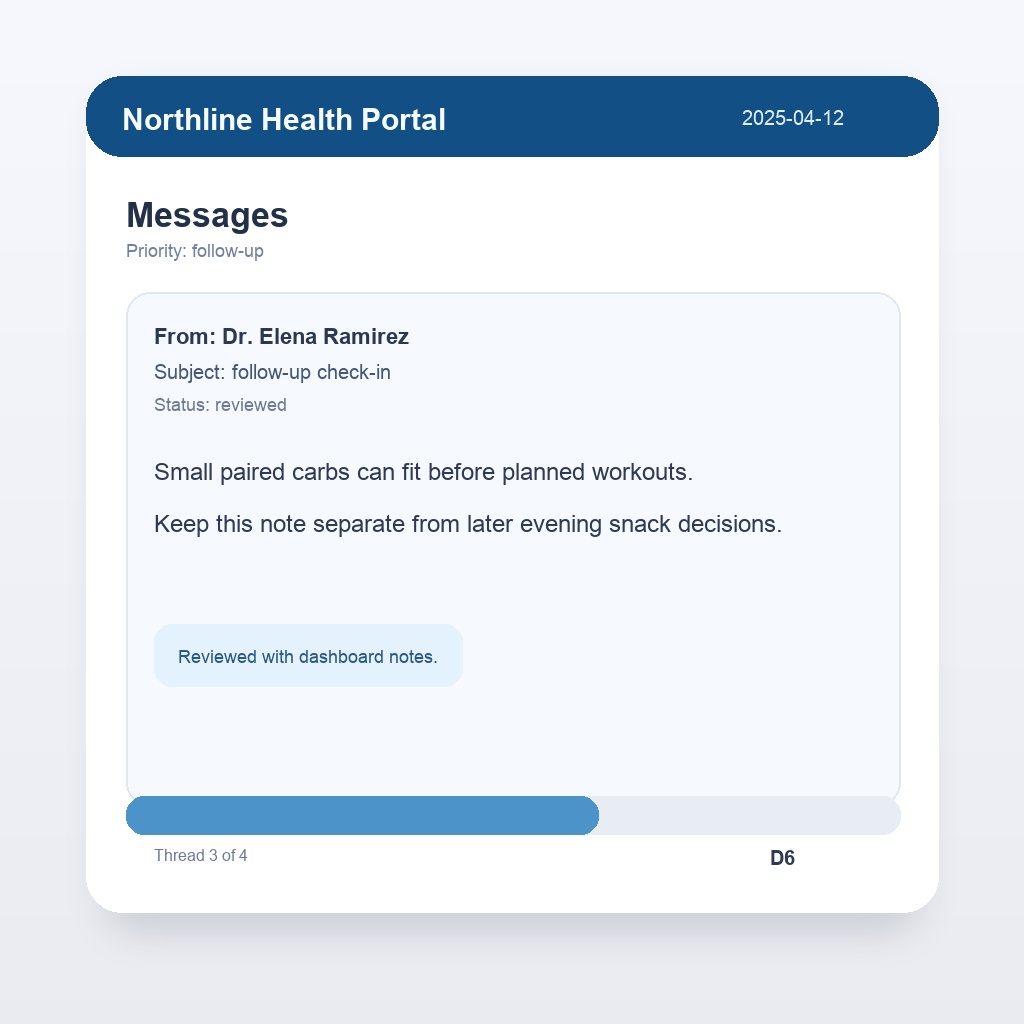}\hfill
\includegraphics[width=0.30\linewidth,height=0.55in,keepaspectratio]{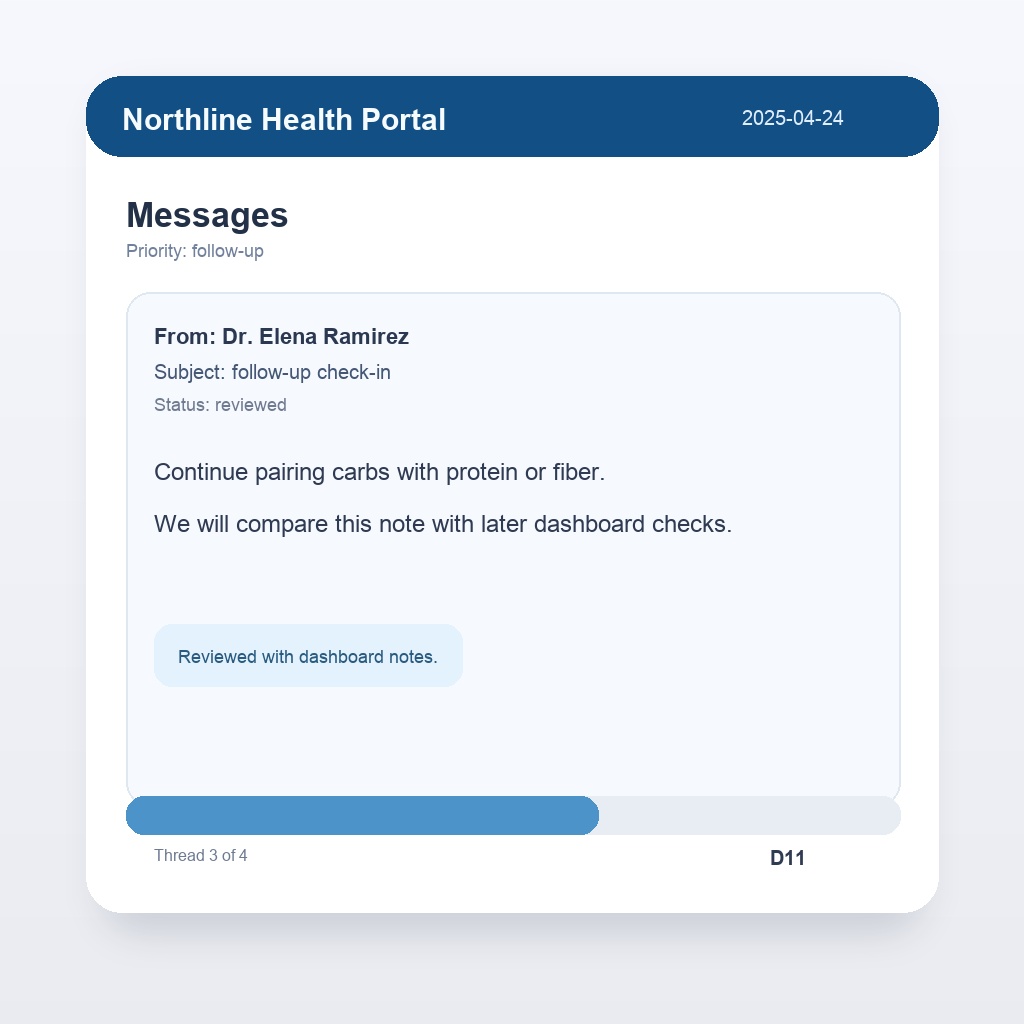}

\vspace{2pt}
\parbox[t]{0.30\linewidth}{\centering\scriptsize D3: ``Pair carbs with protein''}\hfill
\parbox[t]{0.30\linewidth}{\centering\scriptsize D6: ``Manage carb snacks''}\hfill
\parbox[t]{0.30\linewidth}{\centering\scriptsize D11: ``Continue pairing carbs''}
\end{minipage}

\panelrule

\caseqa{0.72in}{
\textbf{Q:} Across all doctor portal messages Maya received during the month, what is the focus of the MOST RECENT guidance from Dr.\ Ramirez?

\vspace{3pt}
\textbf{A:} Continue pairing carbs with protein or fiber.
}

\vspace{1pt}

\casecaptionbox{0.96in}{
\textbf{Captions seen by text-only model:}\\
D3: \textit{``A health portal message from a doctor advises the patient to keep pairing carbs with protein or fiber\ldots''}\\
D6: \textit{``A health portal message shows a doctor advising a patient on managing small carb snacks around workouts.''}\\
D11: \textit{``A patient views a follow-up nutrition message from their doctor in the Northline Health Portal.''}
}

\vspace{4pt}

\casewhy{0.82in}{
\textit{Why caption fails:} D3 and D6 captions reveal their content, but the critical D11 caption is generic and does not preserve which guidance remains current. A text-only model sees that a follow-up exists but lacks the D11 state needed to complete the $A{\to}B{\to}A$ chain, so it answers B.
}

\vspace{4pt}
\end{minipage}
}
\hfill
\fbox{%
\begin{minipage}[t]{0.405\textwidth}
\centering
\vspace{4pt}

\caseheader
{Case 5: Game-State Tracking Under Updates}
{Card Playlog}
{$(X_4,Y_3)$}
{1.00}
{0.25}

\panelrule

\begin{minipage}[t][0.92in][t]{\caseinnerwidth}
\centering
\includegraphics[width=0.72\linewidth,height=0.55in,keepaspectratio]{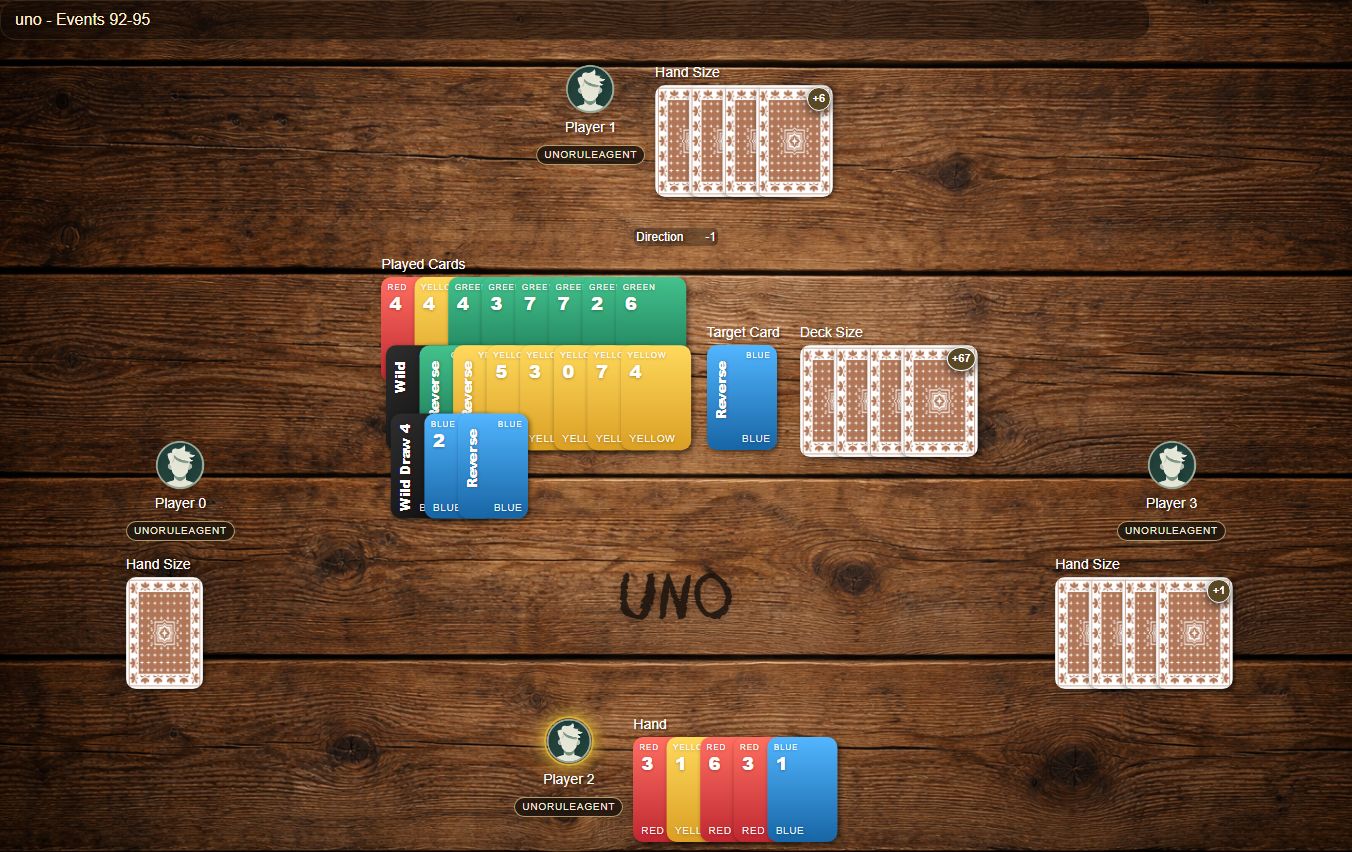}

\vspace{2pt}
\parbox[t]{0.82\linewidth}{\centering\scriptsize UNO state after Player~3's hand changes from 5 to 4 cards}
\end{minipage}

\panelrule

\caseqa{0.72in}{
\textbf{Q:} Immediately after Player~3's visible hand size changes from 5 to 4 for the 1st time, how many red cards does Player~2 hold?

\vspace{3pt}
\textbf{A:} 3
}

\vspace{1pt}

\casecaptionbox{0.96in}{
\textbf{Caption seen by text-only model:}\\
\textit{``Digital UNO game board showing four AI players, a central pile of played cards with a blue Reverse as the target card, a\ldots''}
}

\vspace{4pt}

\casewhy{0.82in}{
\textit{Why caption fails:} The caption describes the board layout generically but omits per-player hand composition. Counting red cards in a specific player's hand at a precise game state requires reading the actual card faces from the screenshot---a fine-grained visual task ($X_4$) combined with temporal state tracking ($Y_3$).
}

\vspace{4pt}
\end{minipage}
}

\vspace{4pt}

\caption{
Two additional Caption-Proof cases highlighting MemEye's $Y_3$ state-evolving synthesis dimension.
Case~4 demonstrates an $A{\to}B{\to}A$ belief reversal across three doctor portal messages: the model must preserve the most recent portal-message state to discover that the final guidance reverts to the initial advice, a fact invisible in the generic D11 caption.
\textbf{Case~5} requires counting specific card colors in a player's hand at a precise temporal state---information that captions never enumerate.
Both cases show that MemEye's hardest region, high $Y$ $\times$ high $X$, demands joint visual fidelity and temporal reasoning that text surrogates cannot support.
}
\label{fig:case-studies-2}
\end{figure*}

Table~\ref{tab:caption_proof_validation} and Figure~\ref{fig:caption-proof-delta-full} report the aggregate Caption-Proof gaps on \texttt{gpt-5.4-mini} for the three matched method families used in the main analysis: Full Context, Semantic RAG, and SimpleMem. Rows aggregate over $Y_1$-$Y_3$ within each visual-granularity level. The table is intended as an X-axis diagnostic: scene- and region-level evidence are often included in captions, whereas instance identity, exact attributes, small text, and temporally valid visual states are more likely to be lost when images are converted to text. The qualitative cases in Figures~\ref{fig:case-studies} and~\ref{fig:case-studies-2} illustrate this failure mode directly.

\subsection{Oracle-Evidence Validation Details}
\label{app:oracle-evidence-validation-details}

Oracle-evidence validation tests whether the annotated axes reflect the intended memory requirements after retrieval difficulty is removed. For each question, the model is given the gold-clue rounds and the original images, while controlling for search over the full history and preserving the evidence needed to answer.

Table~\ref{tab:oracle-diagnostics}(a) validates the $Y$-axis on \texttt{gpt-5.4-mini}. We use open-ended LLM-Judge as the primary metric because multiple-choice options can mask reasoning failures; EM and BLEU-1 are reported as auxiliary metrics. LLM-Judge decreases from $Y_1$ to $Y_3$ ($0.673 \rightarrow 0.601 \rightarrow 0.558$), showing that reasoning remains harder even when the relevant evidence is provided. This supports the interpretation of $Y_1$ as atomic retrieval, $Y_2$ as relational association, and $Y_3$ as evolutionary synthesis over updates, overrides, or conflicts.

Table~\ref{tab:oracle-diagnostics}(b) validates the $X$-axis by comparing text-only and multimodal gold-evidence settings. The visual gap increases from $+0.122$ at $X_1$ to $+0.298$ at $X_4$, indicating that native visual evidence becomes more important as the decisive evidence shifts from scene-level content to instance- and pixel-level details.

\begin{table}[t]
\centering
\small
\caption{
Oracle-evidence diagnostics on \texttt{gpt-5.4-mini}. 
Left: performance by memory-reasoning level when gold clue rounds and original images are provided. 
Right: LLM-Judge visual gap by visual granularity under gold evidence.
}
\label{tab:oracle-diagnostics}
\setlength{\tabcolsep}{4pt}
\renewcommand{\arraystretch}{1.05}
\begin{tabular}{lcccc@{\qquad}lrrrr}
\toprule
\multicolumn{5}{c}{\textbf{(a) Reasoning depth}} 
& \multicolumn{5}{c}{\textbf{(b) Visual granularity}} \\
\cmidrule(lr){1-5}\cmidrule(lr){6-10}
$Y$ & $n$ & EM & B-1 & Judge
& $X$ & $n$ & Text & Visual & $\Delta$ \\
\midrule
$Y_1$ Atomic & 113 & 0.856 & 0.412 & 0.673
& $X_1$ Scene & 49 & 0.653 & 0.776 & +0.122 \\
$Y_2$ Rel. & 198 & 0.633 & 0.426 & 0.601
& $X_2$ Region & 42 & 0.262 & 0.524 & +0.262 \\
$Y_3$ Evol. & 60 & 0.696 & 0.327 & 0.558
& $X_3$ Inst. & 144 & 0.358 & 0.622 & +0.264 \\
&&&&
& $X_4$ Pixel & 136 & 0.335 & 0.632 & +0.298 \\
\bottomrule
\end{tabular}
\end{table}

\subsection{Human Validation Under Oracle Evidence}
\label{app:human-sanity-check}

To verify that MemEye questions are answerable by humans when the relevant evidence is provided, three annotators independently answer a stratified subsample of MCQ questions given only the gold clue rounds and original images. Annotators do not see the full conversation history. For $Y_3$ items, clue rounds are presented in chronological order with instructions to answer based on the latest valid state.

Table~\ref{tab:human-sanity-check} reports accuracy by $Y$ level. We treat this study as a small-scale human oracle sanity check rather than a full human ceiling estimate. Human annotators achieve near-perfect accuracy on $Y_1$, while accuracy decreases on $Y_3$. This lower $Y_3$ score does not by itself indicate that the questions are visually unanswerable: $Y_3$ items require resolving updates, conflicts, or overrides across clue rounds, and annotators can still make state-resolution errors even when retrieval is removed. At the same time, the result shows that $Y_3$ remains harder under oracle evidence, so we do not use this study to claim near-perfect human performance on all evolutionary-synthesis items. Instead, it serves as a sanity check on the sampled questions, while the benchmark construction relies on separate human adjudication and oracle answerability gates to remove ambiguous or under-specified candidates.

\begin{table}[t]
  \centering
  \small
  \caption{Human oracle sanity check. Three annotators independently answer a stratified subsample of questions given only the gold clue rounds and original images. We report majority-vote accuracy and mean individual accuracy by $Y$ level. This serves as a sanity check rather than a full human ceiling estimate.}
  \label{tab:human-sanity-check}
  \setlength{\tabcolsep}{5pt}
  \begin{tabular}{lcccc}
    \toprule
    Metric & $Y_1$ & $Y_2$ & $Y_3$ & All \\
    \midrule
    Majority-vote accuracy    & 1.00 & 0.83 & 0.81 & 0.88 \\
    Mean individual accuracy  & 0.97 & 0.81 & 0.77 & 0.85 \\
    \bottomrule
  \end{tabular}
\end{table}

\section{Evaluation Protocol and Implementation}
\label{app:evaluation-protocol}

\subsection{Benchmark Runner and Method Settings}
\label{app:evaluation-implementation-details}

All backbone runs use deterministic decoding for answer generation, with
temperature $T{=}0$. The main \texttt{gpt-5.4-mini} configuration uses a
maximum generation length of 128 tokens for benchmark answers. For open-ended
questions, LLM-as-a-Judge scoring is enabled with \texttt{gpt-5.2}; the
judge prompt returns a normalized semantic-correctness score in $[0,1]$ and a
short rationale. Multiple-choice scores are computed from the extracted answer
letter and averaged over the four answer-rotation variants of each original
question.

We evaluate 4 open- and closed-source VLM backbones:
Qwen3-VL-8B-Instruct~\citep{bai2025qwen3},
\texttt{gpt-4.1-nano} and \texttt{gpt-5.4-mini}~\citep{openai_models_2026}, and \texttt{gemini-2.5-flash-lite}~\citep{google_gemini_models_2026}. The thirteen memory
methods are Full Context (FC(T/V)), Semantic RAG (SRAG(T/V)), A-Mem,
MemoryOS, Reflexion~\citep{shinn2023reflexion}, Generative
Agents~\citep{park2023generativ}, SimpleMem (T/V)~\citep{simplemem2025,omnisimplemem2026},
MIRIX~\citep{wang2025mirix}, MMA~\citep{lu2026mma}, and
M2A~\citep{feng2026m2a}.

Textual memory methods operate on a captioned stream: each image-bearing round is converted once into a dense caption using \texttt{gpt-5.2}, and the memory method receives the original dialogue text plus these captions rather than the native image. Native multimodal methods receive the original visual inputs. Full-context methods receive the available history directly; retrieval-based methods retrieve a bounded set of memory rounds; agentic and structured memory methods use their method-specific memory-writing and querying procedures under the same benchmark runner.

For retrieval-based comparisons, we set top-$K{=}10$ unless a method's official interface requires an internal iteration budget. Text retrieval uses \texttt{all-MiniLM-L6-v2}. Semantic RAG (V) and M2A use \modelname{siglip2-base-patch16-384} for image embeddings, while MMA uses \modelname{google/siglip-so400m-patch14-384} to match its official implementation. For Semantic RAG (V), text and image dense similarities are combined with equal weight. M2A uses its official memory-manager and query-loop budgets, with up to
15 memory-manager iterations and up to 5 query iterations. MMA uses its default confidence-scoring weights for source, time, and consensus evidence. These settings keep the retrieval budget comparable while preserving method-specific mechanisms that are central to each memory architecture.

\paragraph{Embedding backbones.}
We use each retrieval-based method with the embedding backbone recommended by its original implementation when available. This choice reflects a full-system comparison: the retrieval encoder is part of the deployed memory method rather than a free interchangeable component. Forcing all methods to share a single embedding model can disadvantage methods whose memory writing, indexing, or confidence scoring is designed around a specific encoder. We therefore standardize embeddings where this is compatible with the method design, while preserving official backbones when they are integral to the method implementation. As a result, architecture-level comparisons should be interpreted as comparisons between full memory systems, not isolated encoder ablations.

\subsection{LLM-as-a-Judge Validation}
\label{app:judge-agreement}

To validate the LLM-as-a-Judge metric used for open-ended evaluation, we conduct a human-judge agreement study. We stratify-sample 3 questions per $(X,Y)$ cell and collect the open-ended predictions from two representative methods, FC(V) and SRAG(V), under \texttt{gpt-5.4-mini}. This produces 72 prediction-gold-answer pairs. The items are randomized and method labels are hidden to prevent annotator bias. One human annotator independently scores each prediction as Accept~(1) or Reject~(0), using the same correctness criterion as the automated judge: whether the prediction conveys the same information as the gold answer. We then compare the human labels with the binarized GPT-5.2 judge scores ($\geq 0.5 \to$ Accept, $< 0.5 \to$ Reject).

Agreement is 97.2\% (69/71 items, excluding 1 borderline case), with Cohen's $\kappa = 0.94$. Only two items produce disagreement, both involving the same navigation question where the gold answer describes a route by perceptual attributes while the model describes it by landmarks. The confusion matrix is nearly diagonal (Accept--Accept: 40; Reject-Reject: 29; off-diagonal: 1 each direction), and per-method agreement is balanced (FC(V): 97.2\%; SRAG(V): 97.1\%). These results indicate that the automated LLM-as-a-Judge scores closely track human accept/reject judgments on MemEye open-ended predictions.

\subsection{Prompt Examples}
\label{app:prompt-templates}

\begin{table}[t]
  \centering
  \small
  \caption{
  Key paired bootstrap confidence intervals on \texttt{gpt-5.4-mini}.
  Intervals are percentile 95\% confidence intervals over 10{,}000 question-level resamples with seed 20260430.
  Positive values indicate gains for the first term in each comparison.
  CI-L and CI-U denote the lower and upper confidence bounds.
  }
  \label{tab:bootstrap-ci-key-comparisons}
  \setlength{\tabcolsep}{2.6pt}
  \renewcommand{\arraystretch}{1.03}
  \begin{tabular}{@{}llrrrr@{}}
  \toprule
  Comparison & Slice & $n$ & Mean & CI-L & CI-U \\
  \midrule
  Cap.-Proof Judge & Low-$X$ V--T  & 91  & +0.024 & -0.035 & +0.081 \\
  Cap.-Proof Judge & High-$X$ V--T & 280 & +0.079 & +0.042 & +0.115 \\
  Cap.-Proof EM    & Low-$X$ V--T  & 91  & +0.010 & -0.037 & +0.058 \\
  Cap.-Proof EM    & High-$X$ V--T & 280 & +0.075 & +0.043 & +0.107 \\
  SRAG(V)--FC(V) Judge & All & 371 & +0.058 & +0.005 & +0.111 \\
  SRAG(V)--M2A Judge & $Y_3-Y_1$ & 113/60 & -0.425 & -0.586 & -0.265 \\
  \bottomrule
  \end{tabular}
\end{table}

This section reports the prompts used in our experiments. We group them by role: answer generation, caption generation, question generation, taxonomy annotation, and LLM-as-a-Judge evaluation. For LLM-as-a-Judge scoring, we follow the prompt-based evaluation protocol used by Mem-Gallery~\citep{bei2026mem}.

\newcommand{\promptbox}[2]{%
\begin{tcolorbox}[
  colback=gray!5,
  colframe=black,
  title=\textbf{#1},
  coltitle=white,
  fonttitle=\bfseries,
  colbacktitle=gray!90,
  enhanced,
  breakable,
  boxrule=0.7pt,
  arc=2pt,
  sharp corners=south,
  left=6pt,
  right=6pt,
  top=5pt,
  bottom=5pt,
  width=\linewidth,
  before skip=4pt,
  after skip=6pt
]
\footnotesize
\setlength{\parindent}{0pt}
\setlength{\parskip}{2pt}
#2
\end{tcolorbox}}

\promptbox{Multimodal Open-Ended Answering}{
You are an AI assistant being evaluated on MemEye, a vision-centric multimodal memory benchmark. Answer questions about a long-horizon multimodal conversation using the provided conversation history and images.

Rules: ground every answer in the conversation history and images; do not invent facts; when evidence conflicts, prefer the most recent visual evidence unless the question asks about the conflict; inspect images carefully; be precise about spatial positions, colors, textures, and visual attributes; say the information is absent if it is genuinely unavailable; give only the answer, with no reasoning or restatement. For counting questions, reply with only the number. For yes/no questions, start with ``Yes'' or ``No'' and add only essential corrective detail. For descriptive questions, answer in one or two short sentences.
}

\promptbox{Multimodal Multiple-Choice Answering}{
You are an AI assistant being evaluated on MemEye, a vision-centric multimodal memory benchmark. Answer multiple-choice questions about a long-horizon multimodal conversation using the provided conversation history and images.

Rules: ground every answer in the conversation history and images; do not invent facts; when evidence conflicts, prefer the most recent visual evidence unless the question asks about the conflict; inspect images carefully; reply with only the option letter, e.g., A, B, C, or D.
}

\promptbox{Text-Plus-Caption Open-Ended Answering}{
You are an AI assistant being evaluated on MemEye, a text-plus-caption memory benchmark setting. Answer questions about a long-horizon conversation using the provided conversation history and image captions only.

Rules: ground every answer in the provided conversation history and caption text; do not invent facts; do not assume access to raw image pixels; treat captions as the only visual evidence; if a required visual detail is not stated in the dialogue or captions, say the information is unavailable rather than guessing; be precise and concise; give only the answer. For counting questions, reply with only the number. For yes/no questions, start with ``Yes'' or ``No'' and add only essential corrective detail. For descriptive questions, answer in one or two short sentences.
}

\promptbox{Text-Plus-Caption Multiple-Choice Answering}{
You are an AI assistant being evaluated on MemEye, a text-plus-caption memory benchmark setting. Answer multiple-choice questions about a long-horizon conversation using the provided conversation history and image captions only.

Rules: ground every answer in the provided conversation history and caption text; do not invent facts; do not assume access to raw image pixels; if a visual detail is not stated in the dialogue or captions, choose only if the answer is supported by the provided text, otherwise make the best grounded choice from the evidence; reply with only the option letter, e.g., A, B, C, or D.
}

\promptbox{Caption Generation}{
Generate a detailed caption for this image. Include all visually observable information that may be useful for answering future memory questions: objects, people or characters, their identities or distinguishing attributes, colors, shapes, textures, positions, spatial relations, or changes. Be specific and exhaustive, but do not infer information that is not visible.
}

\promptbox{Question Generation}{
You are helping construct MemEye, a vision-centric benchmark for long-term multimodal agent memory. Generate candidate memory questions for the specified target taxonomy cell.

\emph{Inputs.} Target visual-evidence level: \texttt{\{\{target\_X\}\}}. Target memory-reasoning level: \texttt{\{\{target\_Y\}\}}. Task scenario: \texttt{\{\{task\_scenario\}\}}. Candidate dialogue rounds, timestamps, images, and available image descriptions: \texttt{\{\{evidence\_context\}\}}.

\emph{Taxonomy.} \(X_1\) requires scene-level visual evidence; \(X_2\) requires region-level spatial or local-area evidence; \(X_3\) requires identifying a specific object or person instance; \(X_4\) requires fine visual evidence such as exact color, small text, texture, count, or markings. \(Y_1\) requires one sufficient evidence unit; \(Y_2\) requires linking multiple non-conflicting clues; \(Y_3\) requires resolving updates, conflicts, replacements, disappearance/reappearance, or the latest valid state.

\emph{Requirements.} The question must require the target \(X\) and \(Y\) levels. It should be answerable from the provided evidence, but should not be answerable from general knowledge, answer-option priors, or dialogue text alone when the target requires visual information. For \(Y_3\), include at least one stale or superseded clue and one later clue that establishes the valid state. Avoid ambiguous wording, subjective judgments, and questions whose answer depends on information not visible or not stated in the evidence.

\emph{Output format.} Return a JSON object with fields \texttt{open\_question}, \texttt{ground\_truth}, \texttt{mcq\_question}, \texttt{options}, \texttt{correct\_option}, \texttt{clue\_rounds}, \texttt{target\_X}, \texttt{target\_Y}, and \texttt{explanation}. The explanation should state the decisive visual evidence, why the item fits the target \(X\) level, and why the clue structure fits the target \(Y\) level.
}

\promptbox{Taxonomy Agreement Annotation}{
You are an independent annotator for MemEye, a benchmark that labels each memory question by visual evidence granularity \(X\) and memory reasoning depth \(Y\). Assign one \(X\) label and one \(Y\) label to the item below.

\emph{Inputs.} Question: \texttt{\{\{question\}\}}. Ground-truth answer: \texttt{\{\{ground\_truth\}\}}. Annotator explanation: \texttt{\{\{explanation\}\}}.

\emph{Important instruction.} You do not have access to the original images. Use only the question, ground-truth answer, and annotator explanation. If the explanation describes visual evidence, treat it as the available description of what the human annotator saw. Do not solve the question; label the type of evidence and reasoning required to answer it.

\emph{Highest-bottleneck rule.} Choose the finest \(X\) level required by the decisive evidence, and the deepest \(Y\) operation required after the relevant evidence is available.

\emph{\(X\)-axis labels.} \(X_1\) Scene-level: the answer depends on global scene gist, activity, or overall context. \(X_2\) Region-level: the answer requires a local area, spatial relation, or grouped region, but not identity of one specific instance among similar candidates. \(X_3\) Instance-level: the answer requires identifying a specific object or person instance, especially among similar candidates or across images. \(X_4\) Pixel-level: the answer depends on fine visual details such as exact color, small text, texture, count, symbol, or OCR-like evidence.

\emph{\(Y\)-axis labels.} \(Y_1\) Atomic Retrieval: one evidence unit is sufficient once found. \(Y_2\) Relational Association: multiple non-conflicting clues must be linked across rounds, sessions, modalities, or references. \(Y_3\) Evolutionary Synthesis: evidence includes updates, conflicts, disappearance/reappearance, replacement, or overrides; the answer depends on the current valid state or on resolving a temporal change.

\emph{Boundary rules.} If the main challenge is WHERE something is in a local area, prefer \(X_2\). If the main challenge is WHICH specific instance is being referred to among similar objects or people, prefer \(X_3\). If the answer requires reading exact text, exact color, small markings, or other fine attributes, prefer \(X_4\). For \(Y\), use \(Y_3\) only when later evidence changes or overrides earlier evidence; otherwise use \(Y_2\) for non-conflicting multi-clue association and \(Y_1\) for single-clue retrieval.

\emph{Output format.} Return only a JSON object with fields \texttt{X}, \texttt{Y}, and \texttt{reasoning}. The labels must be one of \texttt{X1}, \texttt{X2}, \texttt{X3}, \texttt{X4} and one of \texttt{Y1}, \texttt{Y2}, \texttt{Y3}. Keep the reasoning to one or two sentences.
}

\promptbox{LLM-as-a-Judge Evaluation}{
You are an impartial judge evaluating the memory capabilities of an AI assistant on a question-answering task. Compare the Assistant's Answer against the Ground Truth and assign one score from \(\{0,0.25,0.5,0.75,1\}\).

\emph{Important principles.} Use semantic equivalence over surface form. Numeric and counting questions are binary: the exact number receives 1, and a wrong number receives 0. For negation-plus-correction questions, the answer must include both the correct polarity and the corrected fact. Identity questions are correct if the same unique person, object, or event is identified, even if the wording differs. Judge whether the answer gives the requested attribute or type of information. Wrong entities should not receive high scores just because they are on topic. Multi-item questions require matching the required set. Chronology and ordering questions require the correct ordered item. Do not over-reward fluent explanations. Opposite relations or orderings are normally scored 0.

\emph{Scoring rubric.} Score 0 for contradictions, wrong yes/no polarity, wrong numbers, wrong answer type, wrong relation or order, hallucination, or no relevant information. Score 0.25 for answers that touch the topic but miss the core entity or value, contain major wrong associations, are excessively vague, or get most of a required set wrong. Score 0.5 for answers that capture the main entity or concept but miss important qualifiers, or for multi-item answers that get a substantial part of the set right. Score 0.75 for largely accurate answers with only minor missing detail or unnecessary filler. Score 1 for accurate, precise answers that contain all core information without hallucinations; exact wording is not required.

\emph{Input fields.} Question: \texttt{\{\{question\}\}}. Ground Truth: \texttt{\{\{ground\_truth\}\}}. Assistant Answer: \texttt{\{\{model\_output\}\}}.

\emph{Output format.} Return only a JSON object with fields \texttt{score} and \texttt{reasoning}. The score must be one of \texttt{0}, \texttt{0.25}, \texttt{0.5}, \texttt{0.75}, or \texttt{1}; the reasoning should be short.
}

\FloatBarrier

\subsection{Bootstrap Confidence Intervals}
\label{app:bootstrap-ci}

We use nonparametric bootstrap confidence intervals to characterize uncertainty in the aggregate comparisons that support the main claims. Unless otherwise stated, each interval is computed with 10,000 bootstrap resamples and a fixed random seed. For each resample, we sample original questions with replacement and recompute the target aggregate metric. For multiple-choice evaluation, each original question contributes its rotation-averaged EM score; the four answer rotations are treated as within-question variants rather than independent samples. For open-ended evaluation, each original question contributes its BLEU-1 or LLM-Judge score.

For paired comparisons, such as native-visual versus caption-only streams or two memory methods evaluated on the same question set, each bootstrap resample is drawn at the question level and the paired score difference is recomputed on the sampled questions. This preserves the correlation induced by evaluating different methods on the same benchmark items. We report percentile 95\% confidence intervals. These intervals reflect sampling uncertainty over questions; they do not account for uncertainty from the judge model or judge prompt. We therefore use the intervals to assess the stability of aggregate trends, while treating individual small $(X,Y)$ cells as diagnostic localization rather than independent significance claims.

Table~\ref{tab:bootstrap-ci-key-comparisons} reports the key intervals used in the main analysis.

\subsection{Memory Method Implementations}
\label{app:method-implementation-details}

All methods are evaluated with the same benchmark runner. Non-agentic methods construct a method-specific context and pass it to the answer VLM. Agentic methods first ingest the full dialogue history into their memory store, then answer via the method-specific query interface. To maintain consistency, all internal VLM/LLM calls use the same backbone as the corresponding answering model, unless otherwise specified. Answer generation uses deterministic decoding with temperature $0$ and the benchmark answer model for the run, unless otherwise specified. 

We group methods into text-based and multimodal families. For text-based methods, images are replaced with dense captions generated by \texttt{gpt-5.2}. For multimodal methods, the original images are retained when supported by the method implementation. When official code is available and compatible with our benchmark runner, we use it with a lightweight adapter for MemEye evaluation formatting. Otherwise, we implement the method as described in the paper and report the relevant configuration below. All methods use the same benchmark prompts, decoding settings, and evaluation pipeline.

\paragraph{Full Context.}
FC(T) and FC(V) are full-history baselines. FC(T) concatenates the captioned dialogue history, while FC(V) passes the native multimodal history. When the estimated history exceeds the context budget, we apply FIFO (First-in-first-out) truncation. The context limit is 128k tokens, with reserved space for the question and answer.

\paragraph{Semantic RAG.}
SRAG(T) indexes each dialogue round with \texttt{all-MiniLM-L6-v2} text embeddings and retrieves the top-$K{=}10$ rounds. Image-bearing rounds are indexed using their captions. SRAG(V) uses the same text encoder and additionally embeds images with \texttt{siglip2-base-patch16-384}. Text and image similarities are combined with equal weights, and the selected top-$K{=}10$ rounds are passed to the answer VLM with native images.

\paragraph{A-Mem~\citep{xu2025amem}.}
A-Mem is evaluated as a text-only agentic memory system. Each dialogue round is converted into a memory note containing the dialog text and image caption lines. A-Mem uses \texttt{all-MiniLM-L6-v2} embeddings and retrieves $K{=}10$ related memories.

\paragraph{Reflexion~\citep{shinn2023reflexion}.}
Reflexion stores each captioned round as an observation in an RFMemory-style store. At query time, it recalls relevant memory text, answers using the shared benchmark prompt, and updates a compact global reflection. The recalled context is truncated to 6000 words. 

\paragraph{Generative Agents~\citep{park2023generativ}.}
Generative Agents stores captioned dialogue rounds as timestamped memories. Retrieval combines text similarity, recency, and importance using \texttt{all-MiniLM-L6-v2} embeddings, top-$K{=}10$ retrieval, a recency decay of 0.995, and a 4000-word recall budget. We enable the reflection mechanism with a threshold of 8.0, two reflection questions, and two generated insights. 

\paragraph{MemoryOS~\citep{kang2025memoryos}.}
MemoryOS is evaluated through its official short-, mid-, and long-term memory interface. Each captioned round is added as a timestamped user/assistant memory. Retrieval uses the internal MemoryOS context retriever with short-term capacity 10, mid-term capacity 2000, long-term knowledge capacity 100, retrieval queue capacity 10, mid-term heat threshold 5.0, and mid-term similarity threshold 0.6. The embedding model is \texttt{all-MiniLM-L6-v2}.

\paragraph{SimpleMem~\citep{simplemem2025,omnisimplemem2026}.}
SimpleMem(T) uses the Omni-SimpleMem orchestrator in caption-based mode. Each round is inserted as a text memory with session, round, image-id, and timestamp metadata. We use \texttt{all-MiniLM-L6-v2} embeddings, retrieve $K{=}20$ memories, and disable SimpleMem's self-evolution loop for benchmark fairness. SimpleMem(V) uses the same orchestrator in multimodal mode. Retrieved memory units retain raw image pointers and load images at answer time.

\paragraph{MIRIX~\citep{wang2025mirix}.}
MIRIX is evaluated through the official runtime adapter. The adapter stores user turns with native images and assistant turns without images, preserving per-turn timestamps. We use the official MIRIX agent wrapper. 

\paragraph{MMA~\citep{lu2026mma}.}
MMA is implemented as a confidence-aware multimodal memory system. It stores user and assistant turns as memory entries, attaches native images to user memories, computes text embeddings with \texttt{all-MiniLM-L6-v2}, and computes image embeddings with \texttt{google/siglip-so400m-patch14-384}. Retrieval ranks memories based on a blended text-image similarity score weighted by a confidence score. Confidence combines source credibility, temporal decay, and consensus with weights 0.45, 0.40, and 0.15, respectively, using a 30-day half-life. The retrieval budget is $K{=}10$.

\paragraph{M2A~\citep{feng2026m2a}.}
M2A is evaluated with its two-phase protocol. During ingestion, sessions are processed in temporal order, and the MemoryManager writes semantic memories from raw dialogue and image evidence. During answering, a fresh query agent retrieves semantic memories and expands their evidence identifiers back to source rounds. We use \texttt{all-MiniLM-L6-v2} for text embeddings and \texttt{siglip2-base-patch16-384} for multimodal embeddings. The memory-manager loop allows up to 15 iterations with a recent-context window of 5 turns, and the query loop allows up to 5 iterations. 

\section{Additional Results and Diagnostics}
\label{app:additional-results}
\subsection{Complete Result Matrices}
\label{app:full-main-matrix}

Table~\ref{tab:memeye_main_matrix_gpt_5_4_mini_full_appendix} reports the complete MemEye evaluation matrix for \texttt{gpt-5.4-mini}, including EM, BLEU-1, and LLM-as-a-Judge. We report additional backbone matrices below.

Tables~\ref{tab:memeye_main_matrix_gpt4_1_nano}, \ref{tab:memeye_main_matrix_qwen3vl8b}, and~\ref{tab:memeye_main_matrix_gemini_2_5_flash_lite} report the full MemEye evaluation matrices for \texttt{gpt-4.1-nano}, Qwen3-VL-8B-Instruct, and \texttt{gemini-2.5-flash-lite}, respectively, using the same coordinate-level reporting format as the main \texttt{gpt-5.4-mini} results.

\subsection{Caption Robustness Ablation}
\label{app:caption-robustness}

The Caption-Proof gap reported in \S\ref{sec:caption-proof-validation} uses generic dense captions generated by \texttt{gpt-5.2}. A natural concern is that the gap at high $X$ may reflect captioner weakness rather than a structural modality boundary. To test this, we regenerate captions with a stronger captioner (\texttt{gpt-5.4-mini}) and a task-aware prompt that explicitly targets OCR content, exact colors, instance identities, spatial relations, and fine-grained visual attributes. The prompt instructs the model to be exhaustive about details that a generic caption might omit, producing captions that are typically 2--3$\times$ longer than the originals.

We evaluate on a stratified subsample of 120 open-ended questions (40 low-$X$ from $X_1$--$X_2$, 80 high-$X$ from $X_3$-$X_4$), drawn by round-robin across all eight tasks. We run FC(T) and SRAG(T) with the task-aware captions and compare against both the original generic-caption runs and the visual-stream runs on the same questions.

Table~\ref{tab:caption-robustness} reports the results. Task-aware captions substantially improve textual-stream performance: SRAG(T) rises from $0.425$ to $0.595$ at low $X$ and from $0.235$ to $0.387$ at high $X$. Critically, the visual-stream gain shows a diminishing-returns pattern. At low $X$, the gap closes entirely: SRAG(T) with task-aware captions matches or exceeds the visual stream ($\Delta = -0.005$). At high $X$, the gap shrinks from $+0.194$ to $+0.041$ but remains positive; at $X_4$ it shrinks from $+0.215$ to $+0.094$. This pattern suggests that scene- and region-level evidence can often be recovered by a sufficiently detailed captioner, while instance- and pixel-level evidence---identity bindings, fine textures, small text, exact color attributes---remains harder to fully recover through textual mediation, even with a stronger task-aware captioning prompt.

\begin{table}[t]
  \centering
  \small
  \caption{Caption robustness ablation on \texttt{gpt-5.4-mini} (LLM-Judge, open-ended).
  Rows denote matched memory families. Native-image input reports the multimodal version,
  FC(V) or SRAG(V), while caption input reports the corresponding caption-based version,
  FC(T) or SRAG(T), using either generic or task-aware captions. $\Delta$ is the gain
  from native images over each caption condition. The stratified subsample contains
  40 low-$X$ and 80 high-$X$ questions across all eight tasks.}
  \label{tab:caption-robustness}
  \setlength{\tabcolsep}{3.5pt}
  \begin{tabular}{llccccc}
    \toprule
    \multirow{2}{*}{Method family} & \multirow{2}{*}{$X$ group} &
    \multirow{2}{*}{Native-image input} &
    \multicolumn{2}{c}{Caption input} &
    \multicolumn{2}{c}{$\Delta$ (Native image $-$ Caption)} \\
    \cmidrule(lr){4-5}\cmidrule(lr){6-7}
    & & & Generic & Task-aware & Generic & Task-aware \\
    \midrule
    FC    & Low-$X$  & 0.449 & 0.528 & 0.636 & $-0.079$ & $-0.188$ \\
    FC    & High-$X$ & 0.432 & 0.298 & 0.418 & $+0.134$ & $+0.014$ \\
    FC    & $X_4$    & 0.485 & 0.308 & 0.386 & $+0.177$ & $+0.100$ \\
    \midrule
    SRAG  & Low-$X$  & 0.590 & 0.425 & 0.595 & $+0.165$ & $-0.005$ \\
    SRAG  & High-$X$ & 0.428 & 0.235 & 0.387 & $+0.194$ & $+0.041$ \\
    SRAG  & $X_4$    & 0.419 & 0.203 & 0.325 & $+0.215$ & $+0.094$ \\
    \bottomrule
  \end{tabular}
\end{table}

\subsection{Effective Visual Information Analysis}
\label{app:effective-visual-information}

Figure~\ref{fig:caption-irreplaceability-bar} compares how much useful visual information remains under different input settings across LoCoMo~\citep{maharana2024evaluating}, MMRC~\citep{xue2025mmrc}, Mem-Gallery~\citep{bei2026mem}, and MemEye. The goal of this analysis is not to evaluate long-context retrieval, but to test whether the answer-relevant image information in each benchmark can be replaced by text. To isolate this factor, we use the gold clue rounds for each question whenever clue annotations are available. This removes memory search as the primary bottleneck and focuses the comparison on how much information native images contribute beyond textual context or captions.

For each benchmark, we evaluate the same answering model, \texttt{gpt-5.4-mini}, under three input settings. In the \textit{No Visual Info.} setting, the model receives the question and the gold textual clue context, but no images or image captions. In the \textit{Caption Only} setting, each image in the gold clue rounds is replaced by a dense textual caption. In the \textit{Multimodal} setting, the model receives the same gold clue rounds with the original images. The question text is kept fixed across all three settings, so differences in performance reflect the additional information provided by captions or native visual inputs.

For MemEye, we use the annotated clue rounds provided by the benchmark. For prior benchmarks, we construct comparable gold-clue contexts from their provided answer-relevant turns, evidence annotations, or, when available, a minimal supporting dialogue context. Captions are generated using \texttt{gpt-5.2} with the same dense captioning prompt used in our caption-based memory experiments. All answers are evaluated in the open-ended format using the same LLM-as-a-Judge protocol as the main experiments, and scores are multiplied by 100 for visualization.

This analysis measures visual irreplaceability. If caption-only performance approaches multimodal performance, then the benchmark's visually grounded questions are largely recoverable from textual substitutions. If multimodal performance substantially exceeds caption-only performance, then native image evidence contains information that captions do not preserve. As shown in Figure~\ref{fig:caption-irreplaceability-bar}, MemEye exhibits a larger gain from caption-only to multimodal input than the prior benchmarks, indicating that its questions contain more irreplaceable visual evidence.

\subsection{Retrieval Diagnostics}
\label{app:retrieval-diagnostics}

We use the annotated clue rounds to diagnose retrieval failures after a question is asked. The goal is not to build a separate retrieval leaderboard, but to separate three sources of error: a system may fail to retrieve relevant evidence, retrieve only part of the required evidence chain, or retrieve evidence that is semantically relevant but temporally stale. For retrieval-based methods, we compare the top-$K$ retrieved rounds with the gold clue rounds. \emph{Any-Clue Recall@K} measures whether at least one gold clue is retrieved. \emph{Coverage@K} measures the fraction of gold clues retrieved. \emph{Full-Clue Recall@K} measures whether the complete clue set is retrieved. For $Y_3$ questions, we additionally report \emph{Latest-Clue Recall@K}, which checks whether the final decisive clue is retrieved, and \emph{Stale-Dominance}, which measures whether stale evidence is ranked above the latest clue or appears without the latest clue.

Table~\ref{tab:retrieval-diagnostics-srag} shows that native visual retrieval improves evidence access, but does not solve evolving-state reasoning. SRAG(V) improves Any-Clue Recall@10 over SRAG(T), especially at $Y_1$ and $Y_3$. However, on $Y_3$, its Full-Clue Recall@10 is only 0.367 and its Latest-Clue Recall@10 is only 0.533. Thus, many failures are not caused by a complete absence of relevant evidence. Instead, the system often retrieves an incomplete update chain or misses the clue that establishes the current state. MMA retrieves $Y_3$ clue evidence slightly more often than SRAG(V), but under the broader source-session stale definition used by the recency probe, its Stale-Dominance remains high (0.750 versus 0.767 for SRAG(V)). M2A should be interpreted carefully because its provenance is expanded from semantic-memory evidence rather than direct raw-round retrieval. Overall, these diagnostics identify a specific failure mode: semantic relevance can surface the right topic while still selecting an outdated visual state.

\begin{table}[t]
  \centering
  \small
  \caption{Clue-round retrieval diagnostics on open-ended \texttt{gpt-5.4-mini} runs. Metrics compare the retrieved top-10 memory rounds with annotated gold clue rounds. The sample count $n$ reflects questions with valid retrieved-round provenance for this diagnostic and can differ slightly from axis-level dataset totals.}
  \label{tab:retrieval-diagnostics-srag}
  \setlength{\tabcolsep}{3.4pt}
  \begin{tabular}{llccccc}
    \toprule
    Method & $Y$ & $n$ & Any-Clue & Coverage & Full-Clue & Latest / Stale \\
    \midrule
    SRAG(T) & $Y_1$ & 113 & 0.832 & 0.819 & 0.805 & -- / -- \\
    SRAG(T) & $Y_2$ & 195 & 0.826 & 0.590 & 0.344 & -- / -- \\
    SRAG(T) & $Y_3$ & 60  & 0.667 & 0.510 & 0.367 & 0.517 / 0.526 \\
    \midrule
    SRAG(V) & $Y_1$ & 113 & 0.885 & 0.870 & 0.858 & -- / -- \\
    SRAG(V) & $Y_2$ & 195 & 0.826 & 0.622 & 0.410 & -- / -- \\
    SRAG(V) & $Y_3$ & 60  & 0.750 & 0.553 & 0.367 & 0.533 / 0.767 \\
    \midrule
    MMA & $Y_1$ & 113 & 0.841 & 0.824 & 0.805 & -- / -- \\
    MMA & $Y_2$ & 195 & 0.831 & 0.612 & 0.395 & -- / -- \\
    MMA & $Y_3$ & 60  & 0.833 & 0.583 & 0.383 & 0.550 / 0.750 \\
    \midrule
    M2A & $Y_1$ & 113 & 0.319 & 0.295 & 0.274 & -- / -- \\
    M2A & $Y_2$ & 195 & 0.477 & 0.234 & 0.051 & -- / -- \\
    M2A & $Y_3$ & 60  & 0.267 & 0.124 & 0.067 & 0.133 / 0.500 \\
    \bottomrule
  \end{tabular}
\end{table}

We next test whether stale-state selection can be reduced by a simple recency signal. Table~\ref{tab:recency-probe-alpha-sensitivity} keeps SRAG(V)'s retrieved candidate pool fixed and re-ranks candidates using a mixture of retrieval similarity and exponential recency:
\[
s_i = \alpha\,\mathrm{sim}_i + (1-\alpha)\exp(-\lambda \Delta t_i),
\]
where $\mathrm{sim}_i$ is the SigLIP2 retrieval similarity and $\Delta t_i$ is the age of candidate round $i$ relative to the query. We set $\lambda=0.02$. The age $\Delta t_i$ is measured as dialogue-round distance between candidate round $i$ and the query round, using the logged \texttt{recency\_age} field, with no additional normalization. This is a retrieval-side diagnostic probe rather than a new memory method: the candidate pool is unchanged, so recency can correct stale-over-latest ranking errors but cannot recover a latest clue that was not retrieved.

We decompose stale evidence selection into two cases. \emph{Rank-Inversion} measures how often both stale and latest evidence are retrieved but stale evidence is ranked higher. \emph{Latest-Miss} measures how often the latest decisive evidence is absent from the retrieved pool. The recency probe reduces Stale-Dominance and Rank-Inversion, showing that part of the $Y_3$ failure comes from ranking stale evidence above the valid update. However, Latest-Miss does not improve, because re-ranking cannot recover missing evidence. Table~\ref{tab:recency-answer-quality-sanity} reports answer-regeneration checks for $\alpha{=}0.7$ and $\alpha{=}0.5$. The $Y_3$ LLM-Judge point estimates improve, but the confidence intervals overlap zero, so the aggregate answer-quality gains are not reliable.

The two recency weights show the same trade-off. The $\alpha{=}0.7$ setting yields the larger $Y_3$ answer-quality point estimate ($+0.067$ Judge), while $\alpha{=}0.5$ yields the larger retrieval-side reduction in Stale-Dominance ($-0.183$ versus $-0.083$). Thus, recency is useful as a diagnostic signal for temporal-authority failures, but it is not a complete solution. A memory system must also retrieve the missing latest evidence and reason over the update chain once the evidence is available.

\begin{table}[t]
  \centering
  \small
  \caption{Recency counterfactual for SRAG(V) on $Y_3$ open-ended questions. Stale-Dominance uses the source-session stale definition: any retrieved round from a source session that occurs before the latest annotated clue is treated as stale evidence. Latest-Miss measures cases where the latest clue is absent from the top-10 candidate set; Rank-Inversion measures cases where the latest clue is retrieved but stale evidence ranks above it. Paired deltas are relative to vanilla SRAG(V), with percentile 95\% bootstrap CIs over 10{,}000 question-level resamples. Y3 Judge is reported only for settings with full answer regeneration.}
  \label{tab:recency-probe-alpha-sensitivity}
  \setlength{\tabcolsep}{4.0pt}
  \begin{tabular}{lccccc}
    \toprule
    Ranking & Latest@10 & Latest-Miss & Stale-Dom. & Rank-Inv. & Y3 Judge \\
    \midrule
    SRAG(V), $\alpha{=}1.0$ & 0.600 & 0.400 & 0.767 & 0.483 & 0.292 \\
    + Recency, $\alpha{=}0.7$ & 0.550 & 0.450 & 0.683 & 0.367 & 0.358 \\
    + Recency, $\alpha{=}0.5$ & 0.550 & 0.450 & 0.583 & 0.283 & 0.325 \\
    \midrule
    \multicolumn{6}{l}{\textit{Paired deltas relative to SRAG(V)}} \\
    \midrule
    Ranking & $\Delta$ Stale-Dom. & 95\% CI & $\Delta$ Rank-Inv. & 95\% CI & $\Delta$ Judge \\
    \midrule
    + Recency, $\alpha{=}0.7$ & $-0.083$ & [$-0.167$, $-0.017$] & $-0.117$ & [$-0.217$, $-0.033$] & $+0.067$ \\
    + Recency, $\alpha{=}0.5$ & $-0.183$ & [$-0.283$, $-0.100$] & $-0.200$ & [$-0.317$, $-0.083$] & $+0.033$ \\
    \bottomrule
  \end{tabular}
\end{table}

\begin{table}[t]
  \centering
  \small
  \caption{Answer-regeneration sanity check for the SRAG(V) recency diagnostic. Values are open-ended LLM-as-a-Judge scores. $\Delta$ reports paired differences relative to SRAG(V), with percentile 95\% bootstrap CIs over 10{,}000 question-level resamples.}
  \label{tab:recency-answer-quality-sanity}
  \setlength{\tabcolsep}{4.0pt}
  \begin{tabular}{llccc}
    \toprule
    Setting & Slice & Judge & $\Delta$ & 95\% CI \\
    \midrule
    SRAG(V) & $Y_1$ & 0.673 & -- & -- \\
    SRAG(V) & $Y_2$ & 0.449 & -- & -- \\
    SRAG(V) & $Y_3$ & 0.292 & -- & -- \\
    SRAG(V) & Avg. & 0.492 & -- & -- \\
    \midrule
    + Recency $\alpha{=}0.7$ & $Y_1$ & 0.677 & $+0.004$ & [$-0.058$, $+0.066$] \\
    + Recency $\alpha{=}0.7$ & $Y_2$ & 0.399 & $-0.051$ & [$-0.106$, $+0.005$] \\
    + Recency $\alpha{=}0.7$ & $Y_3$ & 0.358 & $+0.067$ & [$-0.042$, $+0.175$] \\
    + Recency $\alpha{=}0.7$ & Avg. & 0.477 & $-0.015$ & [$-0.054$, $+0.024$] \\
    \midrule
    + Recency $\alpha{=}0.5$ & $Y_1$ & 0.650 & $-0.022$ & [$-0.093$, $+0.049$] \\
    + Recency $\alpha{=}0.5$ & $Y_2$ & 0.391 & $-0.058$ & [$-0.129$, $+0.013$] \\
    + Recency $\alpha{=}0.5$ & $Y_3$ & 0.325 & $+0.033$ & [$-0.075$, $+0.142$] \\
    + Recency $\alpha{=}0.5$ & Avg. & 0.460 & $-0.032$ & [$-0.080$, $+0.015$] \\
    \bottomrule
  \end{tabular}
\end{table}

\subsection{Evolving Visual State Probe}
\label{app:evolving-visual-state}

The retrieval diagnostics above show that $Y_3$ failures involve stale evidence selection. We now isolate the visual form of this problem. In the evolving visual-state subset, the update is expressed in the images themselves rather than only in dialogue text. The questions require a sequence of visual clues in which later images update, conflict with, or override earlier images. To answer correctly, a memory system must recover the relevant visual evidence and determine which visual state remains valid.

We evaluate three controlled evidence settings. \emph{Latest-only} provides only the latest decisive visual clue and tests whether the current state is visually readable. \emph{Stale-only} provides only earlier visual clues and tests whether outdated evidence supports a plausible but stale answer. \emph{All-clue oracle} provides the full annotated visual evidence chain and tests whether the answering model can use both stale and updated clues when evidence selection is perfect. We compare these controls with representative multimodal memory systems under their normal retrieval or context behavior.

Table~\ref{tab:evolving-visual-state-probe} reports the LLM-as-a-Judge scores. The all-clue oracle obtains the highest score, showing that the questions become substantially more answerable when the full visual update chain is provided. Latest-only is slightly lower, as expected, because some questions ask about the change itself and therefore require earlier visual states for comparison. Stale-only is lower than both latest-only and all-clue oracle, indicating that older visual states are not harmless context; they can support outdated or incomplete answers. All memory systems remain far below the controlled evidence settings. This gap shows that current multimodal memory systems struggle not only to store images, but also to retrieve and prioritize the valid visual state when stale and updated visual evidence coexist.

\begin{figure*}[htbp]
\centering
\footnotesize

\setlength{\fboxsep}{3pt}
\setlength{\fboxrule}{0.4pt}

\newcommand{\panelrule}{\vspace{4pt}\hrule\vspace{5pt}}
\newcommand{\caseinnerwidth}{0.94\linewidth}

\newcommand{\caseheader}[4]{%
\parbox[t][0.76in][t]{\caseinnerwidth}{\centering
\textbf{#1}\\[1pt]
\textit{Task:} #2\\[1pt]
\textit{Cell:} #3\\[1pt]
#4
}%
}

\newcommand{\caseimageblock}[1]{%
\begin{minipage}[t][1.08in][t]{\caseinnerwidth}
\centering
#1
\end{minipage}
}

\newcommand{\caseqa}[1]{%
\parbox[t][0.72in][t]{\caseinnerwidth}{\scriptsize #1}%
}

\newcommand{\casegraybox}[1]{%
\colorbox{gray!10}{%
\parbox[t][1.35in][t]{0.91\linewidth}{\scriptsize #1}%
}%
}

\fbox{%
\begin{minipage}[t]{0.292\textwidth}
\centering
\vspace{4pt}

\caseheader
{Case 6: Silent Tag Override}
{CrossScene Memory}
{$(X_4,Y_3)$}
{FC-V\,=\,0.75 \quad RAG-V\,=\,0.00}

\panelrule

\caseimageblock{
\includegraphics[width=0.46\linewidth,height=0.48in,keepaspectratio]{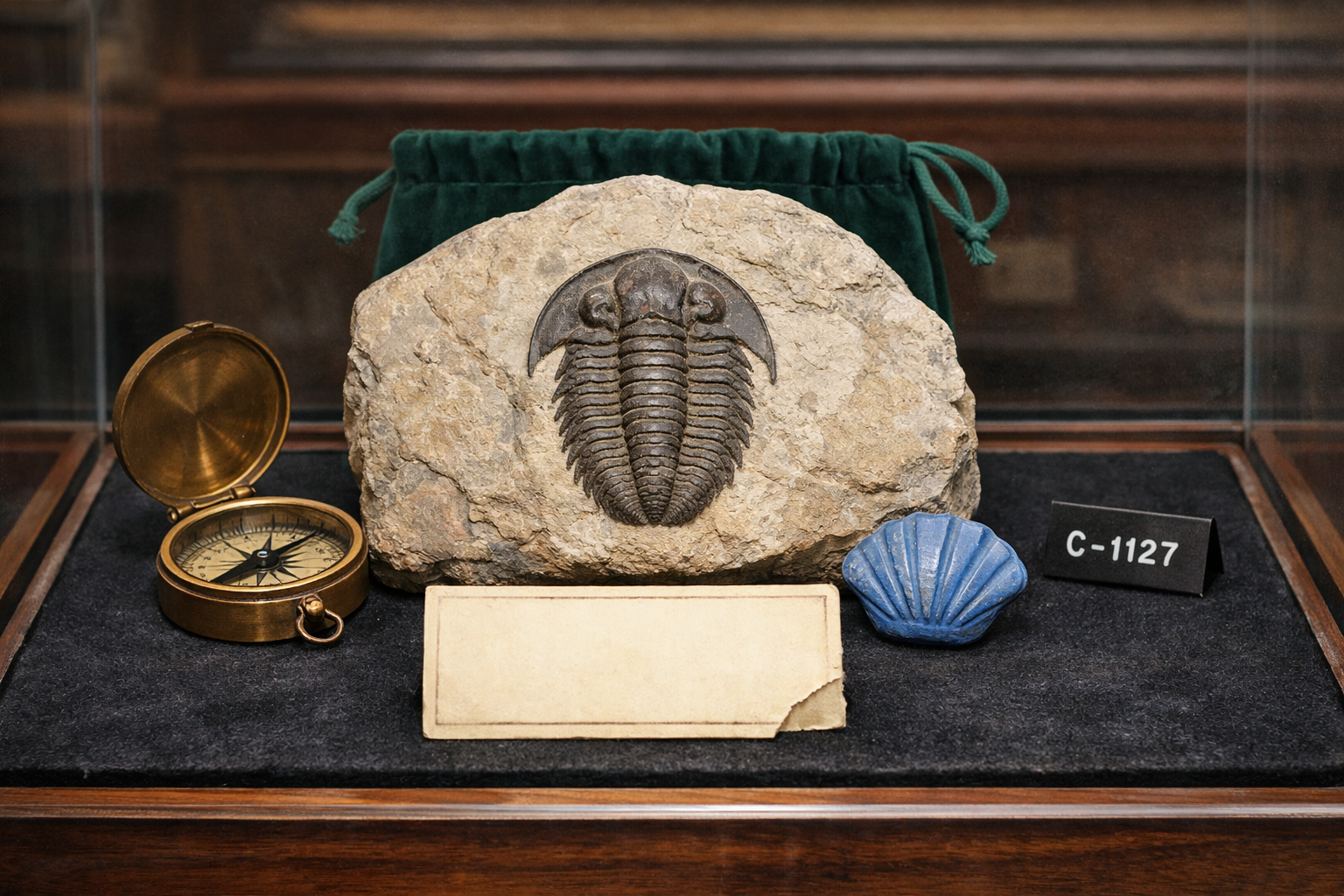}\hfill
\includegraphics[width=0.46\linewidth,height=0.48in,keepaspectratio]{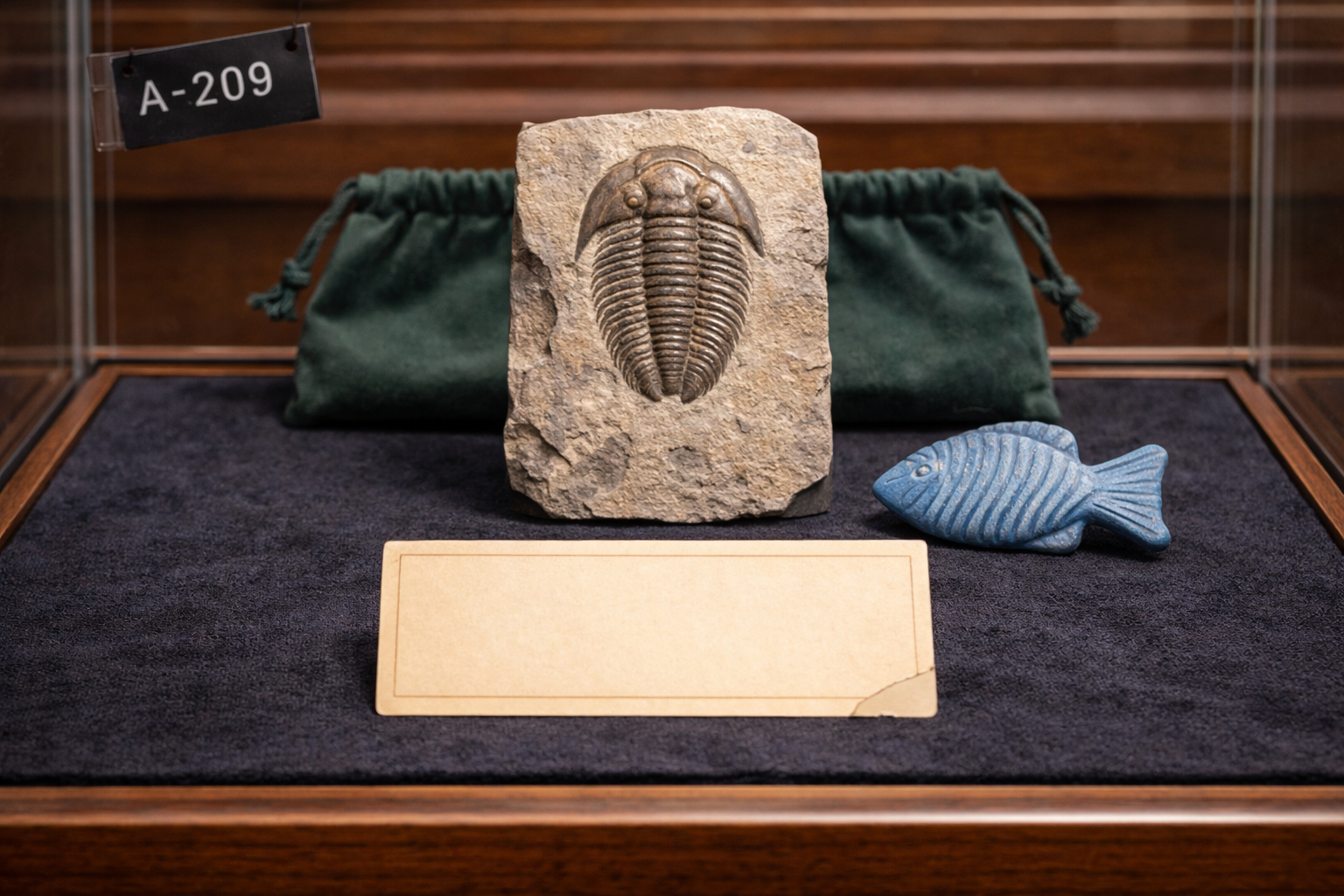}

\vspace{2pt}
\parbox[t]{0.46\linewidth}{\centering\scriptsize S7: Tag reads\\``C-1127''\\($\times 3$ images)}\hfill
\parbox[t]{0.46\linewidth}{\centering\scriptsize S9: Tag reads\\``A-209''\\($\times 1$ image)}
}

\panelrule

\caseqa{
\textbf{Q:} What identification tag number is currently displayed in the fossil room case?

\vspace{3pt}
\textbf{A:} A-209
}

\vspace{1pt}

\casegraybox{
\textbf{Why RAG fails:}\\
Retrieval for ``fossil room tag'' returns 4 images: 3 show ``C-1127'' (S7) and 1 shows ``A-209'' (S9). A frequency-voting retrieval system answers ``C-1127'' (3:1 majority). Only temporal reasoning---understanding that S9 is \textbf{later} than S7---yields the correct \textbf{current} tag ``A-209.''
}

\vspace{4pt}
\end{minipage}
}
\hfill
\fbox{%
\begin{minipage}[t]{0.292\textwidth}
\centering
\vspace{4pt}

\caseheader
{Case 7: Object Migration}
{CrossScene Memory}
{$(X_2,Y_3)$}
{FC-V\,=\,1.00 \quad RAG-V\,=\,0.25}

\panelrule

\caseimageblock{
\includegraphics[width=0.46\linewidth,height=0.48in,keepaspectratio]{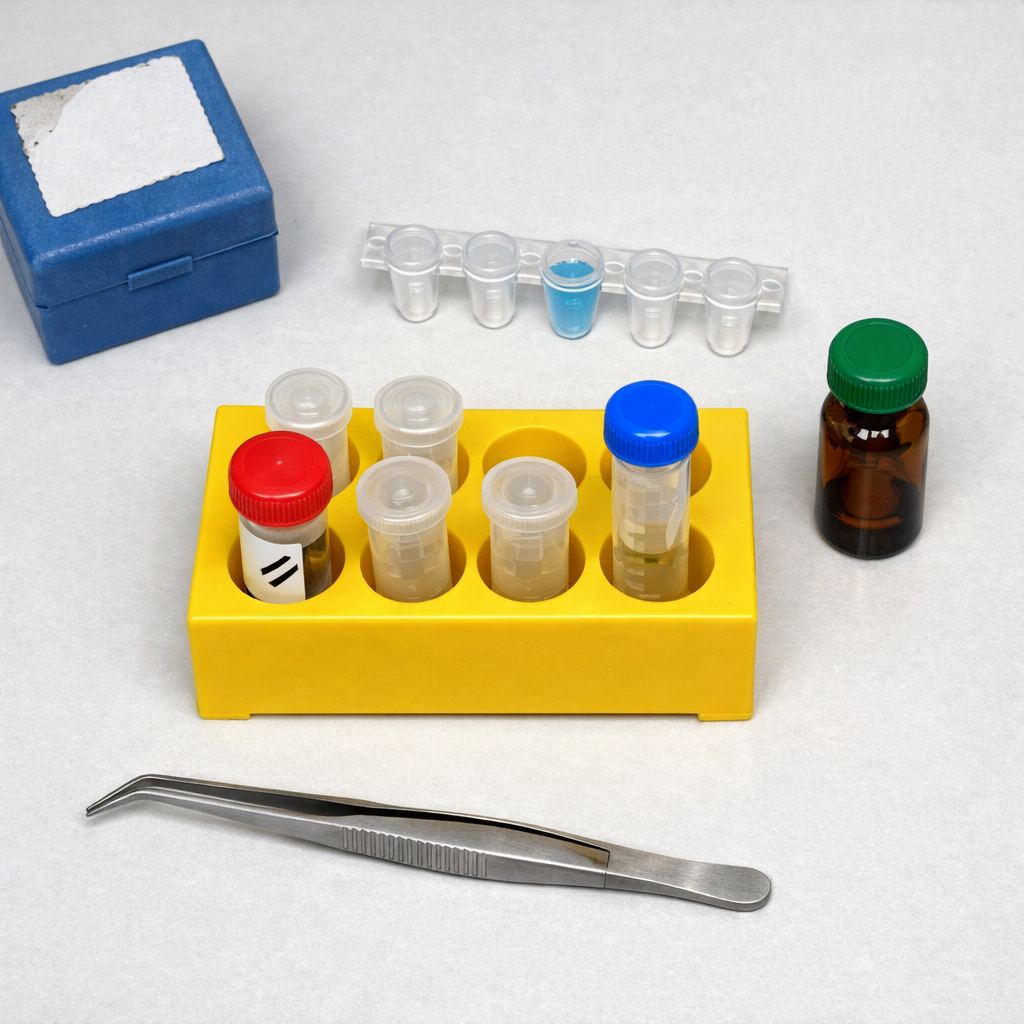}\hfill
\includegraphics[width=0.46\linewidth,height=0.48in,keepaspectratio]{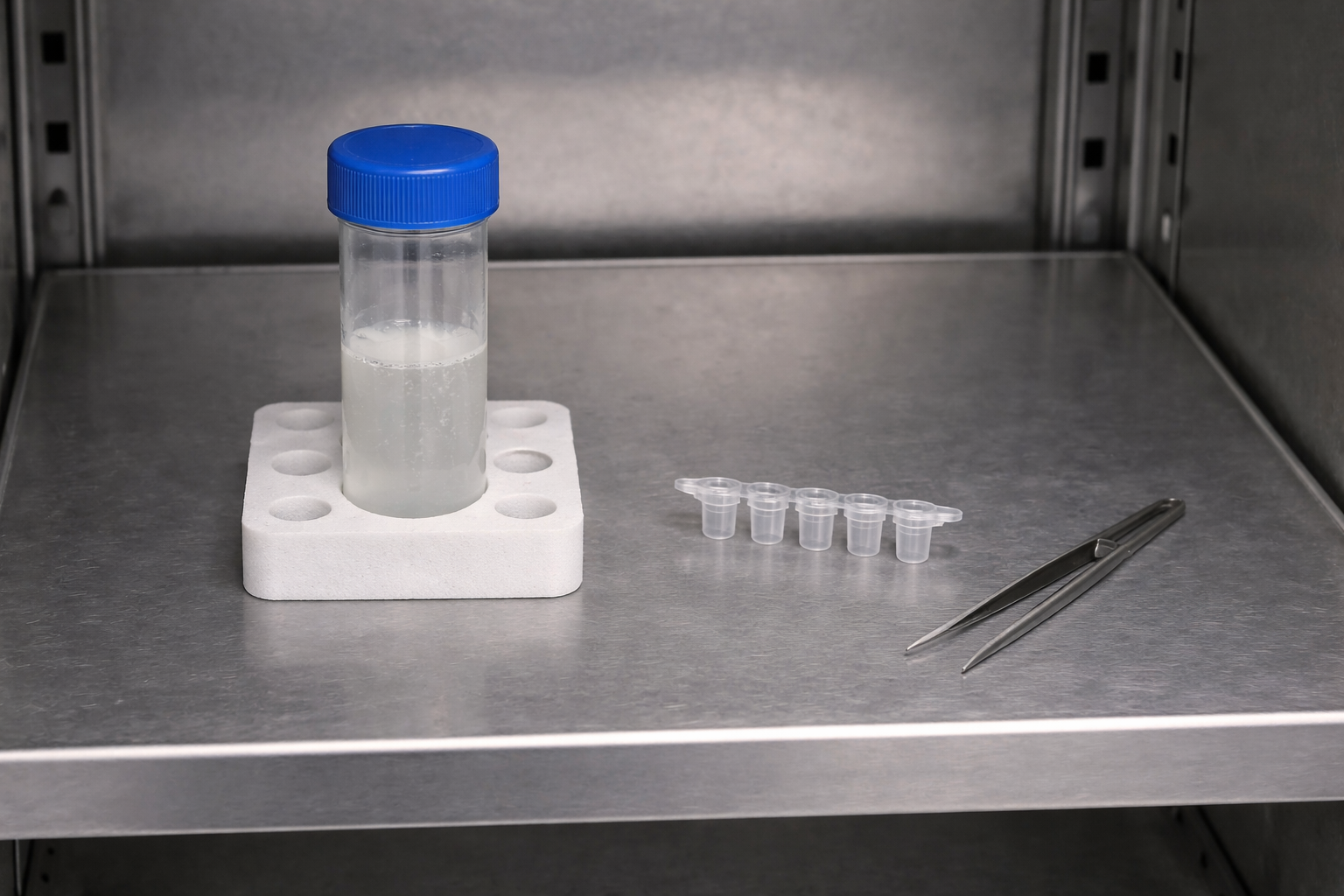}

\vspace{2pt}
\parbox[t]{0.46\linewidth}{\centering\scriptsize S10: Forceps on\\main bench}\hfill
\parbox[t]{0.46\linewidth}{\centering\scriptsize S11: Forceps on\\cold-room shelf}
}

\panelrule

\caseqa{
\textbf{Q:} Where are the bent-tip forceps that were originally on the main lab bench?

\vspace{3pt}
\textbf{A:} On the lower shelf of the cold-room prep area.
}

\vspace{1pt}

\casegraybox{
\textbf{Why RAG fails:}\\
Retrieval for ``bent-tip forceps'' returns images from both the original location (S10, main bench) and the current location (S11, cold-room shelf). Without temporal ordering, RAG cannot determine which is the \textbf{current} position. The question asks ``where are they,'' not ``where were they.''
}

\vspace{4pt}
\end{minipage}
}
\hfill
\fbox{%
\begin{minipage}[t]{0.292\textwidth}
\centering
\vspace{4pt}

\caseheader
{Case 8: Narrative Arc Tracking}
{Cartoon Ent.\ (Comic)}
{$(X_1,Y_3)$}
{FC-T\,=\,0.75 \quad RAG-T\,=\,0.25}

\panelrule

\caseimageblock{
\includegraphics[width=0.30\linewidth,height=0.48in,keepaspectratio]{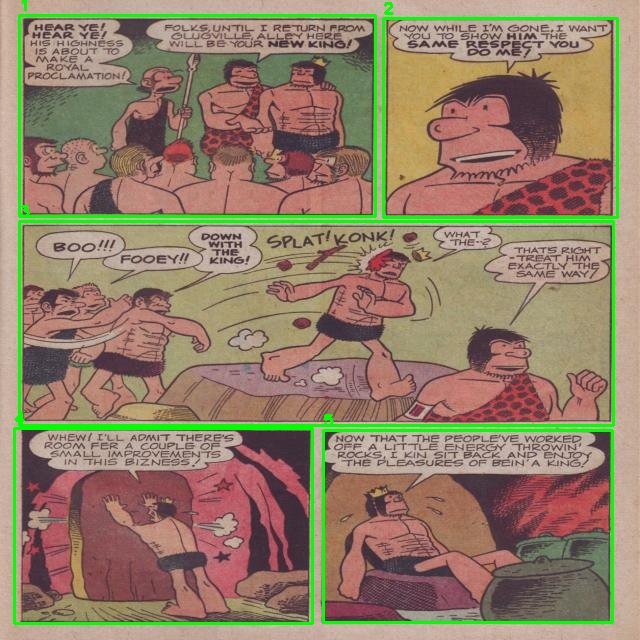}\hfill
\includegraphics[width=0.30\linewidth,height=0.48in,keepaspectratio]{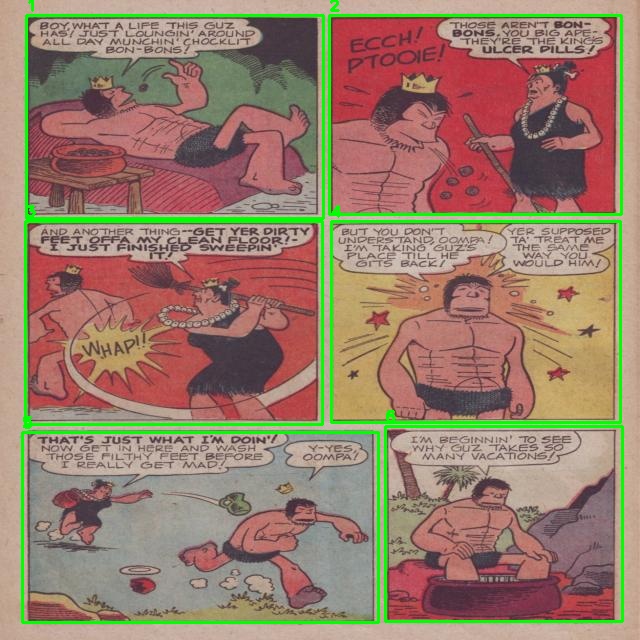}\hfill
\includegraphics[width=0.30\linewidth,height=0.48in,keepaspectratio]{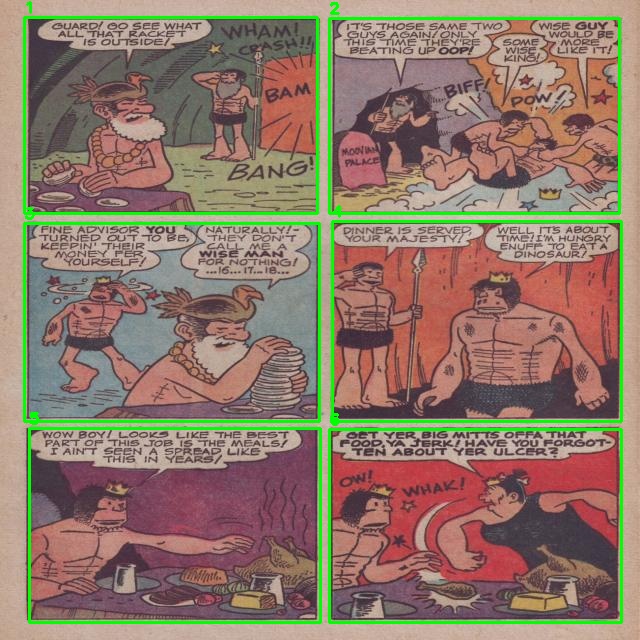}

\vspace{2pt}
\parbox[t]{0.30\linewidth}{\centering\scriptsize P28: Pelted\\by crowd}\hfill
\parbox[t]{0.30\linewidth}{\centering\scriptsize P29: Forced\\to do chores}\hfill
\parbox[t]{0.30\linewidth}{\centering\scriptsize P31: Angry\\outburst}
}

\panelrule

\caseqa{
\textbf{Q:} I feel like the temporary ruler's situation actually got better over time. Did it really get worse or better?

\vspace{3pt}
\textbf{A:} It got worse.
}

\vspace{1pt}

\casegraybox{
\textbf{Why RAG fails:}\\
Individual pages show mixed signals: P28 shows pelting \emph{then} relaxing, P29 shows chores, and P31 shows a steak dinner. A retrieval system may surface the ``relaxing'' or ``steak'' fragments and conclude improvement. Only tracking the \textbf{full temporal arc} across pages reveals a consistent downward trend.
}

\vspace{4pt}
\end{minipage}
}

\vspace{4pt}

\caption{
Three $Y_3$ cases where retrieval-based methods systematically fail and temporal reasoning is required.
\textbf{Case~6}: old evidence outnumbers new evidence 3:1---frequency voting picks the stale tag.
\textbf{Case~7}: the same object appears at two locations across sessions---only temporal ordering identifies the current one.
\textbf{Case~8}: the narrative arc contains local ups and downs---only tracking the full trajectory reveals the overall decline.
These cases illustrate why MemEye's $Y_3$ dimension is necessary: it separates models that can maintain a coherent, temporally ordered world model from those that treat memory as a static bag of retrieved fragments.
}
\label{fig:case-studies-3}
\end{figure*}

This failure mode is difficult to expose in benchmarks where updates are primarily textual. In MemEye, the key challenge is temporal authority over native visual evidence: the system must decide which image-grounded state is current. Figure~\ref{fig:case-studies-3} provides three case studies illustrating this behavior. The cases show that retrieval failures are not uniform: a method may miss the latest clue, over-weight stale evidence, or retrieve locally plausible fragments without tracking the full temporal chain.

\begin{table}[t]
  \centering
  \small
  \setlength{\tabcolsep}{6pt}
  \caption{Evolving visual-state probe on the subset of evolving visual state $Y_3$ questions under \texttt{gpt-5.4-mini}. Scores are open-ended LLM-as-a-Judge averages. Oracle settings use clue-only evidence with raw images. Memory-system settings use each method's original retrieval or context behavior.}
  \label{tab:evolving-visual-state-probe}
  \begin{tabular}{llc}
    \toprule
    Evaluation & Method & LLM-as-a-Judge \\
    \midrule
    \multirow{3}{*}{Oracle control}
      & Stale-only & 0.591 \\
      & Latest-only & 0.712 \\
      & All-clue oracle & 0.727 \\
    \midrule
    \multirow{4}{*}{Memory system}
      & FC(V) & 0.333 \\
      & SRAG(V) & 0.379 \\
      & MMA & 0.394 \\
      & M2A & 0.182 \\
    \bottomrule
  \end{tabular}
\end{table}

\subsection{Cross-topic Dialogue Scaling Ablation}
\label{app:mixed-memory-scaling}

Figure~\ref{fig:experimental-diagnostics}(d) and \S\ref{sec:retrieval-diagnostics} summarize the cross-topic dialogue scaling ablation in the main text. Here we describe the construction protocol and full trends. The goal is to test whether memory systems remain robust when answer-relevant evidence is embedded in a larger, more diverse history containing unrelated tasks.

We construct two controlled four-domain combinations. The first combines Brand Memory, Social Chat, Cartoon Entertainment, and Card Playlog; the second combines Health Care, Home Renovation, CrossScene Memory, and Outdoor Navigation. For each combination, we evaluate three memory scales over the same underlying questions. The 1-dataset point is the QA-weighted average over the four clean single-domain runs. The 2-dataset point is the QA-weighted average over two pairwise combinations. The 4-dataset point evaluates the full combined conversation history. This design keeps the answer-relevant evidence fixed while increasing cross-domain interference, allowing us to isolate whether a memory system can route the correct multimodal evidence from a noisy long-horizon history.

\begin{figure*}[t]
  \centering
  \includegraphics[width=\textwidth]{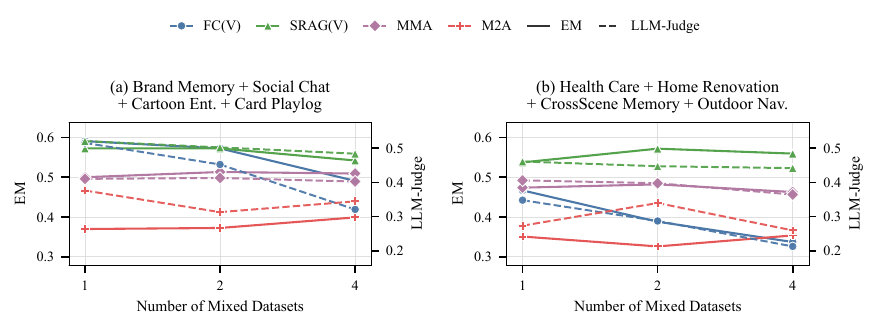}
  \caption{
  Cross-topic dialogue scaling ablation under \texttt{gpt-5.4-mini}. 
  Left: MCQ EM. Right: LLM-as-a-Judge. 
  Each curve evaluates the same questions as memory scale increases from clean single-domain histories, to pairwise cross-domain histories, to the full four-domain history. 
  Results are averaged over two controlled four-domain combinations.
  }
  \label{fig:mixed-memory-scaling-full}
\end{figure*}

Figure~\ref{fig:mixed-memory-scaling-full} shows the full scaling curves for both MCQ EM and open-ended LLM-as-a-Judge. The detailed trends support the main-text interpretation. Full Context (V) is sensitive to cross-topic interference: its performance drops as unrelated histories are added, especially in the Health Care, Home Renovation, CrossScene Memory, and Outdoor Navigation combination. This decline likely reflects two factors. First, the model must filter more irrelevant visual-textual evidence as the history grows. Second, long histories may approach the context window, increasing the risk that early answer-relevant evidence is truncated or diluted.

Retrieval-based and structured memory methods are more stable. Semantic RAG (V) remains comparatively robust across memory scales, suggesting that targeted retrieval reduces irrelevant context before answer generation. MMA also exhibits flatter trends, indicating that structured multimodal memory can reduce interference. M2A shows partial resilience in open-ended evaluation, suggesting that agentic operations such as memory writing, conflict checking, or iterative querying can help manage noisy histories. However, its lower absolute scores indicate that abstraction must still preserve fine-grained visual evidence to be effective.

Overall, this ablation shows that scaling multimodal memory is not simply a matter of increasing the context window. Long-term multimodal agents need evidence-routing mechanisms that filter unrelated history while preserving the visual details and state information needed for the current question.

\subsection{Limitations and Broader Impacts}
\label{app:limitations}

\sys is intended as a diagnostic benchmark rather than a complete simulation of all multimodal agent deployments. Its scenarios, model panel, captioning pipeline, and human validation sample may not cover every real-world setting. Comparisons across memory architectures can also be affected by method-specific encoders and implementations, so we interpret the results as diagnostic evidence about design trade-offs rather than as a universal ranking of memory systems. Finally, the human oracle check in Appendix~\ref{app:human-sanity-check} is a small sanity check rather than a full human ceiling estimate.

Evaluating long-term visual memory also has broader implications. More reliable memory can reduce stale or incorrect visual-state reasoning, but stronger visual memory systems may increase privacy risks when agents store user images, personal environments, or evolving user states. Future multimodal memory systems should therefore combine evidence-preserving memory with consent, data minimization, deletion, and access-control mechanisms.

\begin{table*}[p]
  \centering
  \caption{
  Main results on the MemEye evaluation matrix using \texttt{gpt-5.4-mini}. Columns correspond to memory methods, grouped into text-only and multimodal families. Within each coordinate, EM is reported for multiple-choice questions, while BLEU-1 and LLM-as-a-Judge (LLM-Judge) is reported for free-response questions. The first- and second-performing memory model(s) are highlighted with orange and blue backgrounds, respectively.}
  \label{tab:memeye_main_matrix_gpt_5_4_mini_full_appendix}

  \setlength{\tabcolsep}{3.0pt}
  \renewcommand{\arraystretch}{1.05}

  \resizebox{\textwidth}{!}{%
  \begin{tabular}{lllccccccccccccc}
  \toprule
  \multirow{2}{*}{Y} & \multirow{2}{*}{X} & \multirow{2}{*}{Metric}
  & \multicolumn{7}{c}{Textual memory}
  & \multicolumn{6}{c}{Multimodal memory} \\
  \cmidrule(lr){4-10}\cmidrule(lr){11-16}
  & &
  & FC(T) & SRAG(T) & Refl. & Gen.Ag. & MemOS & A-Mem & SM(T)
  & FC(V) & SRAG(V) & MIRIX & MMA & M2A & SM(V) \\
  \midrule

  \multirow{12}{*}{Y1}
  & \multirow{3}{*}{X1} & EM   & 0.8000 & \secondcell{0.9500} & 0.6750 & 0.2500 & 0.7750 & 0.7750 & 0.8000 & \bestcell{1.0000} & 0.9000 & 0.6750 & 0.5500 & 0.4750 & 0.8500 \\
  & & BLEU-1   & 0.3291 & 0.2654 & 0.2270 & 0.1914 & 0.1596 & 0.1312 & 0.2926 & \bestcell{0.3867} & \secondcell{0.3814} & 0.1178 & 0.2539 & 0.1439 & 0.2922 \\
  & & LLM-Judge   & \bestcell{0.6500} & 0.4500 & \secondcell{0.6000} & 0.3000 & 0.4000 & 0.2500 & 0.5500 & \bestcell{0.6500} & \secondcell{0.6000} & 0.4000 & 0.4500 & 0.4500 & 0.5000 \\
  \cmidrule(lr){2-16}
  & \multirow{3}{*}{X2} & EM   & \secondcell{0.7500} & 0.5455 & 0.2045 & 0.2273 & 0.4545 & 0.4318 & 0.5000 & 0.6818 & \bestcell{0.7727} & 0.5227 & 0.5455 & 0.2500 & 0.4773 \\
  & & BLEU-1   & 0.2282 & 0.2819 & 0.0894 & 0.0829 & 0.3242 & 0.3333 & 0.1220 & 0.2909 & \bestcell{0.6076} & 0.2144 & \secondcell{0.3471} & 0.1738 & 0.0766 \\
  & & LLM-Judge   & 0.4545 & 0.4091 & 0.1364 & 0.1818 & 0.4091 & 0.3182 & 0.2727 & 0.5455 & \bestcell{0.9091} & 0.5000 & \secondcell{0.7273} & 0.3182 & 0.1818 \\
  \cmidrule(lr){2-16}
  & \multirow{3}{*}{X3} & EM   & 0.4662 & 0.4527 & 0.2534 & 0.2500 & 0.3649 & 0.4392 & 0.3209 & \secondcell{0.5709} & \bestcell{0.6554} & 0.5473 & 0.5507 & 0.3615 & 0.3784 \\
  & & BLEU-1   & 0.2692 & 0.2675 & 0.2196 & 0.1685 & 0.2547 & 0.3142 & 0.1109 & \secondcell{0.3597} & \bestcell{0.3918} & 0.1810 & 0.3554 & 0.1889 & 0.1532 \\
  & & LLM-Judge   & 0.3716 & 0.3176 & 0.3108 & 0.2230 & 0.3176 & 0.4459 & 0.2230 & 0.5338 & \bestcell{0.6554} & 0.2568 & \secondcell{0.5946} & 0.3311 & 0.2838 \\
  \cmidrule(lr){2-16}
  & \multirow{3}{*}{X4} & EM   & 0.4722 & 0.5694 & 0.4444 & 0.2361 & \secondcell{0.6389} & 0.5972 & 0.4583 & 0.4722 & \bestcell{0.8056} & 0.3750 & 0.6250 & 0.5000 & 0.5000 \\
  & & BLEU-1   & 0.2861 & 0.2907 & 0.0722 & 0.0327 & 0.2240 & 0.1852 & 0.1139 & 0.1676 & \bestcell{0.3423} & 0.1321 & \secondcell{0.3237} & 0.0658 & 0.1139 \\
  & & LLM-Judge   & \secondcell{0.3889} & 0.3333 & 0.1111 & 0.0556 & 0.2222 & 0.2500 & 0.2222 & 0.3333 & \bestcell{0.6389} & 0.3056 & \bestcell{0.6389} & 0.1667 & 0.2222 \\
  \midrule

  \multirow{12}{*}{Y2}
  & \multirow{3}{*}{X1} & EM   & 0.5086 & 0.5172 & 0.4483 & 0.2500 & 0.4569 & 0.4052 & 0.5000 & \secondcell{0.5345} & 0.5086 & 0.2845 & 0.5086 & 0.3190 & \bestcell{0.5690} \\
  & & BLEU-1   & 0.3851 & \secondcell{0.5148} & 0.4369 & 0.2195 & 0.3098 & 0.3104 & 0.4071 & 0.4787 & \bestcell{0.5661} & 0.3748 & 0.3101 & 0.0848 & 0.3381 \\
  & & LLM-Judge   & 0.4483 & 0.5345 & \secondcell{0.5517} & 0.2759 & 0.3793 & 0.4138 & 0.4828 & 0.5000 & \bestcell{0.6552} & 0.4138 & 0.3621 & 0.2931 & 0.4138 \\
  \cmidrule(lr){2-16}
  & \multirow{3}{*}{X2} & EM   & \bestcell{0.4881} & 0.3810 & 0.1905 & 0.2619 & 0.3333 & 0.3690 & 0.3095 & 0.3810 & 0.2976 & 0.3214 & \secondcell{0.4762} & 0.3333 & 0.3452 \\
  & & BLEU-1   & 0.1208 & 0.1045 & 0.1404 & 0.0469 & 0.0689 & 0.1045 & 0.1254 & 0.0996 & \bestcell{0.1590} & 0.0749 & \secondcell{0.1562} & 0.0989 & 0.0702 \\
  & & LLM-Judge   & 0.1667 & 0.0952 & 0.1667 & 0.0238 & 0.1429 & 0.0952 & 0.1429 & 0.1667 & \bestcell{0.2143} & 0.1429 & \secondcell{0.1905} & \bestcell{0.2143} & 0.0952 \\
  \cmidrule(lr){2-16}
  & \multirow{3}{*}{X3} & EM   & 0.4417 & 0.4125 & 0.3833 & 0.2292 & 0.3333 & 0.3917 & 0.3917 & \secondcell{0.5750} & \bestcell{0.6250} & 0.4583 & 0.4292 & 0.3792 & 0.3708 \\
  & & BLEU-1   & \secondcell{0.2018} & 0.1800 & 0.1255 & 0.1368 & 0.1582 & 0.1759 & 0.0879 & 0.1694 & \bestcell{0.2532} & 0.1259 & 0.1863 & 0.1174 & 0.1322 \\
  & & LLM-Judge   & 0.3750 & 0.2833 & 0.2000 & 0.1167 & 0.2833 & 0.2917 & 0.2333 & \secondcell{0.3917} & \bestcell{0.6000} & 0.2667 & 0.3750 & 0.3583 & 0.2667 \\
  \cmidrule(lr){2-16}
  & \multirow{3}{*}{X4} & EM   & 0.3438 & 0.3523 & 0.2841 & 0.2472 & 0.3409 & 0.3551 & 0.3097 & 0.3636 & \secondcell{0.3722} & 0.3665 & \bestcell{0.4119} & 0.2955 & 0.2869 \\
  & & BLEU-1   & 0.3487 & 0.2786 & 0.3381 & 0.2161 & \bestcell{0.3904} & 0.2305 & 0.2321 & \secondcell{0.3496} & 0.2737 & 0.2827 & 0.2425 & 0.1082 & 0.1923 \\
  & & LLM-Judge   & 0.3807 & 0.3182 & 0.3466 & 0.2614 & \bestcell{0.4205} & 0.3636 & 0.2500 & \secondcell{0.3977} & 0.3352 & 0.3352 & 0.2898 & 0.3352 & 0.2159 \\
  \midrule

  \multirow{12}{*}{Y3}
  & \multirow{3}{*}{X1} & EM   & \secondcell{0.8000} & 0.7500 & 0.7000 & 0.2500 & 0.7000 & 0.4750 & 0.5000 & \bestcell{0.9000} & 0.5750 & 0.6750 & 0.5500 & 0.6000 & 0.5500 \\
  & & BLEU-1   & 0.2037 & 0.2190 & \bestcell{0.3152} & 0.1832 & 0.2106 & 0.2107 & 0.2212 & \secondcell{0.3097} & 0.1251 & 0.2675 & 0.2791 & 0.1311 & 0.2323 \\
  & & LLM-Judge   & \secondcell{0.6500} & 0.6000 & \bestcell{0.7000} & 0.4500 & 0.6000 & 0.5000 & 0.6000 & \secondcell{0.6500} & 0.4500 & 0.6000 & 0.5500 & 0.6000 & 0.6000 \\
  \cmidrule(lr){2-16}
  & \multirow{3}{*}{X2} & EM   & \secondcell{0.7000} & \secondcell{0.7000} & 0.5500 & 0.2500 & 0.4750 & 0.6000 & 0.6500 & 0.6000 & \bestcell{0.7750} & 0.4750 & 0.6750 & 0.4750 & \secondcell{0.7000} \\
  & & BLEU-1   & 0.2582 & 0.2936 & \secondcell{0.2961} & 0.2592 & 0.2065 & \bestcell{0.3395} & 0.2492 & 0.2097 & 0.1978 & 0.1045 & 0.1819 & 0.1126 & 0.2502 \\
  & & LLM-Judge   & 0.4000 & \bestcell{0.5000} & \secondcell{0.4500} & \secondcell{0.4500} & 0.3000 & \secondcell{0.4500} & 0.3000 & 0.2500 & 0.3500 & 0.2000 & 0.3000 & 0.1000 & 0.3000 \\
  \cmidrule(lr){2-16}
  & \multirow{3}{*}{X3} & EM   & 0.6000 & 0.6250 & 0.5750 & 0.2750 & 0.6250 & 0.5750 & 0.5250 & 0.5750 & \secondcell{0.6500} & \secondcell{0.6500} & \bestcell{0.8000} & 0.5000 & 0.6000 \\
  & & BLEU-1   & 0.1259 & 0.1468 & 0.1152 & \secondcell{0.2134} & 0.1995 & \bestcell{0.2275} & 0.0523 & 0.1380 & 0.1078 & 0.1365 & 0.1738 & 0.1783 & 0.1485 \\
  & & LLM-Judge   & \bestcell{0.5500} & \bestcell{0.5500} & \bestcell{0.5500} & 0.4000 & 0.3500 & \bestcell{0.5500} & 0.3500 & 0.4000 & 0.3000 & 0.4000 & \secondcell{0.4500} & \bestcell{0.5500} & 0.3500 \\
  \cmidrule(lr){2-16}
  & \multirow{3}{*}{X4} & EM   & 0.4333 & 0.3250 & 0.2833 & 0.2333 & 0.3417 & 0.3417 & 0.3167 & \bestcell{0.5917} & \secondcell{0.4750} & 0.2583 & 0.3417 & 0.3250 & 0.3083 \\
  & & BLEU-1   & 0.2122 & 0.2630 & 0.2463 & 0.2407 & 0.1096 & 0.1302 & 0.1969 & \bestcell{0.3495} & 0.1613 & 0.1477 & \secondcell{0.2708} & 0.0690 & 0.2136 \\
  & & LLM-Judge   & 0.3000 & 0.3000 & 0.2833 & \secondcell{0.3167} & 0.1667 & 0.3000 & 0.2000 & \bestcell{0.4500} & 0.2167 & 0.1667 & 0.2667 & 0.3000 & 0.2000 \\
  \midrule
  \multirow{3}{*}{Avg.} & \multirow{3}{*}{--} & EM   & 0.5670 & 0.5484 & 0.4160 & 0.2467 & 0.4866 & 0.4797 & 0.4651 & \secondcell{0.6038} & \bestcell{0.6177} & 0.4674 & 0.5386 & 0.4011 & 0.4947 \\
  & & BLEU-1   & 0.2474 & 0.2588 & 0.2185 & 0.1659 & 0.2180 & 0.2244 & 0.1843 & \secondcell{0.2758} & \bestcell{0.2972} & 0.1800 & 0.2567 & 0.1227 & 0.1844 \\
  & & LLM-Judge   & 0.4280 & 0.3909 & 0.3672 & 0.2546 & 0.3326 & 0.3524 & 0.3189 & \secondcell{0.4391} & \bestcell{0.4937} & 0.3323 & 0.4329 & 0.3347 & 0.3025 \\
  \bottomrule
  \end{tabular}%
  }
\end{table*}

\begin{table*}[p]
  \centering
  \caption{
  Main results on the MemEye evaluation matrix using \texttt{gpt-4.1-nano}. Columns correspond to memory methods, grouped into text-only and multimodal families. Within each coordinate, EM is reported for multiple-choice questions, while BLEU-1 and LLM-as-a-Judge (LLM-Judge) are reported for free-response questions. The first- and second-performing memory model(s) are highlighted with orange and blue backgrounds, respectively.}
  \label{tab:memeye_main_matrix_gpt4_1_nano}

  \setlength{\tabcolsep}{3.0pt}
  \renewcommand{\arraystretch}{1.05}

  \resizebox{\textwidth}{!}{%
  \begin{tabular}{lllccccccccccccc}
  \toprule
  \multirow{2}{*}{Y} & \multirow{2}{*}{X} & \multirow{2}{*}{Metric}
  & \multicolumn{7}{c}{Textual memory}
  & \multicolumn{6}{c}{Multimodal memory} \\
  \cmidrule(lr){4-10}\cmidrule(lr){11-16}
  & &
  & FC(T) & SRAG(T) & Refl. & Gen.Ag. & MemOS & A-Mem & SM(T)
  & FC(V) & SRAG(V) & MIRIX & MMA & M2A & SM(V) \\
  \midrule

  \multirow{12}{*}{Y1} & \multirow{3}{*}{X1} & EM   & 0.4750 & \secondcell{0.7500} & \secondcell{0.7500} & 0.2500 & 0.7250 & 0.7000 & 0.6750 & 0.6750 & \bestcell{0.7750} & 0.4750 & 0.6000 & 0.3500 & \secondcell{0.7500} \\
  & & BLEU-1   & 0.2377 & 0.1875 & \bestcell{0.2991} & 0.1168 & 0.1358 & 0.1540 & 0.2008 & 0.1523 & 0.2506 & 0.0594 & 0.1333 & 0.0467 & \secondcell{0.2579} \\
  & & LLM-Judge   & 0.4500 & 0.3500 & \secondcell{0.5500} & 0.2500 & 0.4000 & 0.4500 & \secondcell{0.5500} & 0.4500 & \bestcell{0.6500} & 0.2000 & 0.3000 & 0.1000 & 0.5000 \\
  \cmidrule(lr){2-16}
  & \multirow{3}{*}{X2} & EM   & 0.4444 & 0.4722 & 0.4167 & 0.2222 & 0.4167 & 0.5556 & \secondcell{0.5833} & 0.3333 & 0.2778 & 0.3056 & 0.2500 & 0.5556 & \bestcell{0.6111} \\
  & & BLEU-1   & 0.2100 & 0.1838 & \bestcell{0.2874} & 0.1421 & 0.1295 & 0.1011 & 0.0428 & 0.0921 & \secondcell{0.2364} & 0.0431 & 0.0960 & 0.0657 & 0.0314 \\
  & & LLM-Judge   & 0.3333 & 0.3333 & 0.3333 & 0.3333 & 0.3333 & 0.1111 & 0.3333 & \secondcell{0.3889} & \bestcell{0.6111} & \secondcell{0.3889} & \secondcell{0.3889} & 0.1111 & 0.3333 \\
  \cmidrule(lr){2-16}
  & \multirow{3}{*}{X3} & EM   & 0.3125 & 0.2961 & 0.3388 & 0.2599 & 0.2961 & 0.2632 & 0.2796 & 0.3947 & \bestcell{0.4342} & \secondcell{0.4079} & \secondcell{0.4079} & 0.2796 & 0.2697 \\
  & & BLEU-1   & \secondcell{0.2341} & 0.2007 & 0.2106 & 0.1328 & 0.2069 & 0.1869 & 0.1644 & 0.2275 & \bestcell{0.2480} & 0.0781 & 0.1598 & 0.0724 & 0.2023 \\
  & & LLM-Judge   & 0.2763 & 0.2303 & 0.2895 & 0.2237 & 0.2632 & 0.2039 & 0.2039 & 0.3882 & \bestcell{0.5263} & 0.1645 & \secondcell{0.4868} & 0.1118 & 0.2237 \\
  \cmidrule(lr){2-16}
  & \multirow{3}{*}{X4} & EM   & 0.4167 & 0.5139 & 0.4167 & 0.2778 & \secondcell{0.5694} & 0.5556 & 0.3611 & 0.4861 & \bestcell{0.5972} & 0.5556 & 0.5556 & 0.3472 & 0.3611 \\
  & & BLEU-1   & 0.1283 & \bestcell{0.2408} & 0.1022 & 0.0148 & \secondcell{0.2018} & 0.1963 & 0.1013 & 0.1032 & 0.1259 & 0.0197 & 0.0608 & 0.0356 & 0.0694 \\
  & & LLM-Judge   & 0.2778 & \bestcell{0.4444} & 0.1667 & 0.0000 & 0.2778 & 0.3056 & 0.1111 & 0.2222 & \secondcell{0.3611} & 0.0000 & \secondcell{0.3611} & 0.1389 & 0.1667 \\
  \midrule

  \multirow{12}{*}{Y2} & \multirow{3}{*}{X1} & EM   & 0.3534 & \secondcell{0.3966} & 0.3534 & 0.2414 & 0.3017 & 0.3190 & 0.2931 & \bestcell{0.4052} & \bestcell{0.4052} & 0.2069 & 0.3879 & 0.3276 & 0.3448 \\
  & & BLEU-1   & 0.1246 & 0.2092 & 0.2119 & 0.0503 & \secondcell{0.2144} & 0.1232 & 0.1600 & 0.0815 & 0.1181 & 0.0151 & 0.0778 & 0.0229 & \bestcell{0.2617} \\
  & & LLM-Judge   & 0.3621 & \bestcell{0.5000} & \secondcell{0.4310} & 0.2586 & 0.4138 & 0.3621 & 0.2241 & \bestcell{0.5000} & 0.3621 & 0.1034 & 0.3621 & 0.1034 & 0.3276 \\
  \cmidrule(lr){2-16}
  & \multirow{3}{*}{X2} & EM   & 0.3910 & 0.2949 & 0.3718 & 0.2500 & 0.2692 & 0.3077 & 0.4295 & 0.3782 & \secondcell{0.4551} & 0.3910 & 0.3782 & 0.3462 & \bestcell{0.4744} \\
  & & BLEU-1   & \secondcell{0.1521} & 0.1421 & 0.1426 & 0.1129 & 0.1396 & 0.0947 & 0.1192 & 0.1400 & \bestcell{0.1599} & 0.0785 & 0.1261 & 0.0988 & 0.0875 \\
  & & LLM-Judge   & \secondcell{0.3205} & 0.2436 & 0.1923 & 0.2436 & 0.1282 & 0.1667 & 0.2179 & 0.2308 & \bestcell{0.4103} & 0.1795 & 0.2308 & 0.1538 & 0.2051 \\
  \cmidrule(lr){2-16}
  & \multirow{3}{*}{X3} & EM   & 0.3512 & 0.3393 & 0.3155 & 0.2381 & 0.3571 & 0.2917 & 0.3095 & 0.3631 & \bestcell{0.4048} & \secondcell{0.3988} & 0.3512 & 0.2976 & 0.2976 \\
  & & BLEU-1   & 0.1368 & 0.0871 & 0.1347 & 0.0993 & \secondcell{0.1715} & \bestcell{0.1715} & 0.0843 & 0.1448 & 0.1676 & 0.0420 & 0.0945 & 0.0534 & 0.1322 \\
  & & LLM-Judge   & 0.2500 & 0.1905 & 0.1429 & 0.1429 & 0.2262 & 0.2381 & 0.1667 & \secondcell{0.3571} & \bestcell{0.4048} & 0.0833 & 0.2143 & 0.0833 & 0.1667 \\
  \cmidrule(lr){2-16}
  & \multirow{3}{*}{X4} & EM   & 0.2784 & 0.2756 & 0.2642 & 0.2642 & 0.2699 & 0.2386 & 0.2756 & 0.2585 & \secondcell{0.3324} & 0.2699 & \bestcell{0.3438} & 0.3125 & 0.3182 \\
  & & BLEU-1   & 0.2114 & 0.1705 & 0.1580 & 0.0760 & 0.1988 & 0.1542 & \secondcell{0.2391} & 0.1742 & 0.1233 & 0.0345 & 0.0670 & 0.0566 & \bestcell{0.2955} \\
  & & LLM-Judge   & 0.3295 & 0.3125 & 0.2784 & \bestcell{0.3580} & \bestcell{0.3580} & 0.3295 & 0.2330 & 0.2386 & \secondcell{0.3409} & 0.0966 & 0.2955 & 0.2727 & 0.3182 \\
  \midrule

  \multirow{12}{*}{Y3} & \multirow{3}{*}{X1} & EM   & \secondcell{0.5750} & \secondcell{0.5750} & 0.5000 & 0.2500 & \bestcell{0.6500} & 0.4500 & 0.4250 & \secondcell{0.5750} & 0.4500 & 0.5000 & 0.3500 & 0.3250 & 0.4750 \\
  & & BLEU-1   & 0.2579 & 0.1544 & \secondcell{0.2705} & 0.1493 & 0.1269 & \bestcell{0.3162} & 0.1423 & 0.1519 & 0.1052 & 0.0295 & 0.1243 & 0.0487 & 0.1541 \\
  & & LLM-Judge   & \secondcell{0.5500} & 0.4000 & \bestcell{0.6000} & 0.4000 & 0.2500 & 0.4500 & 0.4500 & 0.4500 & 0.3000 & 0.1000 & 0.4500 & 0.1000 & 0.3500 \\
  \cmidrule(lr){2-16}
  & \multirow{3}{*}{X2} & EM   & 0.6000 & \bestcell{0.7000} & 0.5000 & 0.2500 & 0.4500 & 0.4500 & \secondcell{0.6250} & 0.4500 & 0.4000 & 0.5000 & 0.2500 & 0.3500 & 0.6000 \\
  & & BLEU-1   & \secondcell{0.3175} & 0.3053 & 0.2436 & 0.2180 & 0.1376 & \bestcell{0.3470} & 0.2635 & 0.2366 & 0.1936 & 0.1935 & 0.1309 & 0.0866 & 0.2903 \\
  & & LLM-Judge   & \bestcell{0.8000} & 0.7000 & 0.6000 & \secondcell{0.7500} & 0.2500 & 0.7000 & 0.6500 & 0.5000 & 0.5000 & 0.3000 & 0.3000 & 0.2000 & 0.6500 \\
  \cmidrule(lr){2-16}
  & \multirow{3}{*}{X3} & EM   & 0.4250 & 0.5500 & 0.4500 & 0.3000 & 0.5500 & 0.4750 & 0.5500 & 0.4750 & 0.4250 & \bestcell{0.7000} & \secondcell{0.6000} & 0.3750 & 0.5750 \\
  & & BLEU-1   & 0.1182 & 0.1960 & \secondcell{0.1965} & 0.1255 & 0.1773 & 0.1177 & 0.1497 & 0.1105 & \bestcell{0.2074} & 0.0580 & 0.1585 & 0.0809 & 0.1409 \\
  & & LLM-Judge   & \secondcell{0.3000} & 0.2500 & 0.2500 & 0.2000 & 0.2000 & 0.2500 & 0.1500 & 0.2500 & \bestcell{0.4000} & 0.0000 & 0.2500 & 0.2000 & 0.2000 \\
  \cmidrule(lr){2-16}
  & \multirow{3}{*}{X4} & EM   & 0.1250 & 0.1833 & 0.1083 & \bestcell{0.2417} & \secondcell{0.2333} & 0.0833 & 0.1917 & 0.1417 & 0.1667 & 0.2000 & 0.2000 & 0.2250 & 0.1667 \\
  & & BLEU-1   & 0.0579 & \bestcell{0.1514} & 0.0607 & 0.0429 & 0.0612 & 0.0845 & 0.0888 & \secondcell{0.1125} & 0.0528 & 0.0281 & 0.0416 & 0.0294 & 0.1009 \\
  & & LLM-Judge   & \bestcell{0.3333} & \secondcell{0.3167} & 0.2333 & 0.1667 & 0.2167 & 0.2667 & 0.1000 & 0.2333 & 0.2000 & 0.0167 & 0.2000 & 0.0833 & 0.1000 \\
  \midrule
  \multirow{3}{*}{Avg.} & \multirow{3}{*}{--} & EM   & 0.3956 & \bestcell{0.4456} & 0.3988 & 0.2538 & 0.4240 & 0.3908 & 0.4165 & 0.4113 & 0.4269 & 0.4092 & 0.3895 & 0.3409 & \secondcell{0.4370} \\
  & & BLEU-1   & 0.1822 & \secondcell{0.1857} & \bestcell{0.1931} & 0.1067 & 0.1585 & 0.1706 & 0.1463 & 0.1439 & 0.1657 & 0.0566 & 0.1059 & 0.0582 & 0.1687 \\
  & & LLM-Judge   & \secondcell{0.3819} & 0.3559 & 0.3390 & 0.2772 & 0.2764 & 0.3195 & 0.2825 & 0.3508 & \bestcell{0.4222} & 0.1361 & 0.3200 & 0.1382 & 0.2951 \\
  \bottomrule
  \end{tabular}%
  }
\end{table*}

\begin{table*}[p]
  \centering
  \caption{
  Main results on the MemEye evaluation matrix using Qwen3-VL-8B-Instruct. Columns correspond to memory methods, grouped into text-only and multimodal families. Within each coordinate, EM is reported for multiple-choice questions, while BLEU-1 and LLM-as-a-Judge (LLM-Judge) are reported for free-response questions. The first- and second-performing memory model(s) are highlighted with orange and blue backgrounds, respectively.}
  \label{tab:memeye_main_matrix_qwen3vl8b}

  \setlength{\tabcolsep}{3.0pt}
  \renewcommand{\arraystretch}{1.05}

  \resizebox{\textwidth}{!}{%
  \begin{tabular}{lllcccccccccccc}
  \toprule
  \multirow{2}{*}{Y} & \multirow{2}{*}{X} & \multirow{2}{*}{Metric}
  & \multicolumn{7}{c}{Textual memory}
  & \multicolumn{5}{c}{Multimodal memory} \\
  \cmidrule(lr){4-10}\cmidrule(lr){11-15}
  & &
  & FC(T) & SRAG(T) & Refl. & Gen.Ag. & MemOS & A-Mem & SM(T)
  & FC(V) & SRAG(V) & MMA & M2A & SM(V) \\
  \midrule

  \multirow{12}{*}{Y1} & \multirow{3}{*}{X1} & EM   & \secondcell{0.8500} & \secondcell{0.8500} & \bestcell{0.9250} & 0.2500 & 0.7750 & 0.8000 & 0.5250 & 0.5750 & 0.5000 & 0.5500 & 0.5250 & 0.5500 \\
  & & BLEU-1   & \secondcell{0.2658} & 0.1958 & 0.1976 & 0.1131 & 0.2471 & 0.1537 & 0.2425 & 0.1177 & 0.1093 & 0.1022 & 0.0874 & \bestcell{0.2925} \\
  & & LLM-Judge   & 0.4500 & \bestcell{0.5500} & 0.4500 & 0.2500 & \bestcell{0.5500} & 0.4000 & \secondcell{0.5000} & 0.3000 & 0.3500 & 0.1500 & 0.1500 & 0.4500 \\
  \cmidrule(lr){2-15}
  & \multirow{3}{*}{X2} & EM   & \bestcell{0.7500} & 0.5278 & \secondcell{0.6389} & 0.2500 & 0.5278 & 0.5833 & 0.6111 & 0.3056 & 0.4444 & 0.2222 & 0.5278 & 0.6111 \\
  & & BLEU-1   & \secondcell{0.1561} & 0.1123 & 0.1471 & 0.0736 & 0.0874 & \bestcell{0.2119} & 0.0150 & 0.1544 & 0.1123 & 0.0069 & 0.0712 & 0.0150 \\
  & & LLM-Judge   & \bestcell{0.6111} & \secondcell{0.3333} & \secondcell{0.3333} & 0.2778 & 0.1111 & \bestcell{0.6111} & 0.2222 & 0.2222 & 0.2778 & 0.0000 & 0.1111 & 0.2222 \\
  \cmidrule(lr){2-15}
  & \multirow{3}{*}{X3} & EM   & \bestcell{0.4901} & \bestcell{0.4901} & \secondcell{0.4605} & 0.2599 & 0.4243 & 0.4507 & 0.1118 & 0.3026 & 0.3586 & 0.3783 & 0.3717 & 0.1118 \\
  & & BLEU-1   & \bestcell{0.2823} & 0.2570 & \secondcell{0.2737} & 0.1893 & 0.1690 & 0.1753 & 0.0589 & 0.1792 & 0.2719 & 0.2050 & 0.1192 & 0.0688 \\
  & & LLM-Judge   & 0.3092 & \secondcell{0.3158} & \bestcell{0.3487} & 0.2303 & 0.2171 & 0.2632 & 0.1053 & 0.2500 & 0.2632 & 0.2237 & 0.2039 & 0.1118 \\
  \cmidrule(lr){2-15}
  & \multirow{3}{*}{X4} & EM   & 0.5278 & \bestcell{0.6111} & 0.5278 & 0.2500 & 0.5556 & \secondcell{0.5833} & 0.2639 & 0.4167 & 0.4306 & 0.2083 & 0.2917 & 0.2639 \\
  & & BLEU-1   & \secondcell{0.2051} & \bestcell{0.2097} & 0.1046 & 0.0273 & 0.1525 & 0.1721 & 0.0556 & 0.0779 & 0.1771 & 0.1378 & 0.0377 & 0.0556 \\
  & & LLM-Judge   & \bestcell{0.3889} & \secondcell{0.3611} & 0.2778 & 0.0833 & 0.2222 & 0.2778 & 0.1111 & 0.3333 & 0.3056 & 0.2222 & 0.0833 & 0.1111 \\
  \midrule

  \multirow{12}{*}{Y2} & \multirow{3}{*}{X1} & EM   & 0.4138 & \secondcell{0.4655} & 0.4310 & 0.2414 & 0.3966 & 0.4052 & 0.3621 & 0.2241 & 0.4397 & \bestcell{0.4741} & 0.3103 & 0.3707 \\
  & & BLEU-1   & 0.2811 & 0.3167 & \secondcell{0.3317} & 0.1716 & 0.1968 & 0.2836 & 0.2634 & 0.1179 & \bestcell{0.4219} & 0.2108 & 0.0737 & 0.1939 \\
  & & LLM-Judge   & 0.3621 & 0.3621 & \secondcell{0.4310} & 0.2414 & 0.2414 & 0.3793 & 0.3276 & 0.2069 & \bestcell{0.4828} & 0.2931 & 0.1034 & 0.2414 \\
  \cmidrule(lr){2-15}
  & \multirow{3}{*}{X2} & EM   & \bestcell{0.4679} & 0.3397 & \secondcell{0.4359} & 0.2564 & 0.3654 & 0.4038 & 0.3526 & 0.1923 & 0.4103 & 0.4167 & 0.4295 & 0.3397 \\
  & & BLEU-1   & \secondcell{0.2243} & 0.1091 & \bestcell{0.2341} & 0.2144 & 0.0576 & 0.0785 & 0.0807 & 0.1542 & 0.1995 & 0.1163 & 0.0909 & 0.0765 \\
  & & LLM-Judge   & \bestcell{0.3590} & 0.2179 & \secondcell{0.3205} & 0.2692 & 0.1538 & 0.1410 & 0.1538 & 0.1026 & 0.1795 & 0.1154 & 0.0513 & 0.1410 \\
  \cmidrule(lr){2-15}
  & \multirow{3}{*}{X3} & EM   & \secondcell{0.3690} & 0.3452 & 0.3095 & 0.2500 & 0.3393 & 0.3155 & 0.1964 & 0.2024 & 0.2976 & 0.2679 & \bestcell{0.3750} & 0.2262 \\
  & & BLEU-1   & 0.1373 & 0.1174 & 0.0967 & 0.1155 & \bestcell{0.1423} & 0.1303 & 0.0432 & 0.1012 & \secondcell{0.1376} & 0.0303 & 0.0933 & 0.0457 \\
  & & LLM-Judge   & \bestcell{0.2500} & 0.2143 & 0.1548 & 0.1310 & \secondcell{0.2381} & 0.1786 & 0.0952 & 0.1310 & 0.2024 & 0.0000 & 0.2143 & 0.1310 \\
  \cmidrule(lr){2-15}
  & \multirow{3}{*}{X4} & EM   & 0.3807 & 0.3693 & 0.3494 & 0.2500 & \secondcell{0.3892} & \bestcell{0.4034} & 0.1506 & 0.2756 & 0.3352 & 0.2869 & 0.3040 & 0.1307 \\
  & & BLEU-1   & \bestcell{0.2880} & 0.2265 & 0.2654 & \secondcell{0.2712} & 0.2153 & 0.1769 & 0.0795 & 0.2097 & 0.1744 & 0.1898 & 0.0476 & 0.1003 \\
  & & LLM-Judge   & \bestcell{0.3239} & 0.2443 & \secondcell{0.3011} & 0.2955 & 0.2216 & 0.2102 & 0.1023 & 0.2557 & 0.2102 & 0.2159 & 0.1250 & 0.1364 \\
  \midrule

  \multirow{12}{*}{Y3} & \multirow{3}{*}{X1} & EM   & 0.5750 & 0.5250 & 0.5250 & 0.2500 & \bestcell{0.7500} & \secondcell{0.7250} & 0.5250 & 0.2750 & 0.2250 & 0.2750 & 0.4250 & 0.4750 \\
  & & BLEU-1   & 0.1623 & 0.1427 & 0.1667 & 0.0836 & 0.1283 & 0.2122 & \bestcell{0.2515} & 0.0805 & 0.0408 & 0.0264 & 0.0428 & \secondcell{0.2413} \\
  & & LLM-Judge   & \secondcell{0.5000} & \secondcell{0.5000} & \secondcell{0.5000} & 0.4000 & \bestcell{0.5500} & 0.4500 & \bestcell{0.5500} & 0.2500 & 0.1000 & 0.0000 & 0.0000 & \bestcell{0.5500} \\
  \cmidrule(lr){2-15}
  & \multirow{3}{*}{X2} & EM   & \secondcell{0.8000} & 0.7250 & \bestcell{0.8250} & 0.2250 & 0.7000 & 0.6000 & 0.7000 & 0.3500 & 0.5000 & 0.2500 & 0.2500 & 0.7250 \\
  & & BLEU-1   & 0.2543 & 0.2227 & 0.2147 & 0.1895 & 0.2091 & \bestcell{0.3594} & 0.2910 & 0.2249 & 0.2182 & 0.0367 & 0.0843 & \secondcell{0.2939} \\
  & & LLM-Judge   & \bestcell{0.8000} & \secondcell{0.6000} & \bestcell{0.8000} & 0.5500 & 0.4000 & 0.5500 & \secondcell{0.6000} & 0.4000 & 0.2500 & 0.0000 & 0.0000 & \secondcell{0.6000} \\
  \cmidrule(lr){2-15}
  & \multirow{3}{*}{X3} & EM   & \bestcell{0.5500} & \secondcell{0.5250} & 0.4750 & 0.2500 & 0.4000 & 0.4000 & 0.3000 & 0.3250 & 0.4500 & 0.3750 & 0.4750 & 0.2750 \\
  & & BLEU-1   & 0.1523 & 0.0956 & \bestcell{0.1541} & 0.0979 & 0.1053 & \secondcell{0.1540} & 0.0111 & 0.0764 & 0.1096 & 0.1086 & 0.1054 & 0.0091 \\
  & & LLM-Judge   & \bestcell{0.3000} & 0.1000 & \secondcell{0.2500} & 0.0000 & \bestcell{0.3000} & \secondcell{0.2500} & 0.0000 & 0.0500 & 0.2000 & 0.0500 & 0.0500 & 0.0000 \\
  \cmidrule(lr){2-15}
  & \multirow{3}{*}{X4} & EM   & \secondcell{0.2583} & 0.1667 & 0.2083 & 0.2500 & 0.2250 & 0.1917 & 0.0250 & 0.1250 & 0.1833 & 0.2500 & \bestcell{0.3417} & 0.0417 \\
  & & BLEU-1   & \secondcell{0.1967} & 0.1245 & \bestcell{0.2085} & 0.1504 & 0.0603 & 0.0896 & 0.1191 & 0.0048 & 0.1757 & 0.0770 & 0.0413 & 0.1191 \\
  & & LLM-Judge   & 0.1667 & 0.1333 & \secondcell{0.2000} & \bestcell{0.2167} & 0.0833 & 0.1167 & 0.1000 & 0.0333 & \secondcell{0.2000} & 0.0667 & 0.1333 & 0.1000 \\
  \midrule
  \multirow{3}{*}{Avg.} & \multirow{3}{*}{--} & EM   & \bestcell{0.5361} & 0.4950 & \secondcell{0.5093} & 0.2486 & 0.4873 & 0.4885 & 0.3436 & 0.2974 & 0.3812 & 0.3295 & 0.3856 & 0.3434 \\
  & & BLEU-1   & \bestcell{0.2171} & 0.1775 & \secondcell{0.1996} & 0.1415 & 0.1476 & 0.1831 & 0.1260 & 0.1249 & 0.1790 & 0.1040 & 0.0746 & 0.1260 \\
  & & LLM-Judge   & \bestcell{0.4017} & 0.3277 & \secondcell{0.3639} & 0.2454 & 0.2741 & 0.3190 & 0.2390 & 0.2112 & 0.2518 & 0.1114 & 0.1021 & 0.2329 \\
  \bottomrule
  \end{tabular}%
  }
\end{table*}

\begin{table*}[p]
  \centering
  \caption{
  Main results on the MemEye evaluation matrix using \texttt{gemini-2.5-flash-lite}. Columns correspond to memory methods, grouped into text-only and multimodal families. Within each coordinate, EM is reported for multiple-choice questions, while BLEU-1 and LLM-as-a-Judge (LLM-Judge) are reported for free-response questions. The first- and second-performing memory model(s) are highlighted with orange and blue backgrounds, respectively.}
  \label{tab:memeye_main_matrix_gemini_2_5_flash_lite}

  \setlength{\tabcolsep}{3.0pt}
  \renewcommand{\arraystretch}{1.05}

  \resizebox{\textwidth}{!}{%
  \begin{tabular}{lllcccccccccccc}
  \toprule
  \multirow{2}{*}{Y} & \multirow{2}{*}{X} & \multirow{2}{*}{Metric}
  & \multicolumn{7}{c}{Textual memory}
  & \multicolumn{5}{c}{Multimodal memory} \\
  \cmidrule(lr){4-10}\cmidrule(lr){11-15}
  & &
  & FC(T) & SRAG(T) & Refl. & Gen.Ag. & MemOS & A-Mem & SM(T)
  & FC(V) & SRAG(V) & MMA & M2A & SM(V) \\
  \midrule

  \multirow{12}{*}{Y1} & \multirow{3}{*}{X1} & EM   & 0.7500 & 0.7250 & 0.6000 & 0.2500 & 0.4750 & 0.5750 & 0.1500 & \secondcell{0.8750} & \bestcell{0.9000} & 0.4000 & 0.3000 & 0.3750 \\
  & & BLEU-1   & 0.2689 & 0.1777 & 0.2456 & 0.1278 & 0.1096 & 0.0672 & 0.2119 & \secondcell{0.2841} & \bestcell{0.3328} & 0.2005 & 0.0508 & 0.2099 \\
  & & LLM-Judge   & 0.5500 & 0.4000 & 0.5556 & 0.2000 & 0.2000 & 0.4000 & 0.5500 & \secondcell{0.6000} & \bestcell{0.7500} & 0.3000 & 0.2000 & 0.4500 \\
  \cmidrule(lr){2-15}
  & \multirow{3}{*}{X2} & EM   & \secondcell{0.6667} & 0.6111 & 0.5556 & 0.2500 & 0.4444 & 0.5278 & 0.1944 & \bestcell{0.7500} & 0.5833 & 0.4167 & 0.0833 & 0.3611 \\
  & & BLEU-1   & 0.1325 & 0.0613 & 0.0757 & 0.0807 & \bestcell{0.1431} & 0.0870 & 0.0000 & 0.1192 & \secondcell{0.1382} & 0.1013 & 0.0274 & 0.0190 \\
  & & LLM-Judge   & 0.4444 & 0.2778 & 0.3333 & 0.1667 & 0.2222 & 0.2778 & 0.2222 & \bestcell{0.6111} & \secondcell{0.5000} & 0.1111 & 0.1111 & 0.2222 \\
  \cmidrule(lr){2-15}
  & \multirow{3}{*}{X3} & EM   & 0.4309 & 0.3914 & 0.2599 & 0.2533 & 0.2632 & 0.2763 & 0.0757 & \secondcell{0.6776} & \bestcell{0.7500} & 0.4868 & 0.1053 & 0.0855 \\
  & & BLEU-1   & 0.2585 & 0.2203 & 0.1466 & 0.1061 & 0.0894 & 0.1035 & 0.0655 & 0.2928 & \secondcell{0.3186} & \bestcell{0.3552} & 0.0566 & 0.0293 \\
  & & LLM-Judge   & 0.3553 & 0.2632 & 0.2039 & 0.1184 & 0.0658 & 0.1579 & 0.1382 & 0.4934 & \bestcell{0.6053} & \secondcell{0.5132} & 0.0724 & 0.0987 \\
  \cmidrule(lr){2-15}
  & \multirow{3}{*}{X4} & EM   & 0.4861 & 0.4583 & 0.3472 & 0.2639 & 0.4722 & 0.5000 & 0.0972 & \secondcell{0.7917} & \bestcell{0.8194} & 0.6667 & 0.1389 & 0.0972 \\
  & & BLEU-1   & 0.2180 & 0.1982 & 0.1062 & 0.0205 & 0.0316 & 0.1875 & 0.0556 & \bestcell{0.3300} & 0.2881 & \secondcell{0.3139} & 0.0281 & 0.0556 \\
  & & LLM-Judge   & 0.4167 & 0.3056 & 0.1667 & 0.0000 & 0.0556 & 0.1389 & 0.0556 & \secondcell{0.6111} & \bestcell{0.6389} & 0.5278 & 0.0278 & 0.0556 \\
  \midrule

  \multirow{12}{*}{Y2} & \multirow{3}{*}{X1} & EM   & 0.4397 & 0.4310 & \secondcell{0.4741} & 0.2500 & 0.2759 & 0.3793 & 0.2328 & \bestcell{0.5000} & 0.4655 & 0.3879 & 0.1638 & 0.2586 \\
  & & BLEU-1   & \secondcell{0.4360} & 0.3272 & 0.3461 & 0.1401 & 0.1279 & 0.1607 & 0.2682 & 0.3290 & 0.3624 & \bestcell{0.4404} & 0.0100 & 0.3076 \\
  & & LLM-Judge   & \secondcell{0.5000} & 0.3793 & 0.3966 & 0.1724 & 0.1379 & 0.2069 & 0.3621 & 0.4138 & 0.4310 & \bestcell{0.5345} & 0.0345 & 0.3966 \\
  \cmidrule(lr){2-15}
  & \multirow{3}{*}{X2} & EM   & 0.4295 & 0.3397 & 0.4295 & 0.2500 & 0.3397 & 0.3654 & 0.2500 & \secondcell{0.4551} & \bestcell{0.5256} & 0.3654 & 0.2051 & 0.2564 \\
  & & BLEU-1   & 0.1546 & \secondcell{0.1805} & 0.1795 & 0.1065 & 0.1540 & 0.1441 & 0.1001 & 0.1254 & \bestcell{0.2260} & 0.0994 & 0.0516 & 0.1029 \\
  & & LLM-Judge   & 0.2308 & 0.2051 & 0.2436 & 0.2051 & \secondcell{0.2692} & 0.2308 & 0.1538 & 0.2564 & \bestcell{0.3333} & 0.2308 & 0.0513 & 0.1538 \\
  \cmidrule(lr){2-15}
  & \multirow{3}{*}{X3} & EM   & 0.2857 & 0.2560 & 0.2738 & 0.2440 & 0.2738 & 0.2619 & 0.0536 & \secondcell{0.3869} & \bestcell{0.4048} & 0.2619 & 0.1548 & 0.0774 \\
  & & BLEU-1   & 0.1576 & 0.1232 & 0.0838 & 0.0500 & 0.1460 & 0.1285 & 0.0096 & \bestcell{0.1930} & \secondcell{0.1762} & 0.1761 & 0.0225 & 0.0112 \\
  & & LLM-Judge   & 0.2143 & 0.1071 & 0.0789 & 0.0952 & 0.1548 & 0.1667 & 0.0238 & \bestcell{0.3810} & 0.3452 & \secondcell{0.3571} & 0.0119 & 0.0238 \\
  \cmidrule(lr){2-15}
  & \multirow{3}{*}{X4} & EM   & 0.3210 & 0.2841 & 0.2727 & 0.2500 & 0.2585 & 0.2756 & 0.0511 & \bestcell{0.4517} & \secondcell{0.3778} & 0.2955 & 0.1534 & 0.0540 \\
  & & BLEU-1   & 0.1449 & 0.1299 & 0.0660 & 0.0606 & 0.1315 & 0.0898 & 0.0357 & \bestcell{0.2468} & 0.1323 & \secondcell{0.1955} & 0.0327 & 0.0599 \\
  & & LLM-Judge   & 0.1818 & 0.1648 & 0.1354 & 0.1023 & 0.1136 & 0.1193 & 0.0568 & \bestcell{0.3011} & 0.2273 & \secondcell{0.2443} & 0.0739 & 0.0795 \\
  \midrule

  \multirow{12}{*}{Y3} & \multirow{3}{*}{X1} & EM   & \secondcell{0.7000} & 0.5500 & 0.6000 & 0.2500 & 0.3500 & \bestcell{0.7250} & 0.1000 & 0.6250 & 0.4750 & 0.4750 & 0.1750 & 0.0750 \\
  & & BLEU-1   & \secondcell{0.1971} & 0.1808 & \bestcell{0.2247} & 0.0987 & 0.1637 & 0.1656 & 0.1845 & 0.1279 & 0.1691 & 0.1685 & 0.0269 & 0.1634 \\
  & & LLM-Judge   & \bestcell{0.6000} & 0.4000 & 0.4375 & 0.3000 & 0.2500 & \secondcell{0.4500} & \secondcell{0.4500} & 0.4000 & 0.4000 & 0.4000 & 0.0500 & 0.4000 \\
  \cmidrule(lr){2-15}
  & \multirow{3}{*}{X2} & EM   & \bestcell{0.8500} & 0.6500 & \secondcell{0.8000} & 0.2500 & 0.0750 & 0.7750 & 0.1750 & 0.7500 & 0.5000 & 0.3750 & 0.1750 & 0.2000 \\
  & & BLEU-1   & 0.2189 & 0.1631 & \secondcell{0.2594} & 0.1656 & 0.1794 & \bestcell{0.2737} & 0.1901 & 0.1237 & 0.1498 & 0.0956 & 0.0385 & 0.2227 \\
  & & LLM-Judge   & \bestcell{0.6500} & 0.5000 & \secondcell{0.6000} & \secondcell{0.6000} & 0.2000 & \secondcell{0.6000} & \secondcell{0.6000} & 0.5000 & 0.3000 & 0.0500 & 0.0000 & 0.5000 \\
  \cmidrule(lr){2-15}
  & \multirow{3}{*}{X3} & EM   & 0.5000 & \secondcell{0.5750} & 0.3250 & 0.2750 & 0.4750 & 0.5250 & 0.0500 & 0.3750 & 0.5500 & \bestcell{0.7000} & 0.4250 & 0.0500 \\
  & & BLEU-1   & \secondcell{0.1423} & \bestcell{0.1601} & 0.0657 & 0.0310 & 0.0469 & 0.0709 & 0.0215 & 0.0669 & 0.0871 & 0.1157 & 0.0639 & 0.0015 \\
  & & LLM-Judge   & 0.1000 & \bestcell{0.2500} & \secondcell{0.1500} & 0.0000 & 0.0000 & 0.0500 & 0.0000 & 0.0000 & 0.0000 & \secondcell{0.1500} & 0.0000 & 0.0000 \\
  \cmidrule(lr){2-15}
  & \multirow{3}{*}{X4} & EM   & 0.2667 & 0.2333 & \secondcell{0.2833} & 0.2417 & 0.2000 & 0.2500 & 0.0250 & \bestcell{0.3000} & 0.2083 & 0.2750 & 0.0833 & 0.0500 \\
  & & BLEU-1   & 0.0490 & 0.0808 & 0.0590 & 0.0148 & 0.0966 & 0.0943 & 0.0667 & 0.0470 & \secondcell{0.1087} & \bestcell{0.2124} & 0.0105 & 0.0667 \\
  & & LLM-Judge   & 0.1333 & 0.1167 & 0.1154 & 0.0667 & 0.1000 & \bestcell{0.2500} & 0.0667 & 0.0667 & 0.1500 & \secondcell{0.2167} & 0.0000 & 0.0667 \\
  \midrule
  \multirow{3}{*}{Avg.} & \multirow{3}{*}{--} & EM   & 0.5105 & 0.4588 & 0.4351 & 0.2523 & 0.3252 & 0.4530 & 0.1212 & \bestcell{0.5782} & \secondcell{0.5467} & 0.4255 & 0.1802 & 0.1617 \\
  & & BLEU-1   & 0.1982 & 0.1669 & 0.1549 & 0.0835 & 0.1183 & 0.1311 & 0.1008 & 0.1905 & \bestcell{0.2074} & \secondcell{0.2062} & 0.0350 & 0.1041 \\
  & & LLM-Judge   & 0.3647 & 0.2808 & 0.2847 & 0.1689 & 0.1474 & 0.2540 & 0.2233 & \secondcell{0.3862} & \bestcell{0.3901} & 0.3030 & 0.0527 & 0.2039 \\
  \bottomrule
  \end{tabular}%
  }
\end{table*}

\clearpage


%
%
%
\newpage
\section{Case Studies}
\label{app:case-studies}

\paragraph{Qualitative interpretation.}
The case studies illustrate the main failure modes measured by MemEye. Cases~1-5 focus on visual-evidence loss: replacing images with captions can remove decisive details even when the surrounding dialogue is preserved. Cases~6-11 focus on evolving visual states: retrieval can surface semantically related images, but the model must still determine which visual evidence is temporally valid. The final textual-memory examples show a complementary pattern: compact state abstractions can help track updates, but they must be paired with preservation of visual evidence.

\paragraph{Case~1: Cross-session identity tracking.}
This case requires matching the same two characters across visually different scenes and sessions. The caption-only failure shows that object-category descriptions are insufficient; the model must preserve identity cues, such as body shape, color patterns, and character pairings, across time.

\paragraph{Case~2: Micro-attribute comparison.}
This question compares ceiling details across two cave scenes. Although the first caption mentions stalactites, the second omits the corresponding ceiling feature. As a result, the text-only model lacks the contrastive evidence needed to infer that the later cave contains hanging vegetation.

\paragraph{Case~3: Fine-grained color memory.}
This case isolates a pixel-level attribute that captions often discard: exact eye color. The visual model can inspect the characters directly, while the captioned representation collapses the relevant cue into a generic character description.

\paragraph{Case~4: State-evolving belief revision.}
The doctor-message example tests whether a model can track an $A{\to}B{\to}A$ update chain. The final message is visually decisive because it restores the original recommendation, but its caption is too generic. This shows that state revision can fail when the latest evidence is available only through native visual text.

\paragraph{Case~5: Game-state tracking under updates.}
The UNO example combines temporal triggering with fine-grained visual counting. The answer depends on identifying the first-hand-size transition and then counting another player's red cards at that exact state. A generic board caption does not preserve this structured visual evidence.

\paragraph{Case~6: Silent tag override.}
The fossil-tag example illustrates a stale-majority failure. Retrieval surfaces more old images than new images, so a frequency- or similarity-based strategy favors the outdated tag. The correct answer requires recognizing that the later visual state overrides the earlier one.

\paragraph{Case~7: Object migration.}
The forceps example requires tracking an object as it moves from one location to another. Both locations are semantically relevant, but only temporal ordering distinguishes the historical location from the current one. This exposes the gap between evidence retrieval and state resolution.

\paragraph{Case~8: Narrative arc tracking.}
The comic example shows that local retrieval snippets can be misleading when the answer depends on a trajectory. Individual pages contain both positive and negative events, but the correct interpretation requires integrating the sequence and judging the overall direction of change.

\paragraph{Case~9: Object reappears nearby.}
This visual-state probe shows a retrieval miss rather than a recognition failure. The backpack is absent from the old desk close-up but visible nearby in a later scene. When retrieval misses the later evidence, the model cannot recover the correct state even though the visual cue is unambiguous.

\paragraph{Case~10: Card-state trigger missed.}
This case shows that both the trigger event and the answer state are necessary. Retrieving only the later card screenshot is insufficient unless the system also identifies the preceding hand-size change that defines the relevant moment.

\paragraph{Case~11: Plastic-bag stale trap.}
The plastic-bag example shows how a visually salient old state can dominate retrieval. All three methods retrieve the earlier shelf evidence but miss the later desk evidence near the keyring, leading them to answer with the stale location rather than the updated state.






\newcommand{\vsupanelrule}{\vspace{3pt}\hrule\vspace{4pt}}

\newcommand{\vsumiss}[1]{%
\begin{minipage}[c][0.68in][c]{\linewidth}
\centering
\fbox{%
\strut\hspace{4pt}\scriptsize #1\hspace{4pt}%
}
\end{minipage}
}

\newcommand{\vsuthumb}[2]{%
\begin{minipage}[c][0.68in][c]{\linewidth}
\centering
\includegraphics[width=0.92\linewidth,height=0.38in,keepaspectratio]{#1}\\[1pt]
{\scriptsize #2}
\end{minipage}
}

\newcommand{\vsuquestion}[1]{%
\colorbox{gray!10}{%
\parbox{0.97\linewidth}{\scriptsize #1}%
}%
}

\newcommand{\vsudiag}[1]{%
\begin{minipage}[c][0.68in][c]{\linewidth}
\scriptsize #1
\end{minipage}
}

\newcommand{\vsurowlabel}[1]{%
\begin{minipage}[c][0.68in][c]{\linewidth}
\centering\textbf{#1}
\end{minipage}
}


\begin{figure*}[p]
\centering
\scriptsize
\setlength{\fboxsep}{4pt}
\setlength{\fboxrule}{0.4pt}

\fbox{%
\begin{minipage}{0.965\textwidth}
\centering

\textbf{Case 9: Object Reappears Nearby}\\[2pt]

\vsuquestion{
\textbf{Q:} Which object changed from not visible in the later desk close-up to present elsewhere nearby?
\quad
\textbf{A:} The black backpack with the silver star charm.
}

\vsupanelrule

\setlength{\tabcolsep}{3pt}
\renewcommand{\arraystretch}{1.0}
\begin{tabular}{@{}p{0.11\linewidth}p{0.29\linewidth}p{0.29\linewidth}p{0.23\linewidth}@{}}
&
\centering\textbf{E1: old desk, absent}
&
\centering\textbf{E2: latest doorway}
&
\textbf{Diagnosis}
\\[-1pt]
\multicolumn{4}{@{}c@{}}{\rule{0.97\linewidth}{0.3pt}}\\[-4pt]

\vsurowlabel{Oracle} &
\vsuthumb{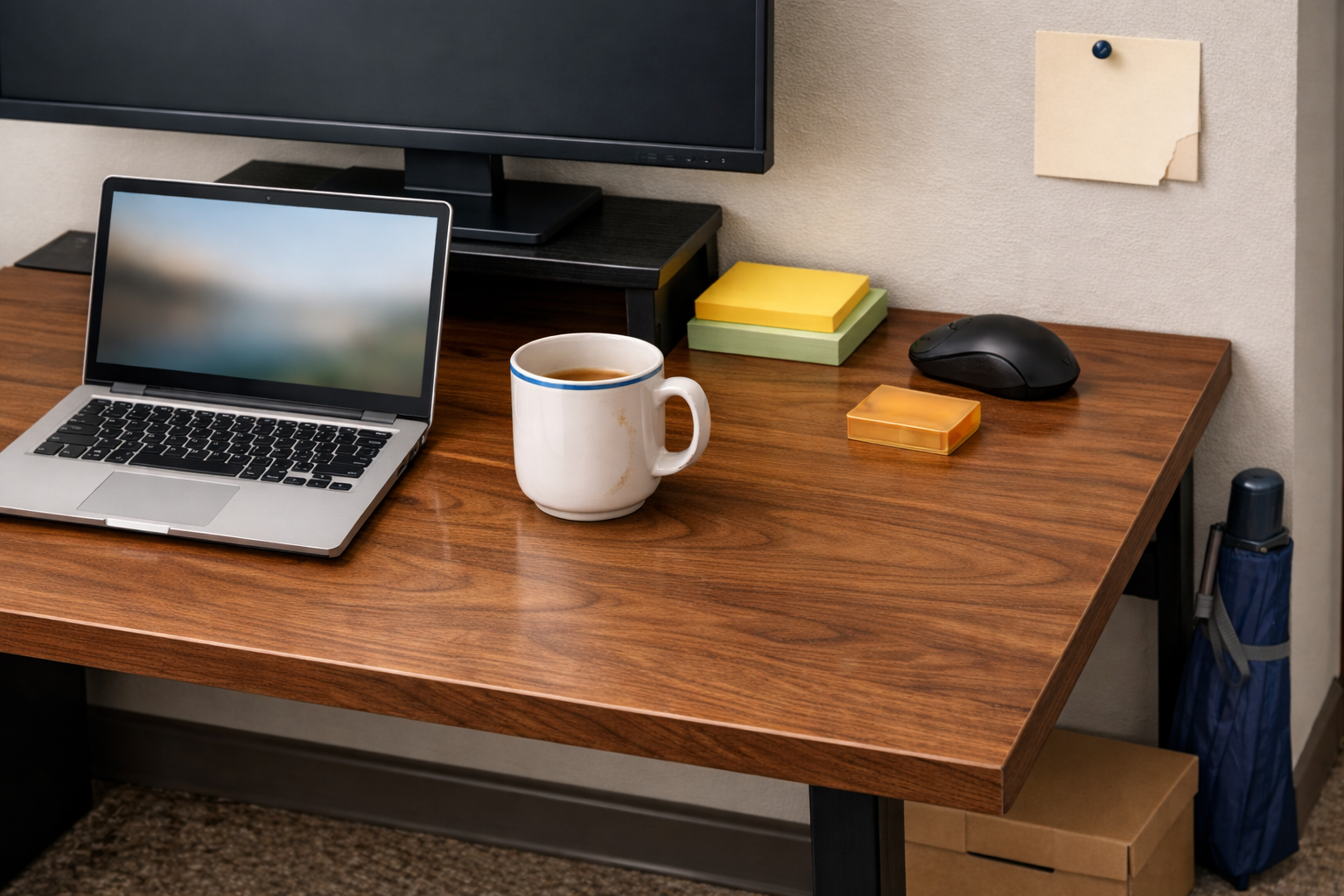}{old desk} &
\vsuthumb{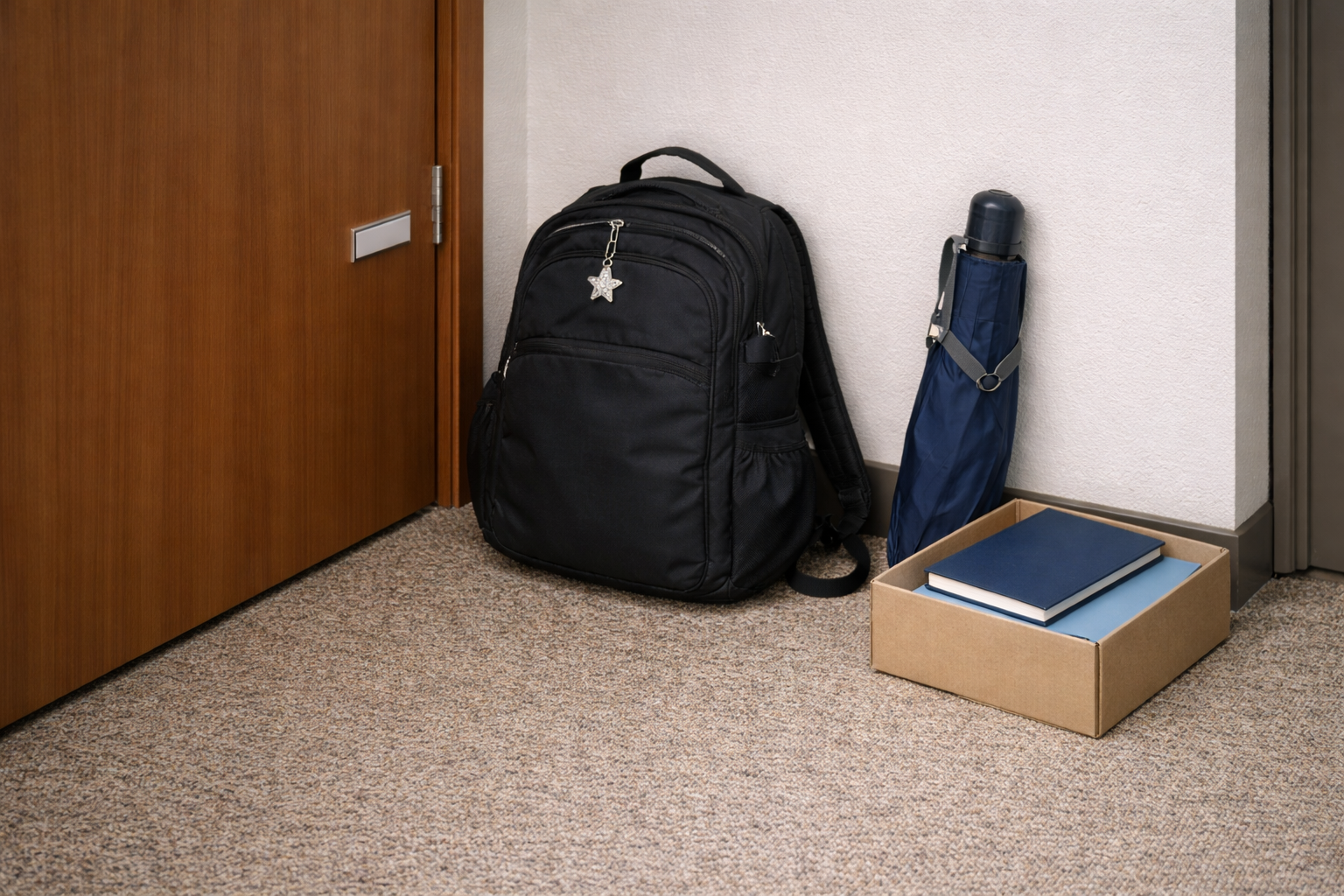}{backpack remains} &
\vsudiag{Full chain.}
\\

\vsurowlabel{SRAG(V)} &
\vsuthumb{case_study/visual_state_q7_E1_S1_R5.png}{retrieved} &
\vsumiss{missing} &
\vsudiag{Misses backpack updates; retrieves cabinet evidence and answers red USB.}
\\

\vsurowlabel{MMA} &
\vsumiss{missing} &
\vsumiss{missing} &
\vsudiag{Misses backpack chain; retrieves off-target desk evidence.}
\\

\vsurowlabel{M2A} &
\vsumiss{missing} &
\vsumiss{missing} &
\vsudiag{No backpack evidence retrieved.}
\end{tabular}

\vspace{2pt}
\end{minipage}
}

\vspace{5pt}

\fbox{%
\begin{minipage}{0.965\textwidth}
\centering

\textbf{Case 10: Card State Trigger Missed}\\[2pt]

\vsuquestion{
\textbf{Q:} How many red cards does Player~3 hold immediately after Player~2's visible hand size changes from 4 to 5 for the first time?
\quad
\textbf{A:} 3.
}

\vsupanelrule

\setlength{\tabcolsep}{3pt}
\renewcommand{\arraystretch}{1.0}
\begin{tabular}{@{}p{0.11\linewidth}p{0.29\linewidth}p{0.29\linewidth}p{0.23\linewidth}@{}}
&
\centering\textbf{E1: trigger state}
&
\centering\textbf{E2: answer state}
&
\textbf{Diagnosis}
\\[-1pt]
\multicolumn{4}{@{}c@{}}{\rule{0.97\linewidth}{0.3pt}}\\[-4pt]

\vsurowlabel{Oracle} &
\vsuthumb{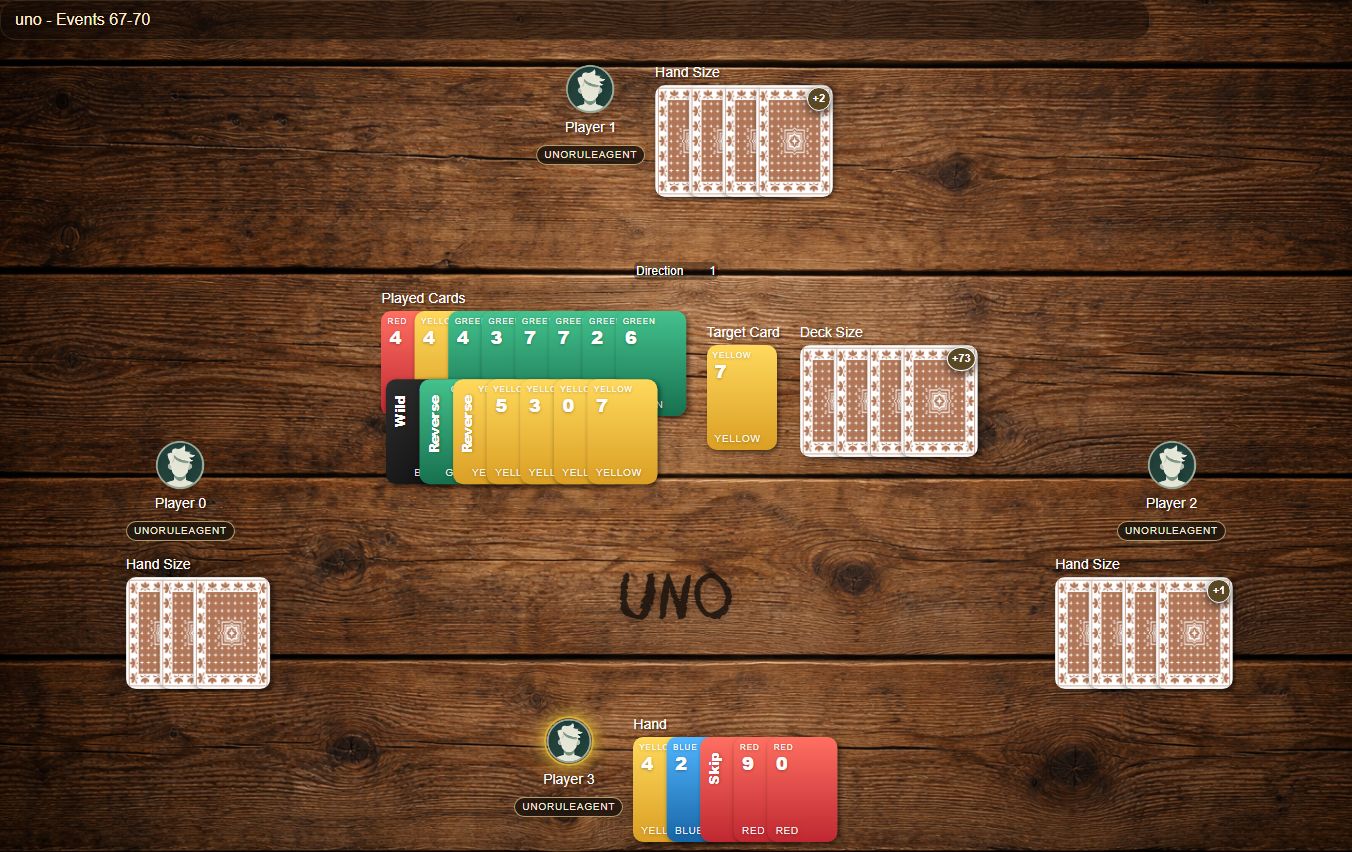}{Player 2: 4 cards} &
\vsuthumb{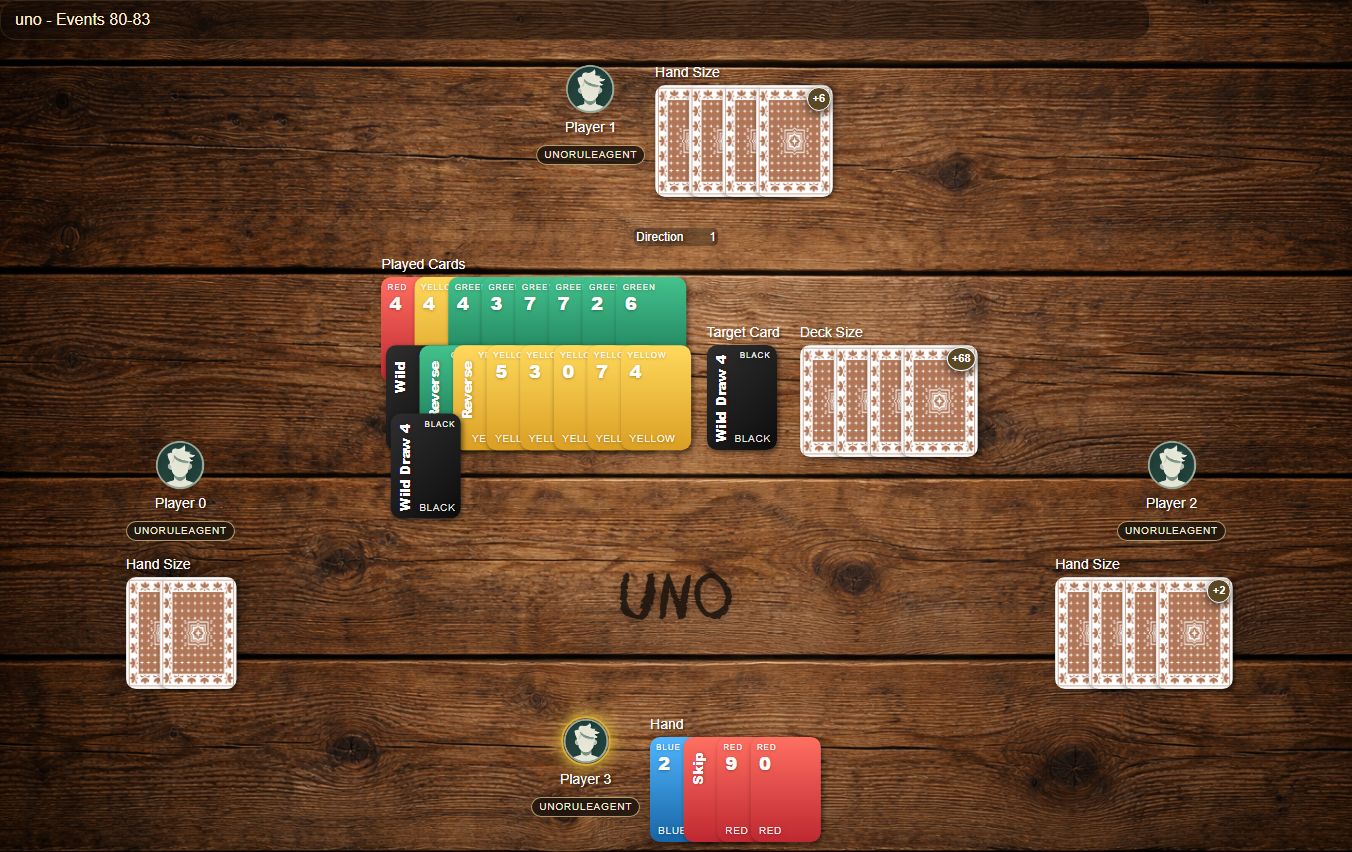}{Player 2: 5 cards} &
\vsudiag{Compare before/after.}
\\

\vsurowlabel{SRAG(V)} &
\vsumiss{missing} &
\vsuthumb{case_study/visual_state_q37_E2_PLAY_S4_R6.jpg}{retrieved} &
\vsudiag{Misses trigger; retrieves old game state and gives wrong count.}
\\

\vsurowlabel{MMA} &
\vsumiss{missing} &
\vsuthumb{case_study/visual_state_q37_E2_PLAY_S4_R6.jpg}{retrieved} &
\vsudiag{Misses trigger; retrieves old game state and gives wrong count.}
\\

\vsurowlabel{M2A} &
\vsumiss{missing} &
\vsumiss{answer state missing} &
\vsudiag{No needed game state.}
\end{tabular}

\vspace{2pt}
\end{minipage}
}

\vspace{4pt}

{\scriptsize\textit{Continued on next page.}}

\caption{
Two $Y_3$ cases where retrieval misses critical state transitions.
\textbf{Case~9}: an object reappears at a nearby location across sessions---methods retrieve stale evidence and miss the update.
\textbf{Case~10}: a card-game state trigger is missed because retrieval returns the wrong temporal snapshot.
}
\label{fig:case-studies-4}
\end{figure*}
\clearpage


\begin{figure*}[htbp]
\centering
\scriptsize
\setlength{\fboxsep}{4pt}
\setlength{\fboxrule}{0.4pt}

\fbox{%
\begin{minipage}{0.965\textwidth}
\centering

\textbf{Case 11: Plastic Bag Stale Trap}\\[2pt]

\vsuquestion{
\textbf{Q:} Which scenario describes the transparent plastic bag?
\quad
\textbf{A:} It was on the cabinet's lower shelf earlier and later appeared on the desk near the keyring.
}

\vsupanelrule

\setlength{\tabcolsep}{3pt}
\renewcommand{\arraystretch}{1.0}
\begin{tabular}{@{}p{0.11\linewidth}p{0.29\linewidth}p{0.29\linewidth}p{0.23\linewidth}@{}}
&
\centering\textbf{E1: old shelf}
&
\centering\textbf{E2: latest desk}
&
\textbf{Diagnosis}
\\[-1pt]
\multicolumn{4}{@{}c@{}}{\rule{0.97\linewidth}{0.3pt}}\\[-4pt]

\vsurowlabel{Oracle} &
\vsuthumb{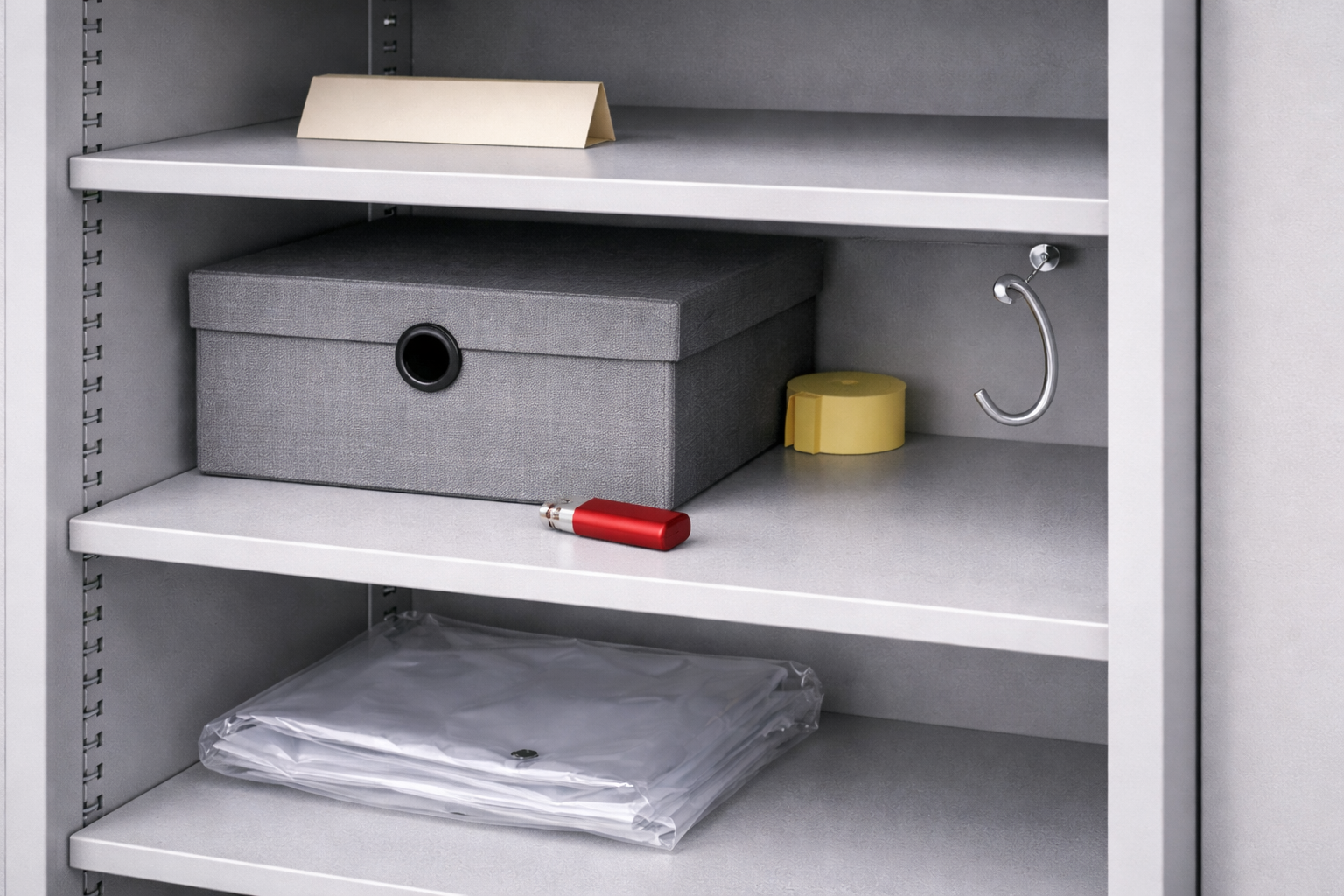}{bag on shelf} &
\vsuthumb{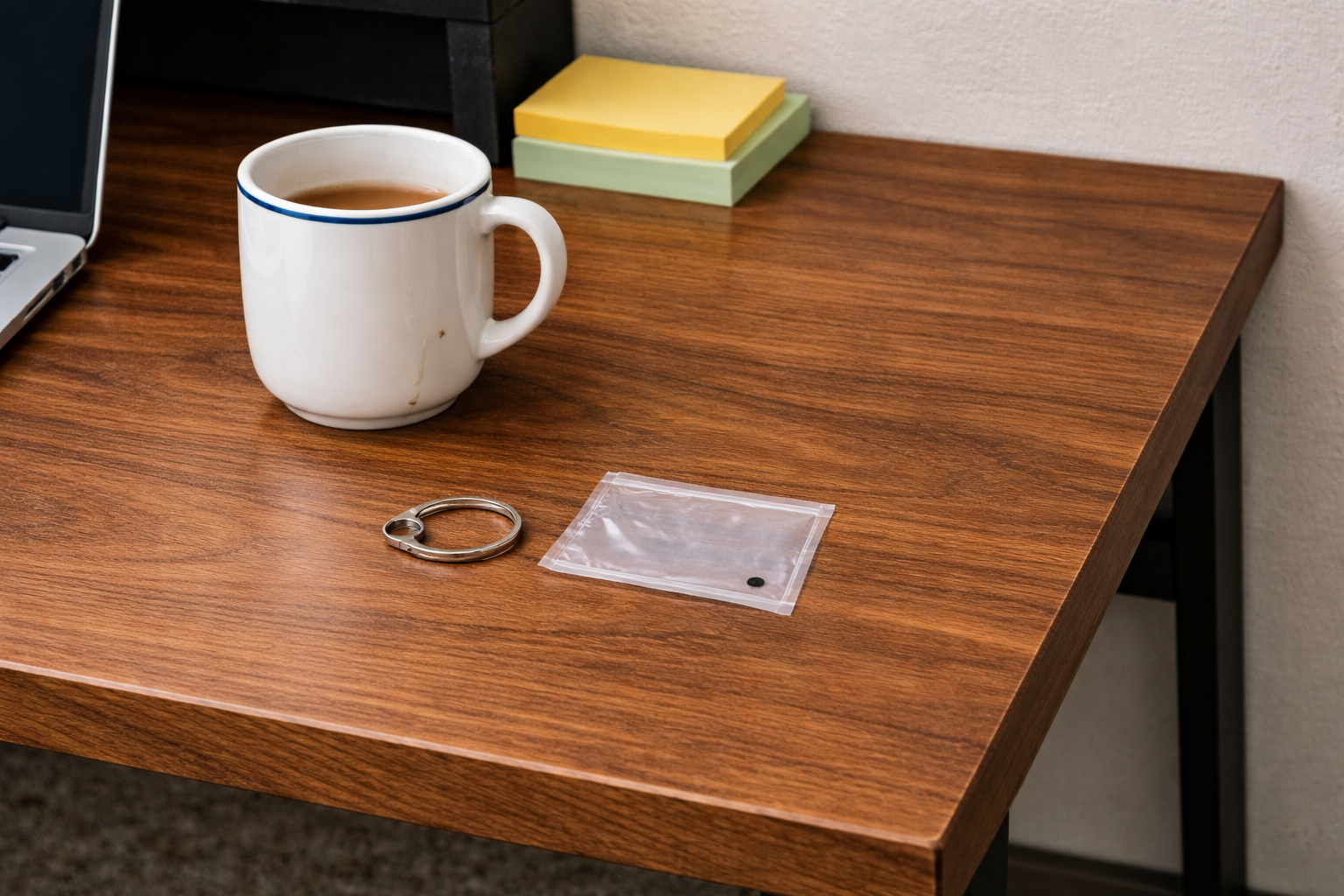}{bag near keyring} &
\vsudiag{Shelf $\rightarrow$ desk.}
\\

\vsurowlabel{SRAG(V)} &
\vsuthumb{case_study/visual_state_q8_E1_S2_R4.png}{retrieved} &
\vsumiss{missing} &
\vsudiag{Retrieves old shelf close-up; answers stale shelf state.}
\\

\vsurowlabel{MMA} &
\vsuthumb{case_study/visual_state_q8_E1_S2_R4.png}{retrieved} &
\vsumiss{missing} &
\vsudiag{Retrieves old shelf close-up; answers stale shelf state.}
\\

\vsurowlabel{M2A} &
\vsuthumb{case_study/visual_state_q8_E1_S2_R4.png}{retrieved} &
\vsumiss{missing} &
\vsudiag{Misses latest desk image.}
\end{tabular}

\vspace{2pt}
\end{minipage}
}

\vspace{4pt}

\caption{
Retrieval-error case studies from the $Y_3$ visual-state update probe.
Each panel aligns the oracle evidence chain with the answer-relevant evidence retrieved by SRAG(V), MMA, and M2A.
\textbf{Case~9} shows a nearby-object reappearance that retrieval misses.
\textbf{Case~10} shows a trigger-state miss in a card-game update.
\textbf{Case~11} shows a stale-evidence trap where the transparent plastic bag is retrieved from its earlier shelf location, while the later desk state is missing.
These examples isolate retrieval-side failures: the methods miss the decisive updated or comparison state and instead supply stale or off-target visual evidence.
Irrelevant top-$k$ distractors are omitted for readability.
}
\label{fig:visual-state-update-case-study}
\end{figure*}
\FloatBarrier

\subsection{Textual Memory Case Study}
\label{app:textual-memory-case-study}

\begin{figure*}[htbp]
\centering
\scriptsize
\setlength{\fboxsep}{5pt}
\setlength{\fboxrule}{0.4pt}

\newcommand{\ytcaseheader}[4]{%
\colorbox{gray!10}{%
\parbox{0.965\linewidth}{%
\textbf{#1}\quad
\textit{Question:} #2 \quad
\textit{Ground truth:} #3\par
\textit{Evidence chain:} #4
}%
}%
}

\newcommand{\yttextcell}[1]{%
\parbox[t]{\linewidth}{\raggedright #1}%
}

\newcommand{\ytanswercell}[1]{%
\parbox[t]{\linewidth}{\raggedright #1}%
}

\newcommand{\ytnoimage}{%
\begin{minipage}[c][0.62in][c]{\linewidth}
\centering
/
\end{minipage}
}

\newcommand{\ytimg}[2]{%
\begin{minipage}[t]{0.31\linewidth}
\centering
\includegraphics[width=\linewidth,height=0.50in,keepaspectratio]{#1}\\[-1pt]
{\tiny #2}
\end{minipage}
}

\newcommand{\ytimagecell}[3]{%
\begin{minipage}[c][0.78in][c]{\linewidth}
\centering
#1\hfill #2\hfill #3
\end{minipage}
}

\newcommand{\ytpanelrule}{\vspace{3pt}\hrule\vspace{4pt}}

\label{tab:y3-textual-memory-case-study}

\fbox{%
\begin{minipage}{0.965\textwidth}
\centering

\ytcaseheader
{Brass compass migration}
{Where is the brass compass now?}
{restoration table}
{fossil case with brass compass $\rightarrow$ compass missing from fossil case $\rightarrow$ same brass compass appears on restoration table.}

\ytpanelrule

\setlength{\tabcolsep}{3pt}
\renewcommand{\arraystretch}{1.12}

\begin{tabular}{@{}p{0.08\linewidth}p{0.36\linewidth}p{0.34\linewidth}p{0.17\linewidth}@{}}
\textbf{Method}
&
\textbf{Retrieved textual state / evidence chain}
&
\textbf{Retrieved image evidence}
&
\textbf{Answer}
\\[-1pt]
\multicolumn{4}{@{}c@{}}{\rule{0.98\linewidth}{0.3pt}}\\[-3pt]

\textbf{A-Mem}
&
\yttextcell{
\textbf{Retrieved textual states:}
(1) fossil display case with a ``brass compass'';
(2) the left-side metal object ``is still missing'' from the fossil case;
(3) the restoration table contains a ``vintage brass compass'' beside the vial and tool.
}
&
\ytnoimage
&
\ytanswercell{
\textbf{Correct.} Uses the updated restoration-table state.
}
\\[4pt]

\textbf{M2A}
&
\yttextcell{
Top semantic memories focus on ``left-side absence remains'' and nearby off-task brass/prop evidence.
The current restoration-table update is not surfaced as the merged state.
}
&
\ytimagecell
{\ytimg{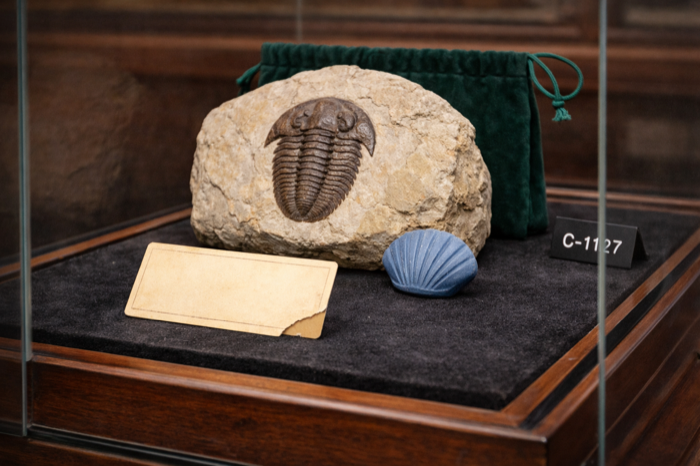}{stale fossil absence}}
{\ytimg{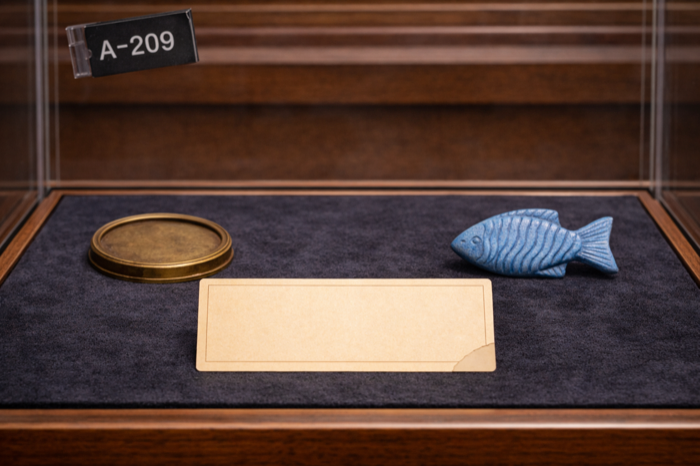}{off-task brass object}}
{\ytimg{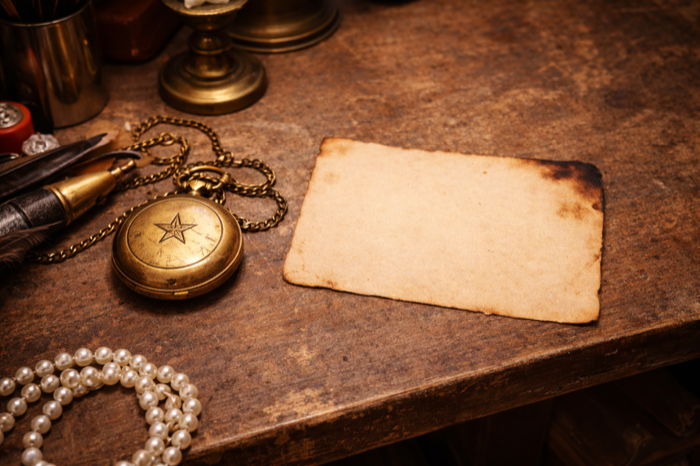}{prop watch distractor}}
&
\ytanswercell{
\textbf{Wrong.} Says the compass is missing or points to an off-target table.
}
\\[4pt]

\textbf{MMA}
&
\yttextcell{
Retrieves raw fossil-room image turns and visually similar later shots, but no explicit ``compass moved to restoration table'' state.
}
&
\ytimagecell
{\ytimg{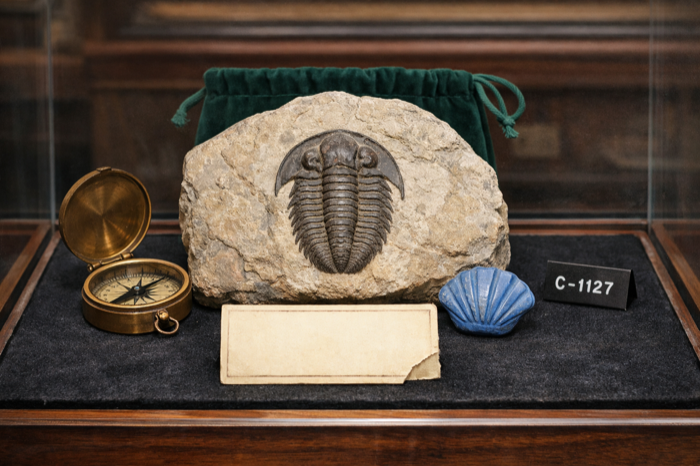}{old fossil case}}
{\ytimg{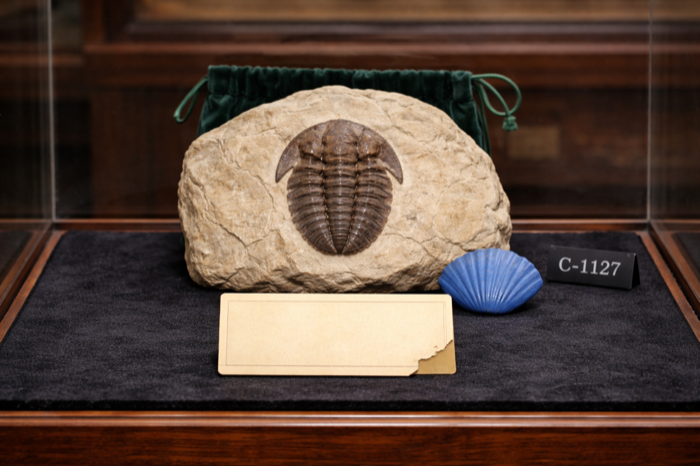}{missing state}}
{\ytimg{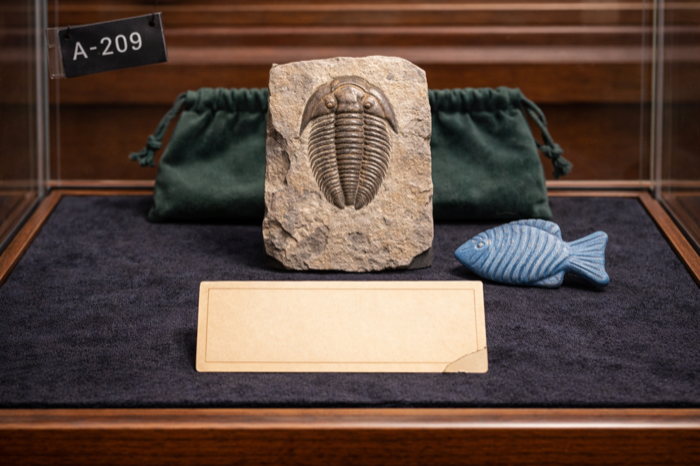}{later fossil room}}
&
\ytanswercell{
\textbf{Wrong.} Keeps a stale/off-target location.
}
\end{tabular}

\vspace{2pt}
\end{minipage}
}

\vspace{6pt}

\fbox{%
\begin{minipage}{0.965\textwidth}
\centering

\ytcaseheader
{Sage-green paint rejected}
{Which tested paint color was not used?}
{sage green}
{sage-green paint test $\rightarrow$ design pivots to terracotta $\rightarrow$ final room uses terracotta, so sage green is the tested color not used.}

\ytpanelrule

\setlength{\tabcolsep}{3pt}
\renewcommand{\arraystretch}{1.12}

\begin{tabular}{@{}p{0.08\linewidth}p{0.36\linewidth}p{0.34\linewidth}p{0.17\linewidth}@{}}
\textbf{Method}
&
\textbf{Retrieved textual state / evidence chain}
&
\textbf{Retrieved image evidence}
&
\textbf{Answer}
\\[-1pt]
\multicolumn{4}{@{}c@{}}{\rule{0.98\linewidth}{0.3pt}}\\[-3pt]

\textbf{A-Mem}
&
\yttextcell{
\textbf{Retrieved textual states:}
(1) a ``sage green paint'' swatch is tested on the wall;
(2) a later wall test uses ``terracotta paint'';
(3) the final living room has a ``terracotta accent wall.''
}
&
\ytnoimage
&
\ytanswercell{
\textbf{Correct.} Names sage green as tested but rejected.
}
\\[4pt]

\textbf{M2A}
&
\yttextcell{
Retrieves the green test and final-room evidence, but abstracts the color as ``muted olive-green'' or generic ``green wall paint,'' losing the exact label.
}
&
\ytimagecell
{\ytimg{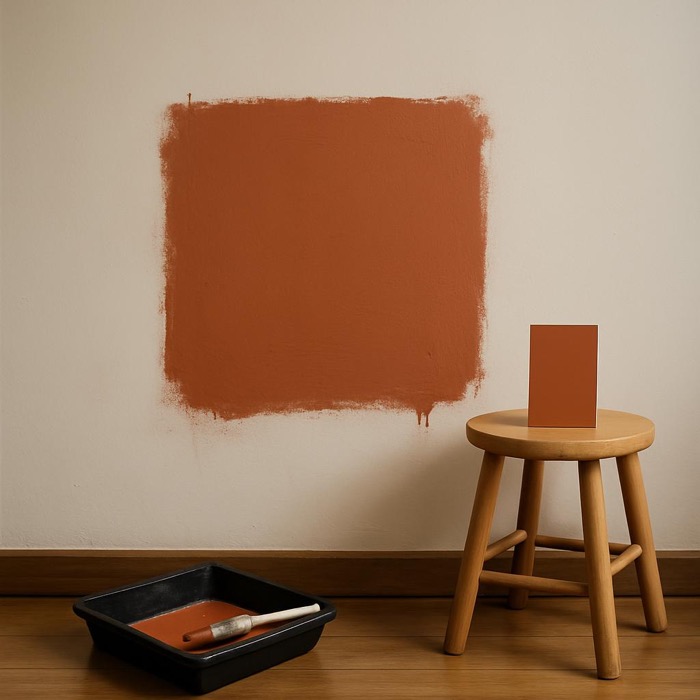}{terracotta test}}
{\ytimg{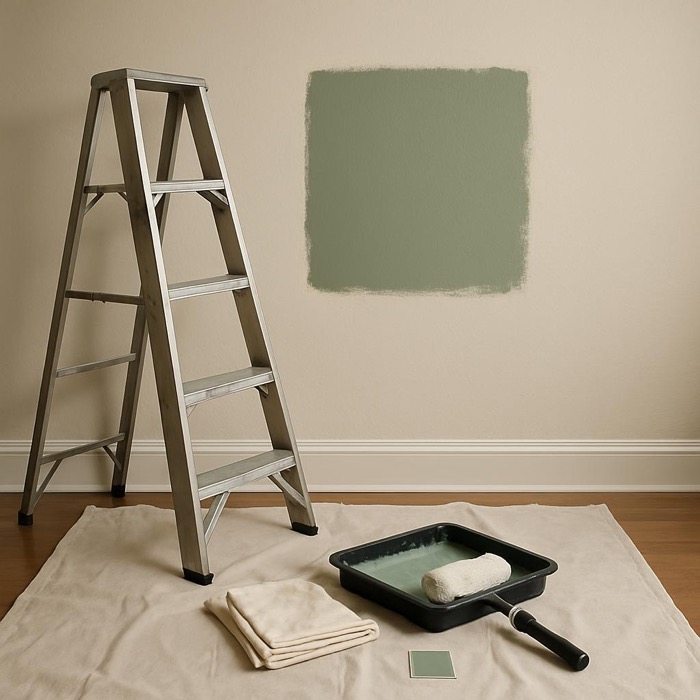}{green test}}
{\ytimg{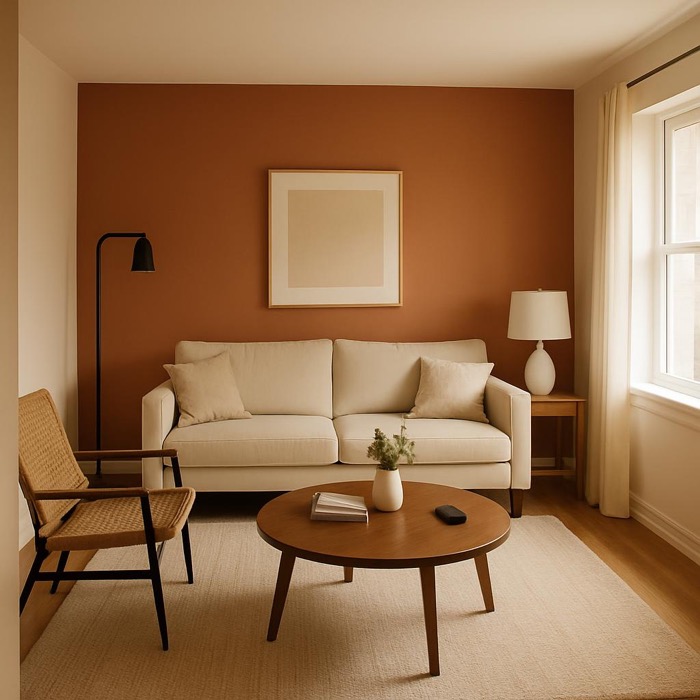}{final room}}
&
\ytanswercell{
\textbf{Wrong.} Loses the exact sage-green answer.
}
\\[4pt]

\textbf{MMA}
&
\yttextcell{
Retrieves separate raw visual entries for terracotta testing, the earlier green test, and the final room, but does not extract ``tested but not final.''
}
&
\ytimagecell
{\ytimg{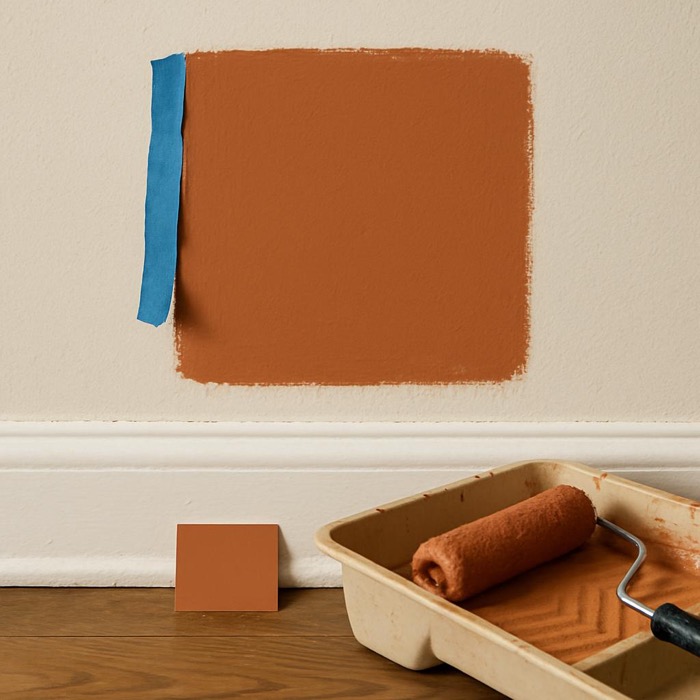}{terracotta test}}
{\ytimg{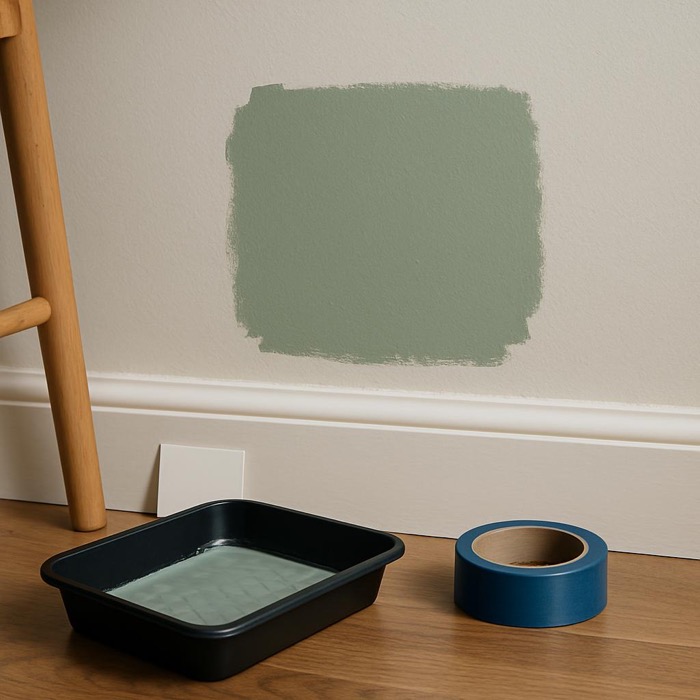}{sage test}}
{\ytimg{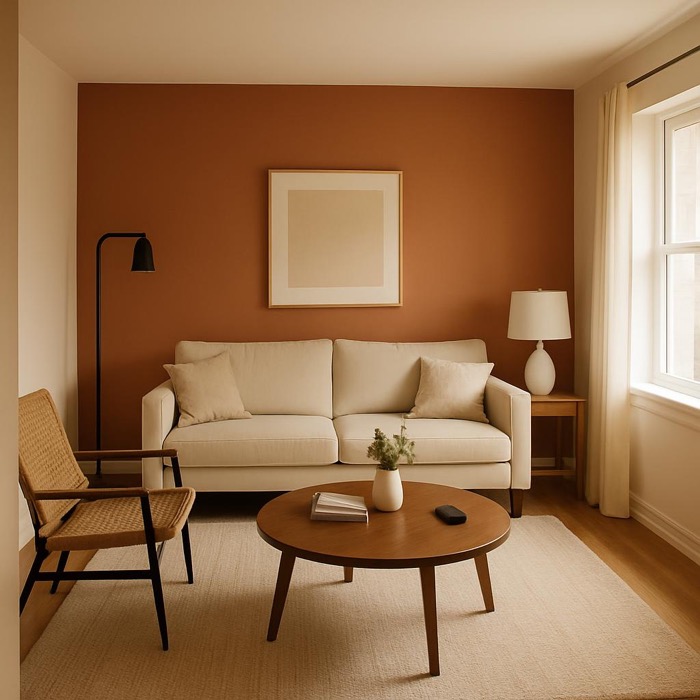}{final room}}
&
\ytanswercell{
\textbf{Wrong.} Selects terracotta, the final color, rather than the rejected test color.
}
\end{tabular}

\vspace{2pt}
\end{minipage}
}
\caption{
Case study of textual memory extraction and image retrieval for two $Y_3$ evolving-state cases.
Each panel first specifies the required evidence chain, then compares the evidence retrieved by A-Mem, M2A, and MMA at answer time.
}
\label{fig:y3-textual-memory-case-study}
\end{figure*}

Figure~\ref{fig:y3-textual-memory-case-study} presents two $Y_3$ case studies where text-based memory preserves the evolving state chain while multimodal memory retrieves stale, conflicting, or visually similar evidence. These cases suggest that the advantage of A-Mem in some $Y_3$ scenarios comes from compact state extraction rather than richer visual recall: when the answer depends on tracking how visual states evolve, a structured textual evidence chain can be more reliable than retrieving visually similar raw images.

\end{document}